\documentclass[11pt,twocolumn]{article}

\usepackage{natbib}
\usepackage{framed,multirow}
\usepackage{authblk}

\usepackage{amssymb}
\usepackage{latexsym}
\usepackage{lipsum,pdflscape}

\usepackage{url}
\usepackage{xcolor}

\usepackage{color}
\usepackage{multirow}
\usepackage{diagbox}
\usepackage{algorithm}
\usepackage{algorithmic}
\usepackage{graphicx}
\usepackage{stackengine}
\usepackage[caption=false,font=footnotesize,labelfont=sf,textfont=sf]{subfig}
 
\newcommand{\etal}{\textit{et al}.}
\newcommand{\ie}{\textit{i}.\textit{e}., }
\newcommand{\eg}{\textit{e}.\textit{g}., }

\def\mat#1{\mathchoice{\mbox{\boldmath $\displaystyle\tt#1$}}
{\mbox{\boldmath$\textstyle\tt#1$}}
{\mbox{\boldmath$\scriptstyle\tt#1$}}
{\mbox{\boldmath$\scriptscriptstyle\tt#1$}}}

\def\vect#1{\mathchoice{\mbox{\boldmath $\displaystyle\bf#1$}}
{\mbox{\boldmath  $\textstyle\bf#1$}}
{\mbox{\boldmath  $\scriptstyle\bf#1$}}
{\mbox{\boldmath  $\scriptscriptstyle\bf#1$}}}

\def\vp{{\vect p}}

\def\vr{{\vect r}}

\def\vt{{\vect t}}

\def\vv{{\vect v}}
\def\vw{{\vect w}}
\def\vx{{\vect x}}

\def\vX{{\vect X}}

\def\vmu{{\vect\mu}}

\def\v0{{\vect 0}}

\def\mDelta{{\mat\mDelta}}

\def\m1{{\mat 1}}

\def\mI{{\mat I}}

\def\mR{{\mat R}}

\def\mX{{\mat X}}

\def\Reales{\mathbb{R}}

\begin{document}

\title{Face Alignment using a 3D Deeply-initialized Ensemble of Regression Trees}

\author[1,4]{Roberto Valle}
\author[2,4]{Jos\'e M. Buenaposada}
\author[3]{Antonio Vald\'es}
\author[1,4]{Luis Baumela}
\affil[1]{Universidad Polit\'ecnica de Madrid, \texttt{\{rvalle,lbaumela\}@fi.upm.es}}
\affil[2]{Universidad Rey Juan Carlos, \texttt{josemiguel.buenaposada@urjc.es}}
\affil[3]{Universidad Complutense de Madrid, \texttt{avaldes@ucm.es}}
\affil[4]{\texttt{http://www.dia.fi.upm.es/$\sim$pcr}}

\date{\vspace{-5ex}}


\twocolumn[
\begin{@twocolumnfalse}
\maketitle
\begin{abstract}
\end{abstract}
Face alignment algorithms locate a set of landmark points in images of faces taken in unrestricted situations. State-of-the-art approaches typically fail or lose accuracy in the presence of occlusions, strong deformations, large pose variations and ambiguous configurations. In this paper we present 3DDE, a robust and efficient face alignment algorithm based on a coarse-to-fine cascade of ensembles of regression trees. It is initialized by robustly fitting a 3D face model to the probability maps produced by a convolutional neural network. With this initialization we address self-occlusions and large face rotations. Further, the regressor implicitly imposes a prior face shape on the solution, addressing occlusions and ambiguous face configurations. Its coarse-to-fine structure tackles the combinatorial explosion of parts deformation. In the experiments performed, 3DDE improves the state-of-the-art in 300W, COFW, AFLW and WFLW data sets. Finally, we perform cross-dataset experiments that reveal the existence of a significant data set bias in these benchmarks.
\end{@twocolumnfalse}
\vspace{1cm}
]




\section{Introduction}\label{sec:introduction}
Face alignment algorithms precisely locate a set of points of interest in the images of faces taken in unrestricted conditions. It has received much attention from the research community~\citep{Jin17} since it is a preliminary step for estimating 3D facial structure~\citep{Zhao16} and many other face image analysis problems such as verification and recognition~\citep{Soltanpour17}, attributes estimation~\citep{Bekios14} or facial expression recognition~\citep{Martinez12}, to name a few. Present approaches typically fail or lose precision in the presence of occlusions, strong deformations produced by facial expressions, large pose variations and ambiguous configurations caused, for example, by strong make-up or the existence of other nearby faces.

Top performers in the most popular benchmarks are based on Convolutional Neural Networks (CNNs) and Ensemble of Regression Trees (ERT), see \eg Tables~\ref{table:300w_public},~\ref{table:300w_private},~\ref{table:cofw},~\ref{table:aflw} and~\ref{table:wflw}. The large effective receptive field of deep models~\citep{Kowalski17,Lv17,Xiao16,Yang17,Wu18} enable them to model context better and produce robust landmark estimations.
However, in these models it is not easy to enforce facial shape consistency, something that limits their accuracy in the presence of occlusions and ambiguous facial configurations. ERT-based models~\citep{Burgos13,Cao14,Kazemi14,Lee15b,Ren16}, on the other hand, are difficult to initialize, but may implicitly impose face shape consistency in their estimations~\citep{Cao14}. This increases their performance in occluded and ambiguous situations. They are also much more efficient than deep models and, as we demonstrate in our experiments, with a good initialization they are also very accurate.

In this paper we present the 3DDE (3D Deeply-initialized Ensemble) regressor, a robust and efficient face alignment algorithm based on a coarse-to-fine cascade of ERTs. It is a hybrid approach that inherits good properties of ERT,  
such as the ability to impose a face shape prior, and the robustness of deep models. It is initialized by robustly fitting a 3D face model to the probability maps produced by a CNN. With this initialization we tackle one of the main drawbacks of ERT, namely the difficulty in initializing the regressor in the presence of occlusions and large face rotations. On the other hand, the ERT implicitly imposes a prior face shape on the solution, addressing the shortcomings of deep models when occlusions and ambiguous face configurations are present. Finally, its coarse-to-fine structure tackles the combinatorial explosion of parts deformation, which is also a key limitation of approaches using shape constraints~\citep{Cao14}.

A preliminary version of our work appeared in~\cite{Valle18}. Here we refine and extend it in several ways. First we improve the initialization by using a RANSAC-like procedure that increases its robustness in the presence of occlusions. We have also introduced early stopping and better data augmentation techniques for increasing the regularization when training both the ERT and the CNN. We also extend the evaluation including the newly released WFLW data base and a detailed ablation study. Finally, 3DDE may also be trained in presence of missing and occluded landmarks in the training set. This has enabled us to perform cross-dataset experiments that reveal the existence of significant data set bias that may limit the generalization capabilities of regressors trained on present data bases. To the best of our knowledge, this is the first time such a problem has been raised in the field.


\section{Related Work}\label{sec:related_work}

Face alignment has been a topic of intense research for more than twenty years. Initial successful results were based on 2D and 3D generative approaches such as the Active Appearance Models (AAM)~\citep{Cootes98} or the 3D Morphable Models (3DMM)~\citep{Blanz03}. Recent approaches are based on a cascaded combination of discriminative regressors. 

In the earliest case these regressors are Random Ferns~\citep{Dollar10}, Ensembles of Regression Trees~\citep{Cao12} or linear models~\citep{Xiong13,Xiong15}. Key ideas in this approach are indexing image description relative to the current shape estimate~\citep{Dollar10}, and the use of a regressor whose predictions lie on the subspace spanned by the training face shapes~\citep{Cao14}, this is the so-called Cascade Shape Regressor (CSR) framework.  \cite{Kazemi14} improved the original cascade framework by proposing a real-time ensemble of regression trees.  \cite{Ren16} used locally binary features to boost the performance up to 3000 FPS. \cite{Burgos13} included occlusion estimation and decreased the influence of occluded landmarks. \cite{Shen14} refine the initial location of face landmarks using a random forest and SIFT features. \cite{Xiong13,Xiong15} also use SIFT features and learn the linear regressor dividing the search space into individual regions with similar gradient directions. Overall, this set of approaches are very sensitive to the starting point of the regression process. For this reason an important part of recent work revolves around how to find good initializations~\citep{Zhu15,Zhu16a}. However, they are extremely efficient and may take advantage of implicit shape constraints~\citep{Cao12,Cao14}.

The recent development of deep learning techniques has also impacted the face alignment field with the widespread use of CNN-based regressors. \cite{Sun13} were pioneers to apply a three-level CNN for locating landmarks. \cite{Zhang14b} proposed a multi-task solution to deal with face alignment and attributes classification. \cite{Lv17} use global and local face parts regressors for fine-grained facial deformation estimation. \cite{Yu16} transform the landmarks rather than the input image for the refinement cascade. \cite{Trigeorgis16} and \cite{Xiao16} are the first approaches that fuse the feature extraction and regression steps of CSR into a recurrent neural network trained end-to-end. \cite{Kowalski17} and \cite{Yang17} use a global similarity transform to normalize landmark locations followed by a VGG-based and a Stacked Hourglass network respectively to regress the final shape. \cite{Wu18} derive face landmarks from boundary lines, which helps to remove the ambiguities in the landmark definition. Deep CNN models have large effective receptive fields that let them model context better and convey these approaches with a high degree of robustness to face rotation, scale, deformation and initialization. However, when used in a cascaded framework they may notably increase the computational requirements. Moreover, it is not clear how to impose facial shape consistency on the estimated set of landmarks. Hence, the regressor accuracy may be harmed in the presence of occlusions or ambiguities.

There is also an increasing number of works based on 3D face models. In the simplest case, they fit a mean model to the estimated image landmarks positions~\citep{Kowalski16} or jointly regress the pose and shape of the face~\citep{Jourabloo17,Xiao17}. \cite{Zhu17} and \cite{Kumar18a} fit a 3DMM in a cascaded way. These approaches provide 3D pose information that may be used to estimate landmark self-occlusions or to train simpler regressors specialized in a given head orientation. However, building and fitting a 3D face model is a difficult task and the results of the full 3D approaches in current benchmarks are not as good as those described above. 

Our proposal tries to leverage on the good properties of the three approaches described above. Using a CNN-based initialization we inherit the robustness of deep learning models. Like the simple 3D approaches we fit a rigid 3D face model to initialize the regressor and estimate the initial face orientation to address self-occlusions and ambiguities. Finally, we use a cascaded ERT within a coarse-to-fine framework to achieve accuracy and efficiency while avoiding the combinatorial explosion of independent parts deformations.


\section{3D deeply-initialized Ensemble}\label{sec:algorithm}
\begin{figure*}
\centering
\includegraphics[width=0.8\textwidth]{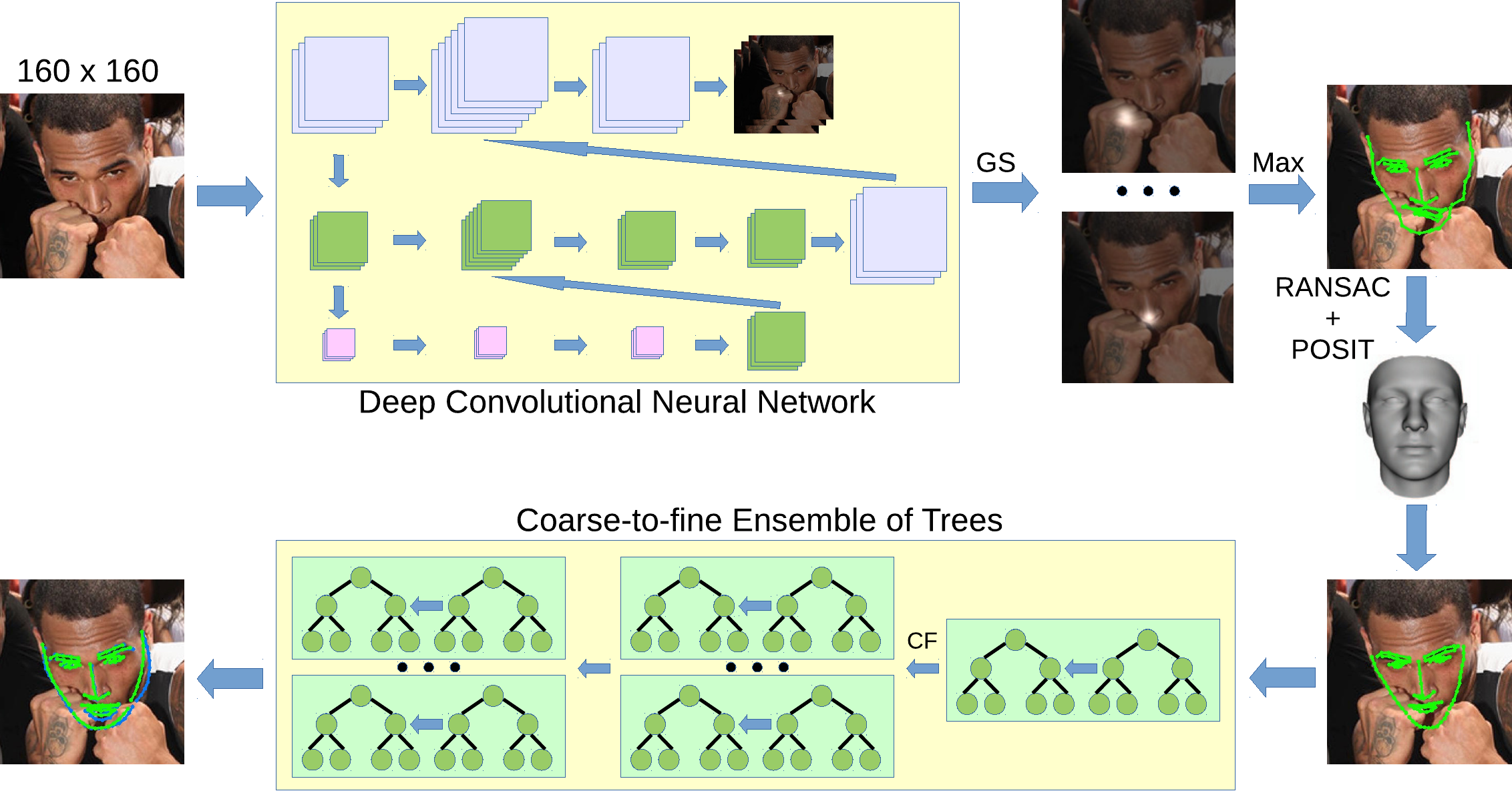}
\caption{3DDE framework diagram. GS, Max and RANSAC+POSIT represent the Gaussian smoothing filter, the maximum of each probability map and the robust 3D pose estimation respectively.}
\label{fig:3dde}
\end{figure*}

In this section we present 3DDE. It consists of two main steps: CNN-based rigid face pose computation and ERT-based non-rigid face deformation estimation, both shown in Fig.~\ref{fig:3dde}.

\subsection{Rigid pose computation}
\label{sec:rigid}

ERT-based regressors require a good initialization to converge. We propose the use of face landmarks location probability maps~\citep{Belhumeur11,Dantone12,Xiao16} to generate plausible shape initialization candidates. We define a UNet-like architecture~\citep{Ronneberger15,Honari16}, with a loss function that handles missing landmarks. We train this CNN to obtain a set of probability maps, $\mathcal{P}(\mI)$, that model the position of each landmark in the input image (see Fig.~\ref{fig:3dde}). The maximum of each smoothed probability map determines our initial landmark positions. Note in Fig.~\ref{fig:3dde} that these predictions are sensitive to occlusions, ambiguities and may not be a valid face shape. Compared to typical CNN-based approaches, \eg~\cite{Yang17}, our CNN is much simpler, since we only require a rough estimation of landmark locations.

To start the ERT with a plausible face, we compute the initial shape by fitting a rigid 3D head model to the estimated 2D landmarks locations. To this end we use the softPOSIT algorithm proposed by \cite{David04} within a robust scheme. Unlike~\cite{Valle18}, here we use a set of the distinct landmarks to establish the correspondences between the CNN predictions and the 3D face model. This avoids problems related to ambiguous landmarks around the jaw that do not correspond always to the same 3D points and produce wrong initializations, mainly in profile faces. Moreover, we have also implemented a RANSAC-like procedure, that runs softPOSIT several times with subsets of correspondences, to obtain a robust estimation (see Algorithm~\ref{alg:initial}). 

\begin{algorithm}
\footnotesize
\caption{Initialization algorithm ($g_0$)}
\label{alg:initial}
\begin{algorithmic}
\renewcommand{\algorithmicrequire}{\textbf{Input:}}
\renewcommand{\algorithmicensure}{\textbf{Output:}}
\REQUIRE {$\mathcal{P}(\mI)$, $\vX$ }
\STATE{// Select coordinates of maximum probability}
\STATE{$\{\vx(l)$ = $\arg\max(\mathcal{P}^l(\mI))\}_{l=1}^L$}
\STATE { $p^* = 0$}
\FOR{z=1 \TO $Z$}
\STATE {// Select subset from distinct landmarks}
\STATE {$\vx_s, \vX_s$ = chooseLandmarksSubset($\vx$, $\vX$)}
\vskip 2ex
\STATE {// Compute projection matrix between $\vx_s, \vX_s$}
\STATE {$\mR, \vt$ = softPOSIT($\vx_s$, $\vX_s$)}
\vskip 2ex
\STATE {// Project 3D face model using previous matrix}
\STATE {$\vx_z,\vv_z$ = projectPoints($\vX$, $\mR$, $\vt$)}
\vskip 2ex
\STATE {// Evaluate the goodness of the initialization}
\STATE {$p(\vx_z) = \sum_{l=1}^L \mathcal{P}^l(\mI)[\vx_z(l)]$}
\IF{$p(\vx_z) > p^*$}
\STATE{$p^* = p(\vx_z)$, $\mR^* = \mR$, $\vt^* = \vt$}
\ENDIF
\ENDFOR
\STATE {$\vx^0,\vv^0$ = projectPoints($\vX$, $\mR^*$, $\vt^*$)} 
\ENSURE $\vx^0,\vv^0$
\end{algorithmic}
\end{algorithm}

Let $\vX\in\Reales^{L\times 3}$ be the 3D coordinates of the $L$ landmarks on the 3D face model, $\vx\in\Reales^{L\times 2}$ their 2D projections onto the image plane and $\vv\in\{0,1\}^{L}$ their visibilities. 
We produce subsets of correspondences $(\vx_s,\vX_s)$ from the distinct landmarks shown in Fig.~\ref{fig:cross:a}, estimate the 3D face model pose $(\mR,\vt)$ with softPOSIT and evaluate the goodness of each estimation as the sum of landmarks probabilities, 
\[
p(\vx_z) = \sum_{l=1}^L \mathcal{P}^l(\mI)[\vx_z(l)],
\]
where $\vx_z(l)$ are the 2D coordinates of the $l$-th landmark and $\mathcal{P}^l(\mI)$ is the probability map for landmark $l$. Finally, we select the rigid transformation $(\mR, \vt)$ with highest $p(\vx_z)$. As a result, we project the 3D model onto the image using the most likely estimated rigid transformation. This provides the ERT with a rough estimation of the scale, translation and 3D pose of the target face (see Fig.~\ref{fig:3dde}), and the visibility estimation of the self-occluded parts of the face.

Let $\vx^0=g_0(\mathcal{P}(\mI), \mX)$ be the \emph{initial shape}, the output of the initialization function $g_0$ after processing the input image $\mI$. With our initialization we enforce two key requirements for the convergence of the ERT. First, that $\vx^0$ lies on the face with an approximately correct 3D face pose. Second, that $\vx^0$ is a valid face shape. The latter guarantees that the predictions in the next step of the algorithm will also be valid face shapes~\citep{Cao14}.

\subsection{ERT-based non-rigid shape estimation}
\label{sec:non_rigid}

Let $\mathcal{S}=\{s_i\}_{i=1}^N$ be the set of training face shapes, where $s_i=(\mI_i,\vx^g_i, \vv_i^g, \vw_i^g,\vx_i^0,\vv_i^0)$. Each shape $s_i$ has its own training image, $\mI_i$, ground truth shape, $\vx_i^g$, ground truth visibility label, $\vv_i^g$, annotated landmark label, $\vw_i^g\in\{0,1\}^L$, initial shape, $\vx_i^0$, and visibilities, $\vv_i^0$, for training the ERT regressor. 
In our implementation we use shape-indexed features~\citep{Lee15b}, $\phi(\mathcal{P}(\mI_i), \vx_i^t, \vw_i^g)$, that depend on the current shape $\vx_i^t$ of the landmarks in image $\mI_i$ and whether they are annotated or not, $\vw_i^g$.

We divide the regression process into a maximum of $T$ stages. We learn an ensemble of $K$ regression trees for the $t$-th stage, $\mathcal{C}_t(f_i) = \vx_i^{t-1} + \sum_{k=1}^K g_k(f_i)$, where $f_i=\phi(\mathcal{P}(\mI_i), \vx_i^{t-1}, \vw_i^g)$ and $\vx^j$ are the coordinates of the landmarks estimated in $j$-th stage. To train the ERT we use the $N$ training shapes in $\mathcal{S}$ to generate an augmented training set of samples, $\mathcal{S}_A$, and a validation set, $\mathcal{S}_V$, with cardinality $N_A=|\mathcal{S}_A|$ and $N_V=|\mathcal{S}_V|$ respectively. The total number of samples is $N_T=N_A+N_V$. Instead of using a fixed number of stages, like~\cite{Valle18}, we stop training when the validation error stops improving. In this way the regressor has a variable number of stages. We compute the initialization for each sample using the 3D projections produced by $g_0$ (see generated initializations in Fig.~\ref{fig:initials}). We also improve the data augmentation used in~\cite{Valle18}. 
To this end we add random noise to the yaw, pitch and roll angles, of
the rotation matrix $\mR^*$ estimated with $g_0$, to generate new training initializations for each sample in $\mathcal{S}_A$.

Following \etal~\cite{Burgos13} and \cite{Kazemi14}, we attach to each landmark in $\mathcal{S}$ the binary labels $\{\vv,\vw\}\in\{0,1\}$ that model respectively whether it is visible and annotated. We learn these labels in the ERT together with the landmark location. Each initial shape is progressively refined by estimating a shape and visibility increments $\mathcal{C}^{\vv}_t(\phi(\mathcal{P}(\mI_i), \vx^{t-1}_i, \vw^g_i))$ where $\vx^{t-1}_i$ represents the current shape of the $i$-th sample (see Algorithm~\ref{alg:ert}). $\mathcal{C}^{\vv}_t$ is trained to minimize only the landmark position errors but on each tree leaf, in addition to the mean shape, we also output the mean of all training shapes visibilities, $\vv^g_i$, that belong to that node. We define $\mathcal{A}_{t-1} = \{ (\vx_i^{t-1}, \vv_i^{t-1} )\}_{i=1}^{N_A}$ and $\mathcal{V}_{t-1} = \{ (\vx_i^{t-1}, \vv_i^{t-1} )\}_{i=1}^{N_V}$ as the set of all current shapes and corresponding visibility vectors for all training and validation data, respectively. 

\begin{algorithm}
\footnotesize
\caption{Training an Ensemble of Regression Trees}
\label{alg:ert}
\begin{algorithmic}
\renewcommand{\algorithmicrequire}{\textbf{Input:}}
\renewcommand{\algorithmicensure}{\textbf{Output:}}
\REQUIRE  { $\mathcal{S}$, $T$}
\STATE {// Generate an augmented training set of samples}
\STATE {$\mathcal{S}_A, \mathcal{S}_V$ = dataAugmentation($\mathcal{S}$)}
\REPEAT
\STATE {// Extract training ($\mathcal{F}_A$) and validation ($\mathcal{F}_V$) features}
\STATE {$\mathcal{F}_A \cup \mathcal{F}_V = \{f_i\}_{i=1}^{N_T} = \{\phi(\mathcal{P}(\mI_i), \vx^{t-1}_i, \vw_i^g)\}_{i=1}^{N_T}$}
\vskip 2ex
\STATE {// Apply Algorithm~\ref{alg:p_parts_regressors} using training samples}
\STATE {$\mathcal{C}^{\vv}_t$ = learnCoarseToFineRegressor($\mathcal{S}_A$, $\mathcal{F}_A$, $\mathcal{A}_{t-1}$, $K$, $P$)}
\vskip 2ex
\STATE {// Update validation samples}
\STATE {$\mathcal{V}_{t}= \mathcal{V}_{t-1} + \{\mathcal{C}^{\vv}_t(f_i)\}_{i=1}^{N_V}$}
\vskip 2ex
\STATE {// Increase P when $NME(\{\vx^{t}_i,\vx^{g}_i\}_{i=1}^{N_A}) < NME(\{\vx^{t}_i,\vx^{g}_i\}_{i=1}^{N_V})$}
\STATE {// Compute validation error improvement}
\STATE {$\Delta\varepsilon = NME(\{\vx^{t-1}_i,\vx^{g}_i\}_{i=1}^{N_V}) - NME(\{\vx^{t}_i,\vx^{g}_i\}_{i=1}^{N_V})$}
\UNTIL{t $>$ $T$ \textbf{or} $\Delta\varepsilon < 1\%$}
\ENSURE $\{\mathcal{C}^{\vv}_t\}_{t=1}^{T^*}$ // ${T^*}$ is the last trained stage
\end{algorithmic}
\end{algorithm}

Compared with conventional ERT approaches, our ensemble is simpler. It will require fewer trees because we only have to estimate the non-rigid face deformation, since the 3D rigid component has already been estimated in the previous step. In the following we describe the details.

\subsubsection{Initial shapes for regression} 
The selection of the starting point in the ERT is fundamental to reach a good solution. The simplest choice is the mean of the ground truth training shapes, $\bar{\vx}^0 = \sum_{i=1}^N{\vx_i^g}/N$. However, such a poor initialization leads to wrong alignment results in test images with large pose variations. Alternative strategies run the ERT several times with different initializations~\citep{Burgos13},  initialize with other ground truth shapes $\vx^0_i \leftarrow \vx_j^g$ where $i \neq j$~\citep{Kazemi14}, or randomly deform the initial shape~\citep{Kowalski17}.

\begin{figure}
\centering
\includegraphics[width=0.11\textwidth]{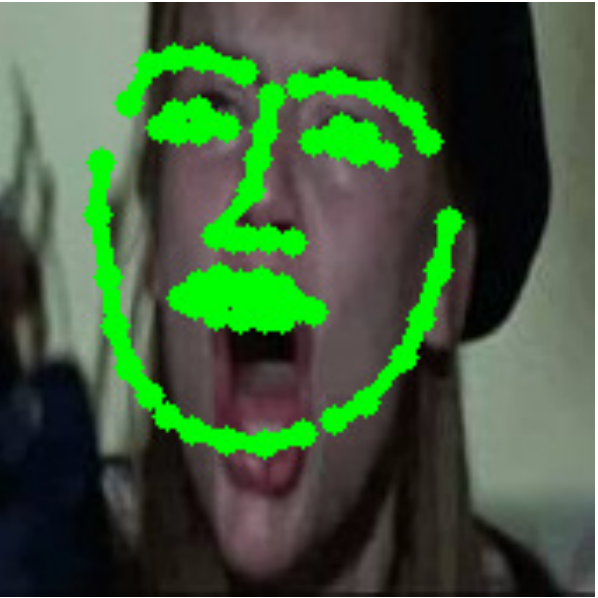}
\includegraphics[width=0.11\textwidth]{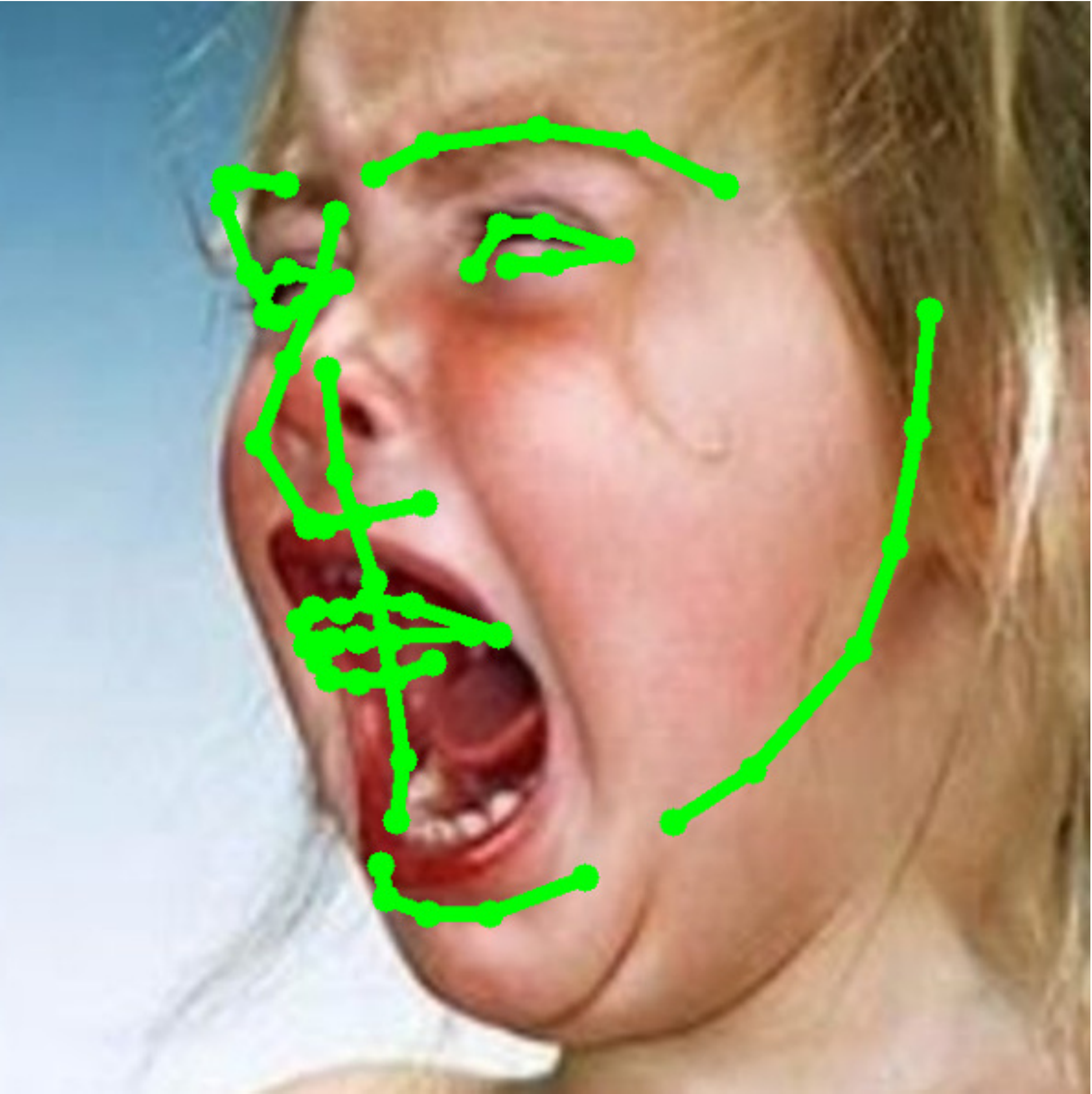}
\includegraphics[width=0.11\textwidth]{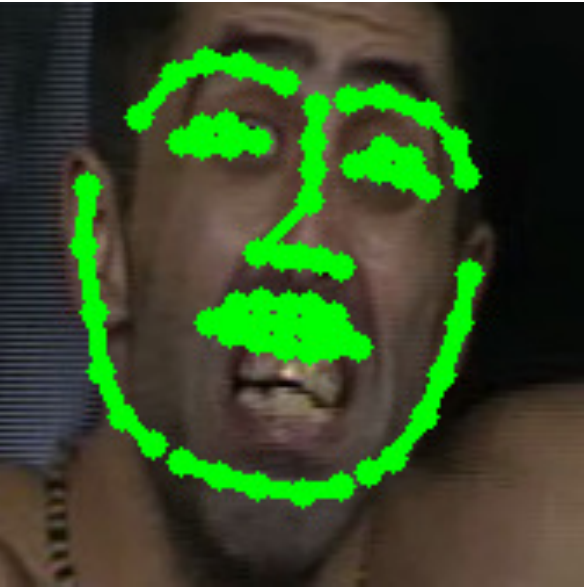}
\includegraphics[width=0.11\textwidth]{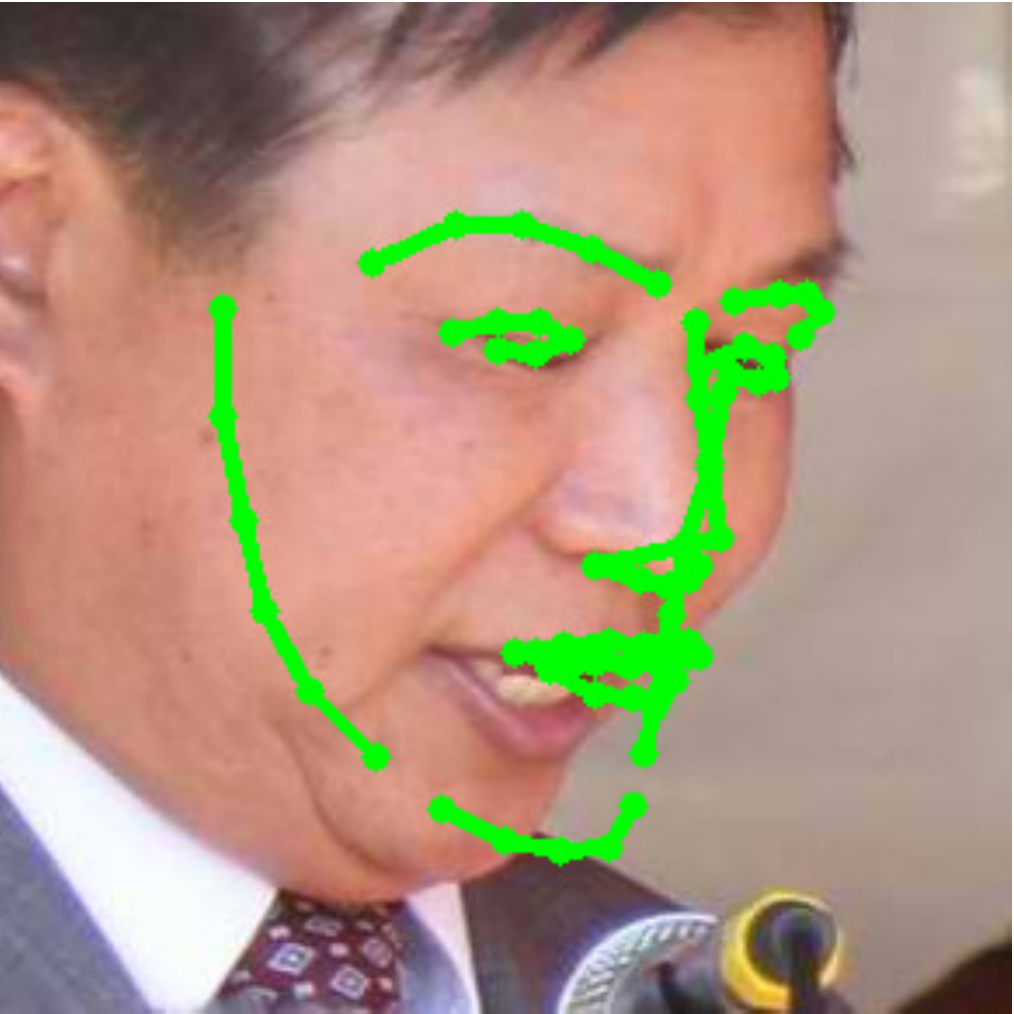}\\
\vspace{0.1cm}
\includegraphics[width=0.11\textwidth]{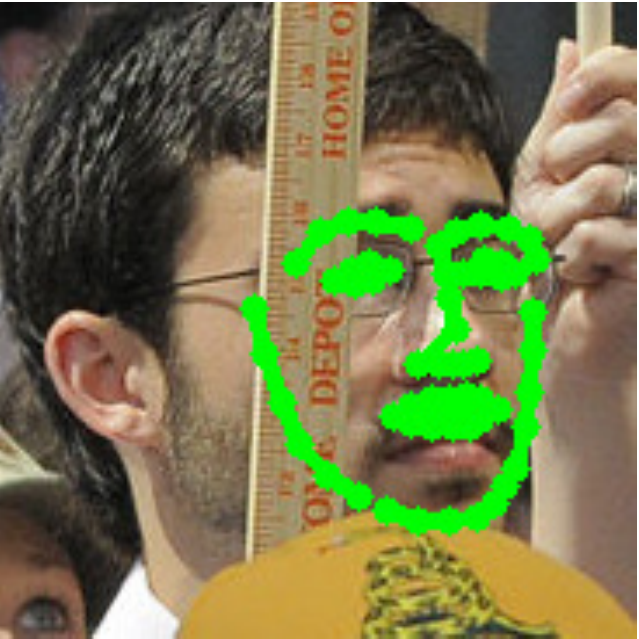}
\includegraphics[width=0.11\textwidth]{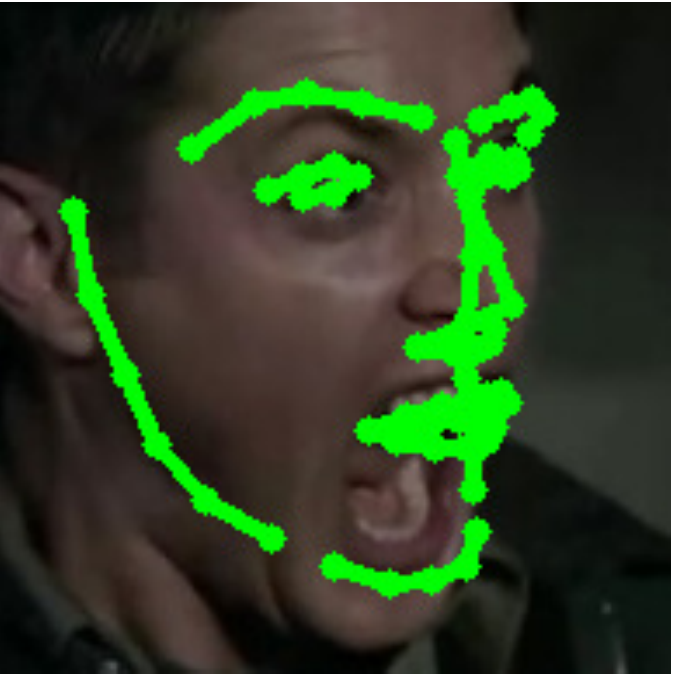}
\includegraphics[width=0.11\textwidth]{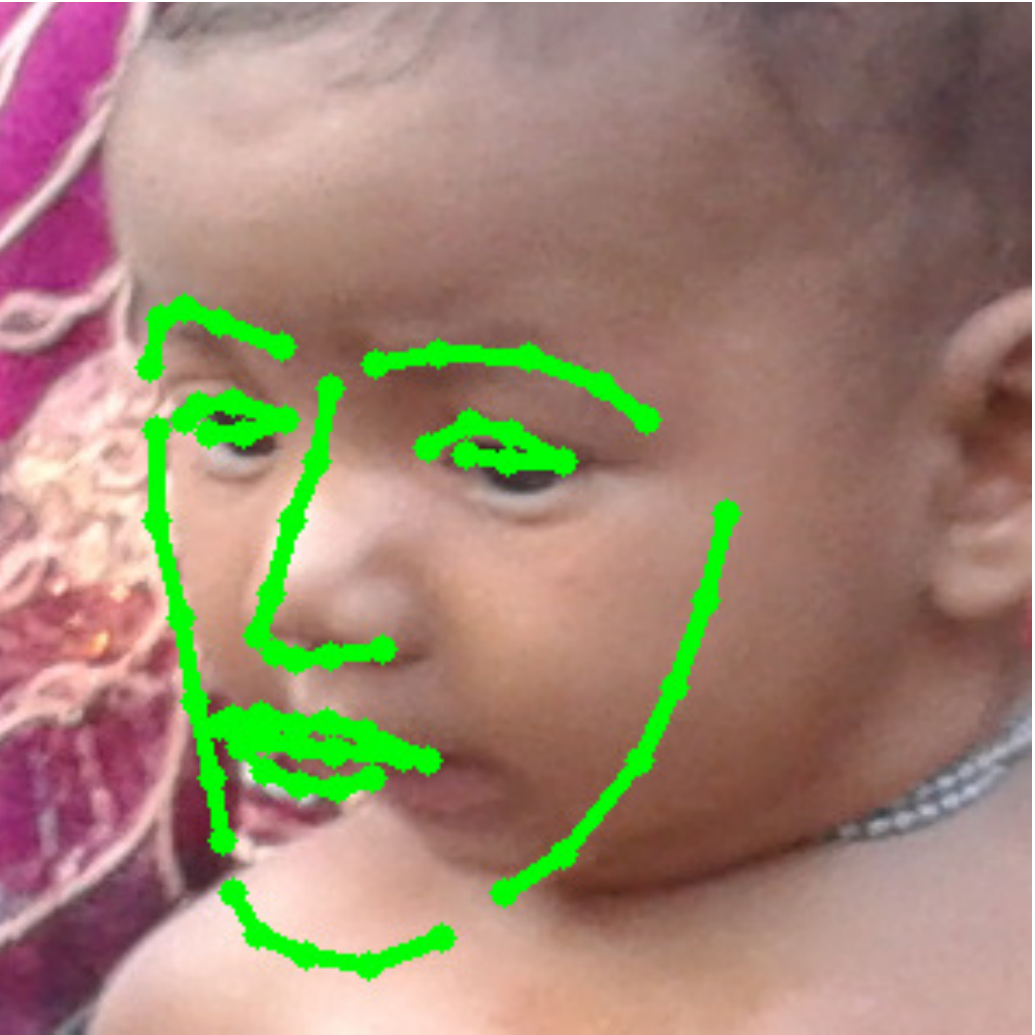}
\includegraphics[width=0.11\textwidth]{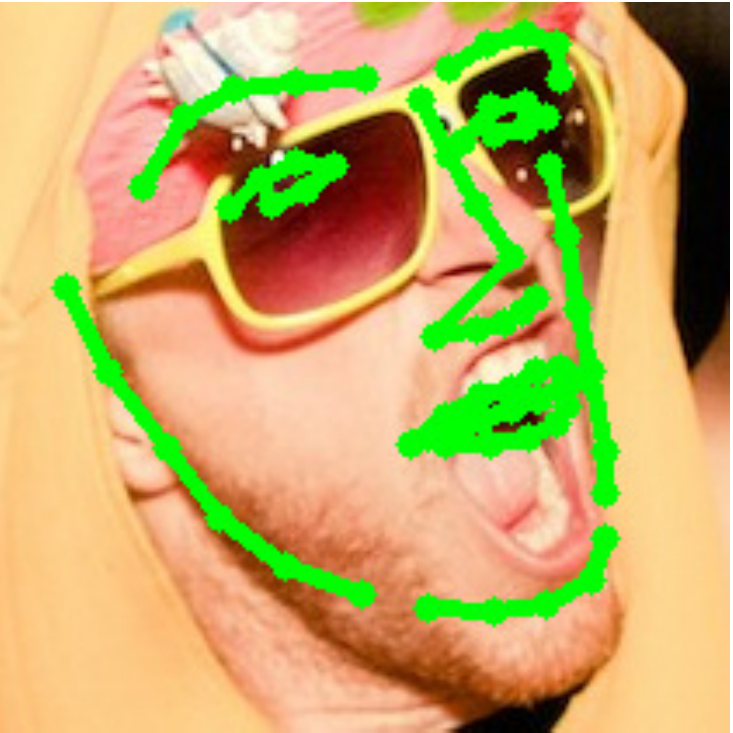}
\caption{The 8 worst initial shapes for the 300W private test set produced by $g_0$ (CNN+3D).}
\label{fig:initials}
\end{figure}

In our approach we initialize the ERT using the algorithm described in section~\ref{sec:rigid}, that  provides a robust pose and a valid shape for initialization (see Fig.~\ref{fig:initials}). Hence, the ERT only needs to estimate the non-rigid deformation component of the face.

\subsubsection{Feature Extraction}
ERT efficiency depends on the feature extraction step. In general, descriptor features such as SIFT used by~\cite{Xiong13} and~\cite{Zhu15} improve face alignment results, but have higher computational cost compared to simpler features such as plain pixel value differences~\citep{Cao14,Burgos13,Kazemi14,Ren16}. In our case, a simple feature suffices, since shape landmarks are close to their ground truth location.

We use the probability maps $\mathcal{P}(\mI)$ to extract features for the cascade. To this end, we select a landmark $l$ and its associated probability map $\mathcal{P}^l(\mI)$. The feature is computed as the difference between two pixels values in $\mathcal{P}^l(\mI)$ from a FREAK descriptor pattern~\citep{Alahi12} around $l$, similar to those in \cite{Lee15b}. However, ours are defined on the probability maps, $\mathcal{P}(\mI)$, instead of the image, $\mI$. We let the training algorithm select the most informative landmark and pair of pixels in each iteration.

\subsubsection{Learn a coarse-to-fine regressor} 
To train the $t$-th stage regressor, $\mathcal{C}^{\vv}_t$, we fit an ERT. Thus, the goal is to sequentially learn a series of weak learners to greedily minimize the regression loss function:
\begin{equation}
   \footnotesize
   \label{eq:loss_regression}
   \mathcal{L}_t(\mathcal{S}_A, \mathcal{F}_A, \mathcal{A}_{t-1}) = \sum_{i=1}^{N_A} 
   || \vw_i^g\odot (\vx_i^g - \vx_i^{t-1} - \sum_{k=1}^{K} g_k(f_i)) ||^2
\end{equation}
where $\odot$ is the Hadamard product. There are different ways of minimizing Equation~\ref{eq:loss_regression}. \cite{Kazemi14} present a general framework based on Gradient Boosting for learning an ensemble of regression trees. \cite{Lee15b} establish an optimization method based on Gaussian Processes also learning an ensemble of regression trees but outperforming previous literature by reducing the overfitting.  In our approach we adopt a Gradient Boosting scheme (see Algorithm~\ref{alg:p_parts_regressors}).

A crucial problem when training a global face landmark regressor is the lack of examples showing all possible combinations of face parts deformations. Hence, these regressors quickly overfit and generalize poorly to combinations of part deformations not present in the training set.
To address this problem we introduce the coarse-to-fine ERT architecture.

\begin{figure}
\centering
\includegraphics[width=0.11\textwidth]{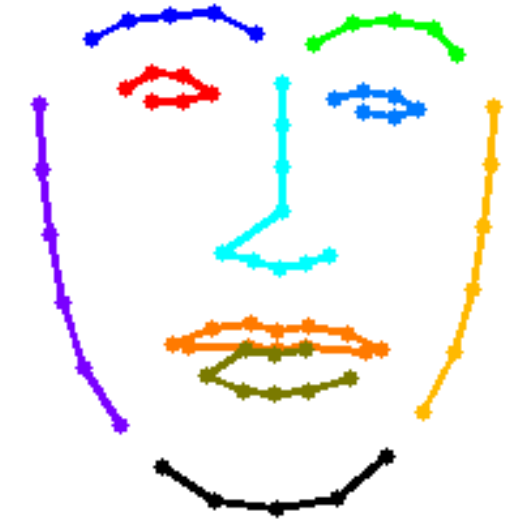}
\includegraphics[width=0.11\textwidth]{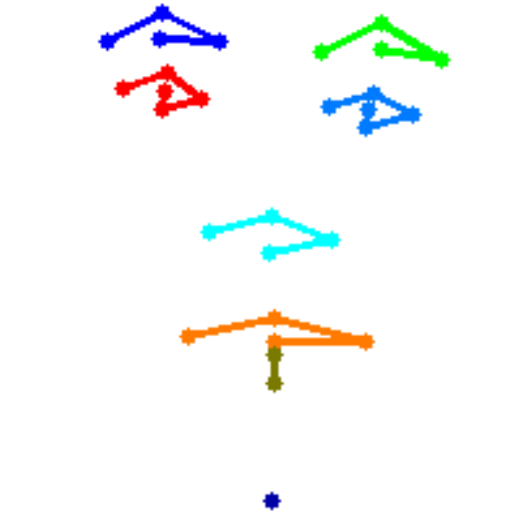}
\includegraphics[width=0.11\textwidth]{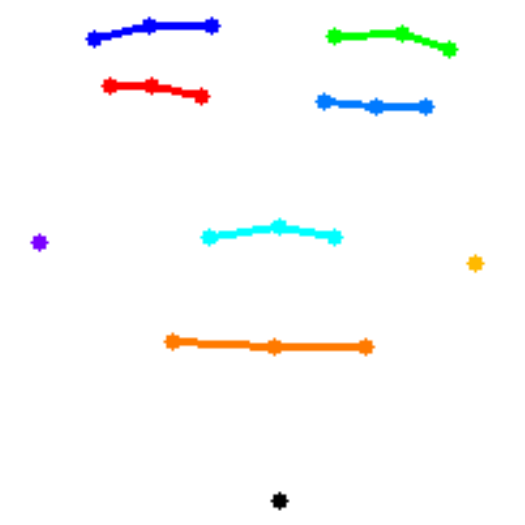}
\includegraphics[width=0.11\textwidth]{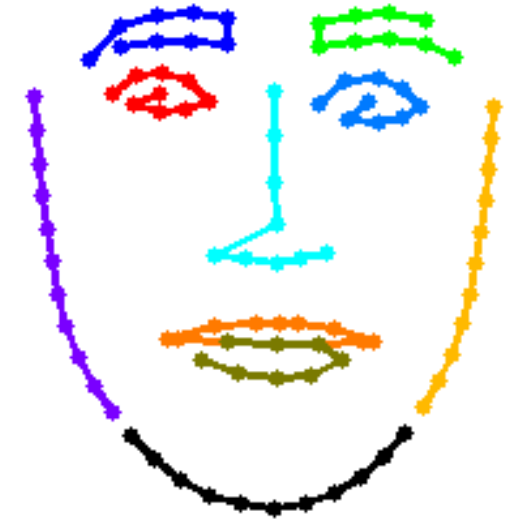}
\caption{The $P=10$ face parts of 300W, COFW, AFLW and WFLW data bases in the fine stage of our coarse-to-fine ERT.}
\label{fig:parts_configuration}
\end{figure}

The goal is to be able to cope with combinations of face part deformations not seen during training. A single monolithic regressor is not able to estimate these local deformations (see the difference between monolithic and coarse-to-fine NME curves in Fig.~\ref{fig:learning:a}). Our algorithm is agnostic in the number of parts and stages of the coarse-to-fine estimation. Algorithm~\ref{alg:p_parts_regressors} details the training of $P$ face parts regressors (each one with a subset of the landmarks) to build a coarse-to-fine regressor. Note that $\mathcal{A}_{k-1}$ in this context is the shape and visibility vectors from the last regressor output (\eg the previous part regressor or a previous full stage regressor). In our implementation the coarse-to-fine scheme has two stages. The coarse stage has one part, $P=1$, that involves all landmarks and $K_1$ trees. The fine stage has ten parts, $P=10$, left/right eyebrow, left/right eye, nose, top/bottom mouth, left/right ear and chin (see Fig.~\ref{fig:parts_configuration}), with $K_2$ trees.

\begin{algorithm}
\footnotesize
\caption{Training $P$ parts regressors}
\label{alg:p_parts_regressors}
\begin{algorithmic}
\renewcommand{\algorithmicrequire}{\textbf{Input:}}
\renewcommand{\algorithmicensure}{\textbf{Output:}}
\REQUIRE $\mathcal{S}_A$, $\mathcal{F}_A, \mathcal{A}_{t-1}, \nu, K, P$
\FOR{k=1 \TO $K$} 
\FOR{p=1 \TO $P$} 
\STATE {// Compute residuals: }
\STATE {//   $\odot$ is the Hadamard product}
\STATE {//   $(p)$ selects elements of vectors in that part}
\STATE {$\{\vr^k_i(p) = \vw_i^g(p) \odot (\vx_i^g(p) - \vx^{k-1}_i(p))\}_{i=1}^{N_A}$}
\vskip 2ex
\STATE {$g_k^p$ = fitRegressionTree($\{\vr^k_i(p)\}_{i=1}^{N_A}$,$\mathcal{F}_A(p)$)}
\vskip 2ex
\STATE {// Update samples with the regression tree estimation,}
\STATE {// $\nu$, shrinkage factor to scale each tree contribution}
\STATE {$\mathcal{A}_{k}(p)= \mathcal{A}_{k-1}(p) + \nu \cdot \{g_k^p(f_i(p))\}_{i=1}^{N_A}$}
\ENDFOR
\ENDFOR
\ENSURE $\{\mathcal{C}^p\}_{p=1}^P$, being $\mathcal{C}^p=\{g_k^p\}_{k=1}^K$
\end{algorithmic}
\end{algorithm}

\subsubsection{Fit a regression tree} 
The training objective for the $k$-th regression tree is to minimize the sum of squared residuals, taking into account the annotated landmark labels:
\begin{equation}
   \label{eq:loss_tree}
   \mathcal{E}_k = \sum_{i=1}^{N_A} ||\vr^k_i||^2 = \sum_{i=1}^{N_A} ||\vw_i^g \odot (\vx_i^g - \vx^{k-1}_i)||^2
\end{equation}
We learn each regression binary tree by recursively splitting the training set into the left (l) and right (r) child nodes. The tree node split function is designed to minimize $\mathcal{E}_k$ from Equation~\ref{eq:loss_tree} in the selected landmark. To train a regression tree node we randomly generate a set of candidate split functions, each of them involving four parameters $\theta = (\tau, \vp_1, \vp_2, l)$, where $\vp_1$ and $\vp_2$ are pixels coordinates on a fixed FREAK structure around the $l$-th landmark coordinates in $\vx_i^{k-1}$. The feature value corresponding to $\theta$ for the $i$-th training sample is $f_i(\theta) = \mathcal{P}^l(\mI_i)[\vp_1] - \mathcal{P}^l(\mI_i)[\vp_2]$, the difference of probability values in the maps for the given landmark.
Finally, we compute the split function thresholding the feature value, $f_i(\theta) > \tau$.

Given $\mathcal{N} \subset \mathcal{S}_A$ the set of training samples at a node, fitting a tree node for the $k$-th tree, consists of finding the parameter $\theta$ that minimizes $E_k(\mathcal{N},\theta)$
\begin{equation}
   \footnotesize
   \arg\min_{\theta} E_k(\mathcal{N},\theta) = \arg\min_{\theta} \sum_{b\in{\{l,r\}}}{\sum_{s\in{\mathcal{N}_{\theta,b}}}{ ||\vr_s^k-\vmu_{\theta,b}||^2}} 
    \label{eq:entropy}
\end{equation}
where $\mathcal{N}_{\theta,l}$ and $\mathcal{N}_{\theta,r}$ are, respectively, the samples sent to the left and right child nodes due to the decision induced by $\theta$. The mean residual $\vmu_{\theta,b}$ for a candidate split function and a subset of training data is given by
\begin{equation}
\vmu_{\theta,b} = \frac{1}{|\mathcal{N}_{\theta,b}|}\sum_{s\in{\mathcal{N}_{\theta,b}}} \vr_s^k
\label{eq:mean}
\end{equation}

Once we know the optimal split each leaf node stores the mean residual, $\vmu_{\theta,b}$, as the output of the regression for any example reaching that leaf. We also output the mean visibility of the samples reaching the tree leaf.


\section{Experiments}\label{sec:experiments}
To train and evaluate our proposal, we perform experiments with 300W, COFW, AFLW and WFLW that are considered the most challenging public data sets:
\begin{itemize}
\item \textbf{300W}. It provides 68 manually annotated landmarks, \cite{Sagonas16}. We follow the most established approach and divide the 300W annotations into 3148 training and 689 testing images (public competition). Evaluation is also performed on the 300W private competition using the previous 3837 images as training and 600 newly updated images as testing set.
\item \textbf{COFW}. This benchmark, presented in \cite{Burgos13} focuses on occlusion. Commonly, there are 1345 training faces in total. The testing set is made of 507 images. The annotations include the landmark positions and the binary occlusion labels for 29 points.
\item \textbf{AFLW}. It provides a collection of 25993 in-the-wild faces, with 21 facial landmarks annotated depending on their visibility, \cite{Koestinger11}. We have found several annotations errors and, consequently, removed these faces from our experiments. From the remaining faces we randomly choose 19312 images for training/validation and 4828 instances for testing. 
\item \textbf{WFLW}. It consists of 7500 extremely challenging training and 2500 testing faces divided into six subgroups, pose, expression, illumination, make-up, occlusion and blur,
with 98 fully manual annotated landmarks, \cite{Wu18}.
\end{itemize}

\subsection{Evaluation}
We use the Normalized Mean Error (NME) as a metric to measure the shape estimation error
\begin{equation}
        \footnotesize
        NME = \frac{100}{N} \sum\limits_{i=1}^{N} \left( \frac{1}{||\vw_i^g||_1} \sum\limits_{l=1}^{L} \left( \frac{\vw_i^g(l)\cdot{} \left\|{\vx_i(l)-\vx_i^g(l)}\right\|}{d_i} \right) \right)
    \label{eq:nme}
\end{equation}
It computes the mean euclidean distance between the ground-truth and estimated landmark positions normalized by $d_i$. We report our results using different values of $d_i$: the ground truth distance between the eye centers (\emph{pupils}), the ground truth distance between the outer eye corners (\emph{corners}) and the ground truth bounding box size (\emph{height}). 

In addition, we also compare our results using Cumulative Error Distribution (CED) curves. We calculate $AUC_\varepsilon$ as the area under the CED curve for images with an NME smaller than $\varepsilon$ and $FR_\varepsilon$ as the failure rate representing the percentage of testing faces with NME greater than $\varepsilon$. We use precision/recall percentages to compare occlusion prediction.

To train our algorithm we shuffle the training set of each data base and split it into 90\% train-set and 10\% validation-set.

\subsection{Implementation}
All experiments have been carried out with the settings described in this section. For each data set, we train from scratch the CNN selecting the model parameters with lowest validation error. We crop faces using the ground truth  bounding boxes annotations enlarged by 30\%. We generate different training samples in each epoch by applying random in plane rotations between $\pm45^\circ$, scale changes by $\pm15\%$ and translations by $\pm5\%$ of bounding box size, randomly mirroring images horizontally and generating random rectangular occlusions. We use Adam stochastic optimization with $\beta_1=0.9$, $\beta_2=0.999$ and $\epsilon=1e^{-8}$ parameters. We train until convergence with an initial learning rate $\alpha=0.001$. When validation error levels out for 10 epochs, we multiply the learning rate by $decay=0.05$. In the CNN the cropped input face is reduced from 160$\times$160 to 1$\times$1 pixels gradually dividing by half their size across $B=8$ branches applying a stride 2 convolution with kernel size 2$\times$2\footnote{5$\times$5 images are reduced to 2$\times$2 pixels applying a kernel size of 3$\times$3}. We apply batch normalization after each convolution. All layers contain 68 filters to describe the required landmark features. We apply a Gaussian filter with $\sigma=33$ to the output probability maps to stabilize the initialization, $g_0$.

We train the coarse-to-fine ERT with the Gradient Boosting algorithm~\citep{Hastie09}. It requires a maximum of $T=20$ stages of $K=50$ regression trees per stage. The depth of trees is set to 4. The number of tests to choose the best split parameters, $\theta$, is set to 200. We resize each image to set the face size to 160$\times$160 pixels. For feature extraction, the FREAK pattern diameter is reduced gradually in each stage (\ie in the last stages the pixel pairs for each feature are closer). We generate $Z=25$ initializations in the robust softPOSIT scheme of $g_0$. We augment the shapes of each face training image to create a set, $\mathcal{S}_A$, of at least $N_A=60000$ samples to train the cascade. To avoid overfitting we use a shrinkage factor $\nu=0.1$ and subsampling factor $\eta=0.5$ in the ERT. Our regressor triggers the coarse-to-fine strategy once the training error is below the validation error, \eg $t=5$ in Fig.~\ref{fig:learning:a}.

Training the CNN and the coarse-to-fine ensemble of trees takes 48 hours using a NVidia GeForce GTX 1080Ti (11GB) GPU and an dual Intel Xeon Silver 4114 CPU at 2.20GHz (2$\times{}$10 cores/20 threads, 128 GB of RAM) with a batch size of 32 images. At runtime our method process test images on average at a rate of 12.5 FPS, where the CNN takes 75 ms and the ERT 5 ms per face image using C++, Tensorflow and OpenCV libraries.

\subsection{Experiments using public code}
\begin{figure*}
  \centering
    \includegraphics[width=0.45\textwidth]{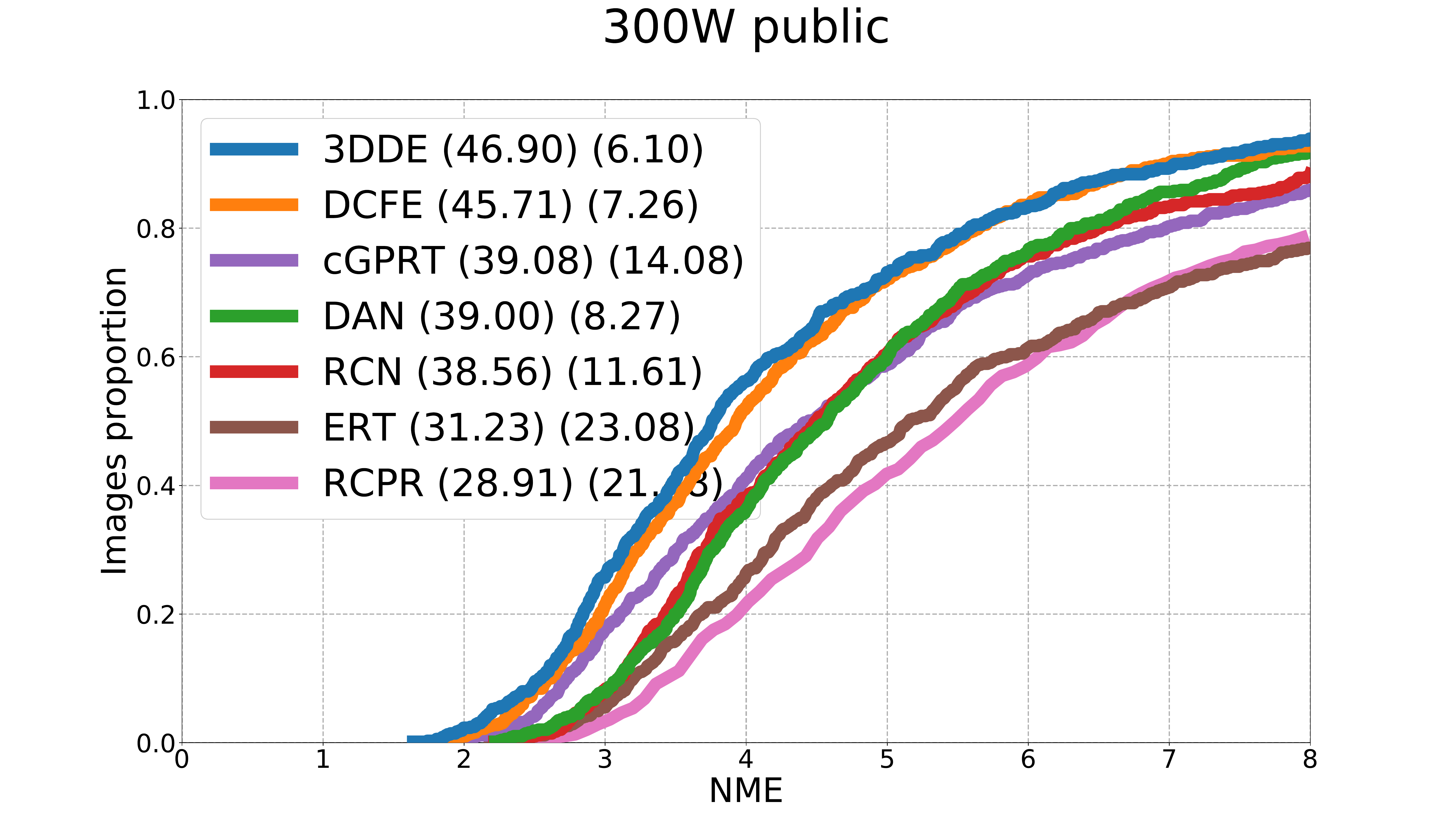}
    \includegraphics[width=0.45\textwidth]{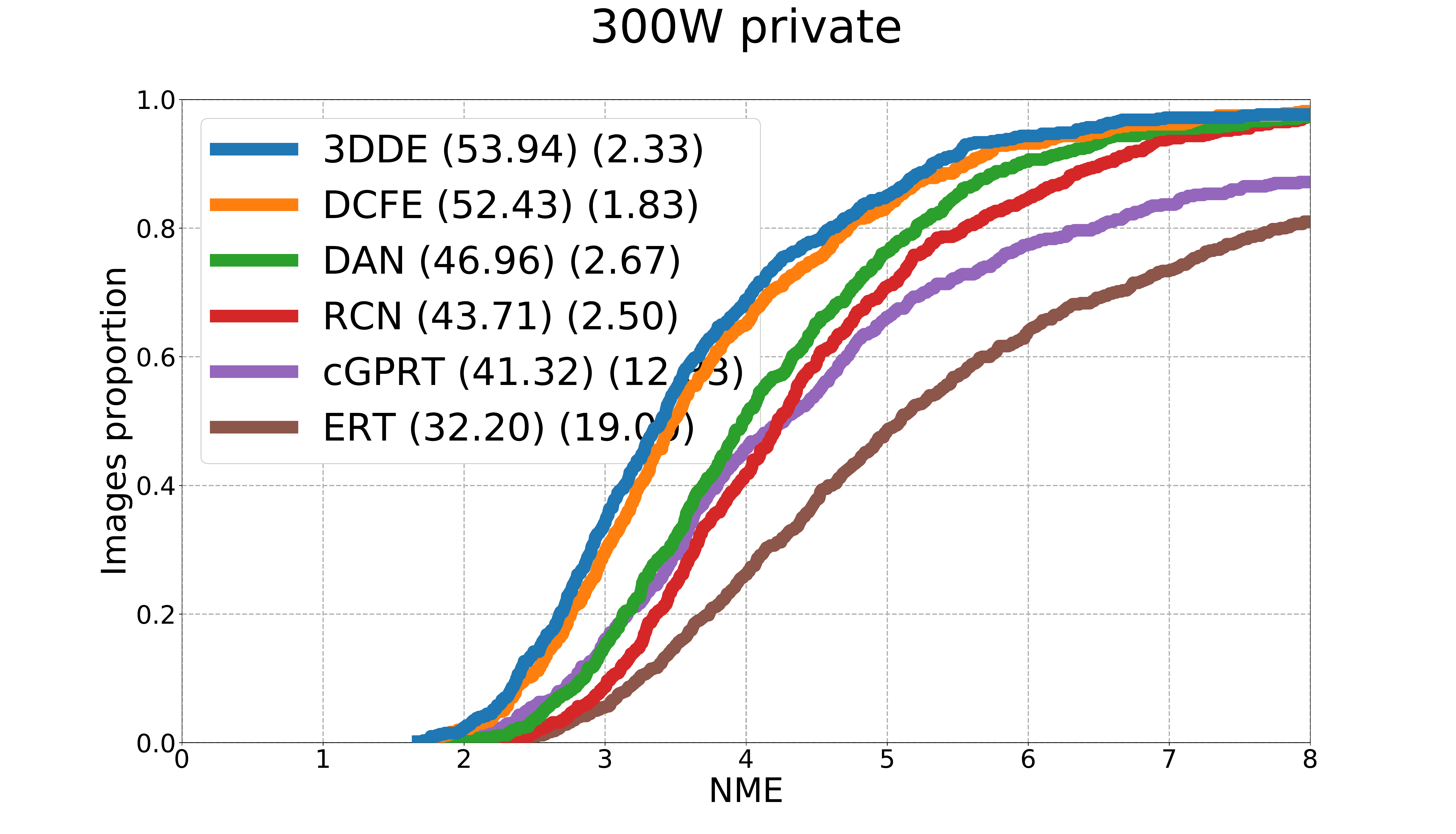}\\
    \includegraphics[width=0.45\textwidth]{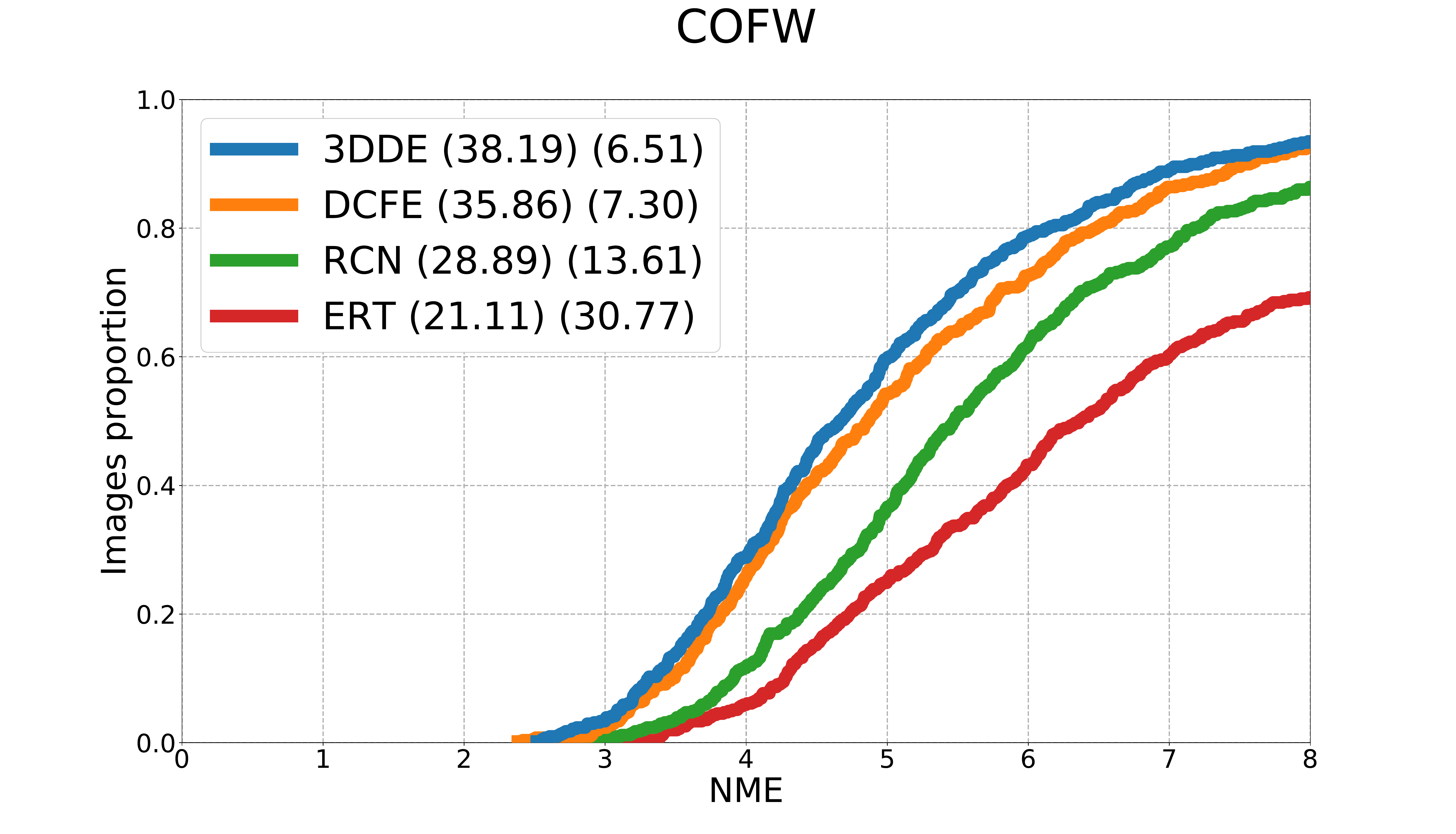}
    \includegraphics[width=0.45\textwidth]{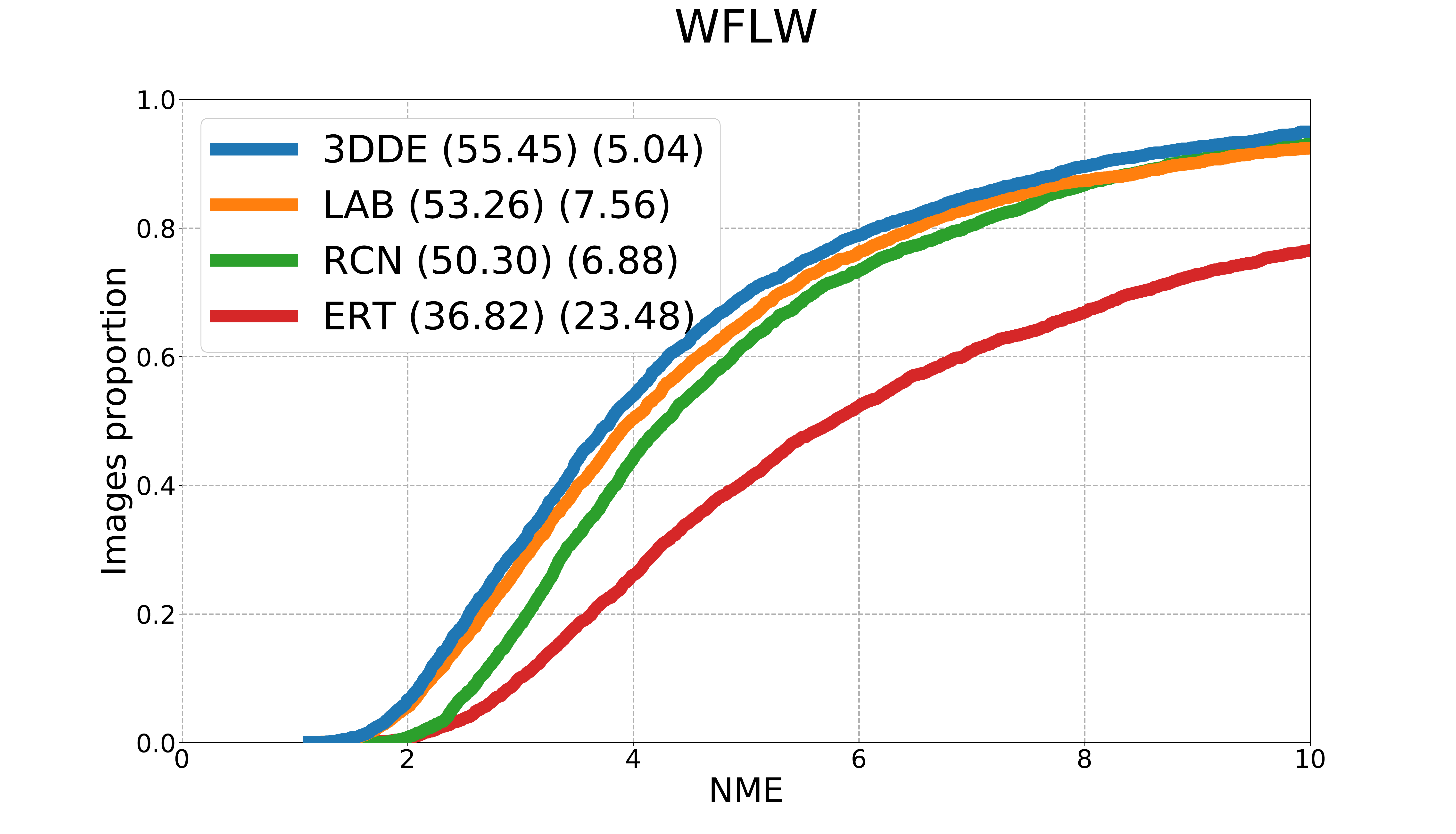}
  \caption{Cumulative error distribution curves sorted by AUC.}
  \label{fig:auc}
\end{figure*}

Published results in the literature are sometimes not fully comparable. In this section we use publicly available code to ensure a fair comparison between 3DDE and DCFE~\citep{Valle18}, LAB~\citep{Wu18}, DAN~\citep{Kowalski17}, RCN~\citep{Honari16}, cGPRT~\citep{Lee15b}, RCPR~\citep{Burgos13} and ERT~\citep{Kazemi14} with the same settings (including same training, validation and bounding boxes), in different benchmarks: 300W public, 300W private, COFW and WFLW. Note that LAB~\citep{Wu18} only provides a trained model for the WFLW data set. In addition, DAN~\citep{Kowalski17} provides code using 68 landmarks, for this reason we only report results in 300W. In Fig.~\ref{fig:auc} we plot the CED curves for all data bases. In the legend we provide the $AUC$ and $FR$ values for each algorithm.

The selected algorithms are representative of the three main families of solutions: a) ensembles of regression trees (cGPRT, RCPR, ERT), b) CNN-based approaches (LAB, DAN, RCN) and c) mixed approaches with deep nets and ensembles of regression trees (3DDE, DCFE). Overall, 3DDE is better than any other providing a public implementation in the literature. We improve over our preliminary algorithm, DCFE~\cite{Valle18}, because of the better 3D initialization and regularization (see a complete analysis in section~\ref{sec:ablation}). In general we are able to improve by a large margin other ERT methods as RCPR, ERT or cGPRT because of the better initialization and the robust features provided by the CNN. We also outperform RCN (without any denoising model), a CNN architecture like the one used in 3DDE. Even DAN and LAB, that implement a cascade of CNN regressors, can not compete with the regularization obtained by using the cascade of ERT in 3DDE (see Fig.~\ref{fig:auc}). The fact that the largest margin is in COFW reflects the importance of the implicit shape model in our cascade to address occlusions.

\subsection{Experiments using published results}
In this section we compare 3DDE with other methods in the literature by using their published results. Since our method is able to train with unannotated landmarks and visibilities, we are able to train and evaluate all data sets in the literature.

First we test our method against the 300W benchmark. Our approach obtains the best overall performance in the indoor and outdoor subsets of the private competition (see Table~\ref{table:300w_private}) and in the full subset of the 300W public test set (see Table~\ref{table:300w_public}). This is due to the excellent accuracy achieved by the coarse-to-fine ERT scheme enforcing valid face shapes and the deep robust features extracted from the CNN. In the challenging subset of the 300W public competition, SHN~\citep{Yang17} gets better results than 3DDE. This is due to 3DDE failing to estimate good landmark probability maps for images with large scale variations. Our method exhibits superior capability in handling typical cases in the data base, since we achieve the best NME full set results in 300W public, 4.39, and in 300W private, 3.73.

\begin{table*}
\begin{center}
\footnotesize
\setlength\tabcolsep{0.25pt}
\begin{tabular}{l|c|c|c|c|c|ccc}
\hline
\multirow{3}{*}{Method} & \multicolumn{2}{c|}{Common} & \multicolumn{2}{c|}{Challenging} & \multicolumn{4}{c}{Full}\\
 & pupils & corners & pupils & corners & pupils & \multicolumn{3}{c}{corners}\\
 & $NME$ & $NME$ & $NME$ & $NME$ & $NME$ & $NME$ & $AUC_8$ & $FR_8$\\
\hline
RCPR \citep{Burgos13} & 6.18 & - & 17.26 & - & 8.35 & - & - & -\\
ESR \citep{Cao12} & 5.28 & - & 17.00 & - & 7.58 & - & 43.12 & 10.45\\
SDM \citep{Xiong13} & 5.60 & - & 15.40 & - & 7.52 & - & 42.94 & 10.89\\
ECSAN \citep{Zhang18b} & 5.42 & - & 11.80 & - & 6.67 & - & - & -\\
ERT \citep{Kazemi14} & - & - & - & - & 6.40 & - & - & -\\
LBF \citep{Ren16} & 4.95 & - & 11.98 & - & 6.32 & - & - & -\\
cGPRT \citep{Lee15b} & - & - & - & - & 5.71 & - & - & -\\
CFSS \citep{Zhu15} & 4.73 & - & 9.98 & - & 5.76 & - & 49.87 & 5.08\\
DDN \citep{Yu16} & - & - & - & - & 5.65 & - & - & -\\
TCDCN \citep{Zhang14b} & 4.80 & - & 8.60 & - & 5.54 & - & - & -\\
MDM \citep{Trigeorgis16} & - & - & - & - & - & - & 52.12 & 4.21\\
3DDFA \citep{Zhu17} & 5.09 & - & 8.07 & - & 5.63 & - & - & -\\
RCN \citep{Honari16} & 4.67 & - & 8.44 & - & 5.41 & - & - & -\\
DAN \citep{Kowalski17} & 4.42 & 3.19 & 7.57 & 5.24 & 5.03 & 3.59 & 55.33 & \textbf{1.16}\\
TSR \citep{Lv17} & 4.36 & - & 7.56 & - & 4.99 & - & - & -\\
RAR \citep{Xiao16} & 4.12 & - & 8.35 & - & 4.94 & - & - & -\\
SHN \citep{Yang17} & 4.12 & - & \textbf{7.00} & \textbf{4.90} & 4.68 & - & - & -\\
DCFE \citep{Valle18} & 3.83 & 2.76 & 7.54 & 5.22 & 4.55 & 3.24 & 60.13 & 1.59\\
PCD-CNN \citep{Kumar18a} & \textbf{3.67} & - & 7.62 & - & 4.44 & - & - & -\\
\hline
3DDE & 3.73 & \textbf{2.69} & 7.10 & 4.92 & \textbf{4.39} & \textbf{3.13} & \textbf{61.24} & 1.30\\
\hline
\end{tabular}
\end{center}
\caption{Error of face alignment methods on the 300W public test set.}
\label{table:300w_public}
\end{table*}

We may assess the improvement achieved by the 3D initialization and the coarse-to-fine ERT by comparing the results of 3DDE in the full subset of 300W, 4.39, with Honari's RCN using the denoising model~\citep{Honari16}, 5.41. It roughly represents a 19\% improvement in the inter-pupils NME. 

\begin{table*}
\footnotesize
\begin{center}
\setlength\tabcolsep{0.25pt}
\begin{tabular}{l|ccc|ccc|ccc}
\hline
\multirow{3}{*}{Method} & \multicolumn{3}{c|}{Indoor} & \multicolumn{3}{c|}{Outdoor} & \multicolumn{3}{c}{Full}\\
 & \multicolumn{3}{c|}{corners} & \multicolumn{3}{c|}{corners} & \multicolumn{3}{c}{corners}\\
 & $NME$ & $AUC_8$ & $FR_8$ & $NME$ & $AUC_8$ & $FR_8$ & $NME$ & $AUC_8$ & $FR_8$ \\
\hline
ESR~\citep{Cao12} & - & - & - & - & - & - & - & 32.35 & 17.00 \\
cGPRT~\citep{Lee15b} & - & - & - & - & - & - & - & 41.32 & 12.83 \\
CFSS~\citep{Zhu15} & - & - & - & - & - & - & - & 39.81 & 12.30 \\
MDM~\citep{Trigeorgis16} & - & - & - & - & - & - & 5.05 & 45.32 & 6.80 \\
DAN~\citep{Kowalski17} & - & - & - & - & - & - & 4.30 & 47.00 & 2.67 \\
SHN~\citep{Yang17} & 4.10 & - & - & 4.00 & - & - & 4.05 & - & -  \\
DCFE~\citep{Valle18} & 3.96 & 52.28 & 2.33 & 3.81 & 52.56 & \textbf{1.33} & 3.88 & 52.42 & \textbf{1.83} \\
\hline
3DDE & \textbf{3.74} & \textbf{53.93} & \textbf{2.00}& \textbf{3.71} & \textbf{53.95} & 2.66 & \textbf{3.73} & \textbf{53.94} & 2.33\\
\hline
\end{tabular}
\end{center}
\caption{Error of face alignment methods on the 300W private test set.}
\label{table:300w_private}
\end{table*}

Table~\ref{table:cofw} compares the performance of our model using the COFW data set. This is the standard to evaluate occlusions. 3DDE obtains the best results, NME 5.11, establishing a new state-of-the-art. This shows the importance of the face shape model implicit in the cascade of ERT to cope with severe occlusions. In terms of landmark visibility estimation, we have obtained better precision with an overall better recall than the best previous approach, DCFE. Again, the regularization together with the new initialization contributes to improve DCFE. 

In Table~\ref{table:aflw} we show the results of our evaluation with AFLW. This is a challenging data set not only because of its size and the large variability of face poses, but also because of the large number of samples with occluded landmarks, that are unannotated. 
Although the results in Table~\ref{table:aflw} are not strictly comparable, because each paper uses its own train and test subsets, we get an NME of 2.06 with the full 21 landmarks set. Again, it is a new state-of-the-art, since most competing approaches do not use the two most difficult landmarks, each located in one earlobe (see 19 landmarks results in Table~\ref{table:aflw}). We have also evaluated 3DDE without the two earlobe landmarks. In this case we get an NME of 2.01, the best reported result.

\begin{table*}
    \footnotesize
    \begin{center}
    \setlength\tabcolsep{1.25pt}
    \begin{tabular}{l|ccc|c}
    \hline
    \multirow{2}{*}{Method} & \multicolumn{3}{c|}{pupils} & occlusion\\
    & $NME$ & $AUC_8$ & $FR_8$ & precision/recall\\
    \hline
    RCPR~\citep{Burgos13} & 8.50 & - & - & 80/40\\
    TCDCN~\citep{Zhang14b} & 8.05 & - & - & -\\
    RAR~\citep{Xiao16} & 6.03 & - & - & -\\
    DAC-CSR~\citep{Feng17} & 6.03 & - & - & -\\
    Wu \etal~\citep{Wu15} & 5.93 & - & - & 80/49.11\\
    SHN~\citep{Yang17} & 5.6 & - & - & -\\
    PCD-CNN~\citep{Kumar18a} & 5.77 & - & - & -\\
    DCFE~\citep{Valle18} & 5.27 & 35.86 & 7.29 & 81.59/49.57\\
    \hline
    3DDE & \textbf{5.11} & \textbf{38.18} & \textbf{6.50} & \textbf{85.92/51.04}\\
    \hline
    \end{tabular}
    \end{center}
    \caption{Error of face alignment methods on COFW.}
    \label{table:cofw}
\end{table*}

\begin{table*}
    \footnotesize
    \begin{center}
    \setlength\tabcolsep{1.25pt}
    \begin{tabular}{l|c|c}
    \hline
    \multirow{3}{*}{Method} & 19 landmarks & 21 landmarks\\
    & height & height\\
    & $NME$ & $NME$\\
    \hline
    PIFAS~\citep{Jourabloo17} & - & 4.45\\
    CFSS~\citep{Zhu15} & 3.92 & -\\
    CCL~\citep{Zhu16a} & 2.72 & -\\
    DAC-CSR~\citep{Feng17} & 2.27 & -\\
    Binary-CNN~\citep{Bulat17} & - & 2.85\\
    PCD-CNN~\citep{Kumar18a} & - & 2.40\\
    TSR~\citep{Lv17} & 2.17 & -\\
    DCFE~\citep{Valle18} & 2.12 & 2.17\\
    \hline
    3DDE & \textbf{2.01} & \textbf{2.06}\\
    \hline
    \end{tabular}
    \end{center}
    \caption{Error of face alignment methods on AFLW.}
    \label{table:aflw}
\end{table*}

\begin{figure}
\centering
\stackunder[5pt]{\includegraphics[width=0.11\textwidth]{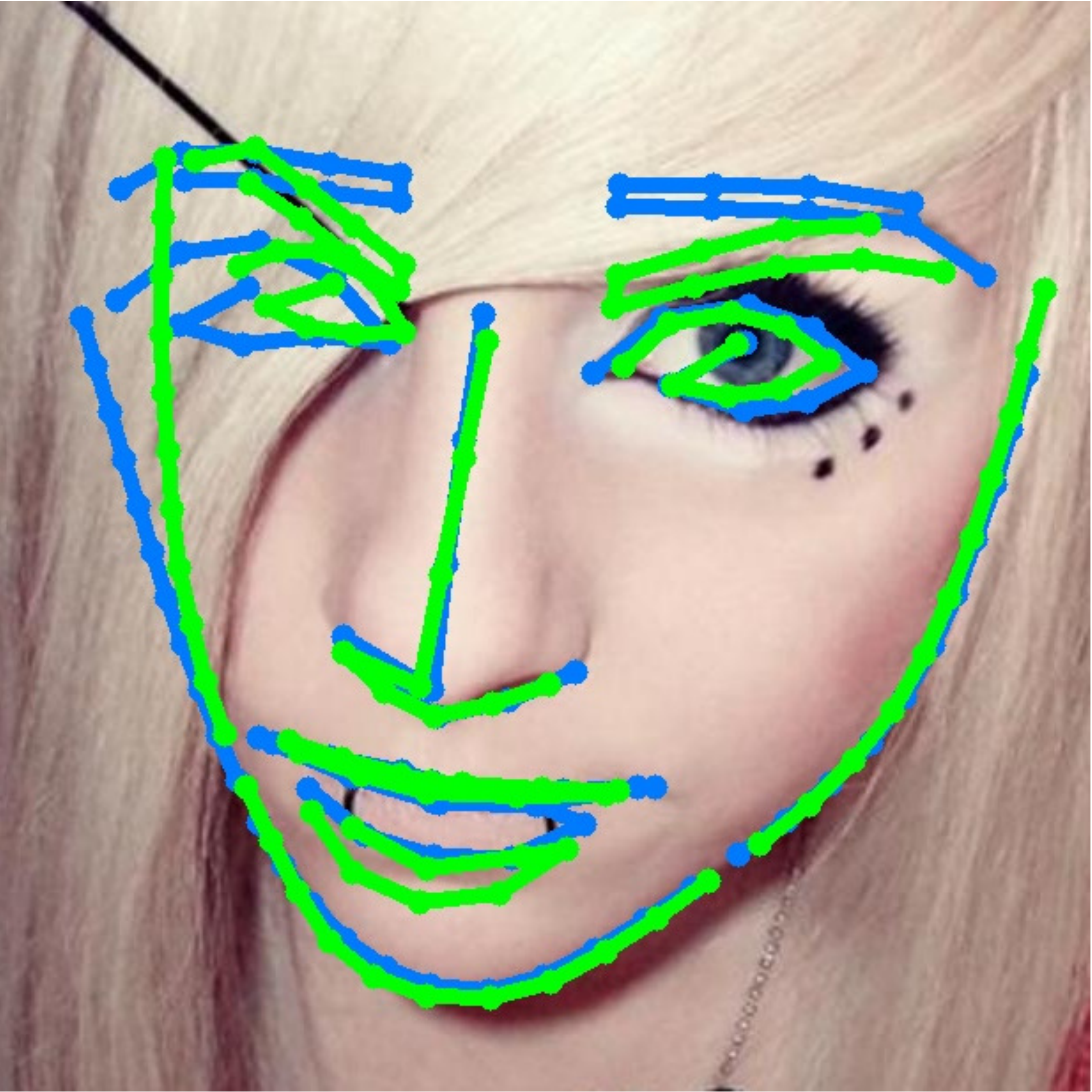}}{6.407}
\stackunder[5pt]{\includegraphics[width=0.11\textwidth]{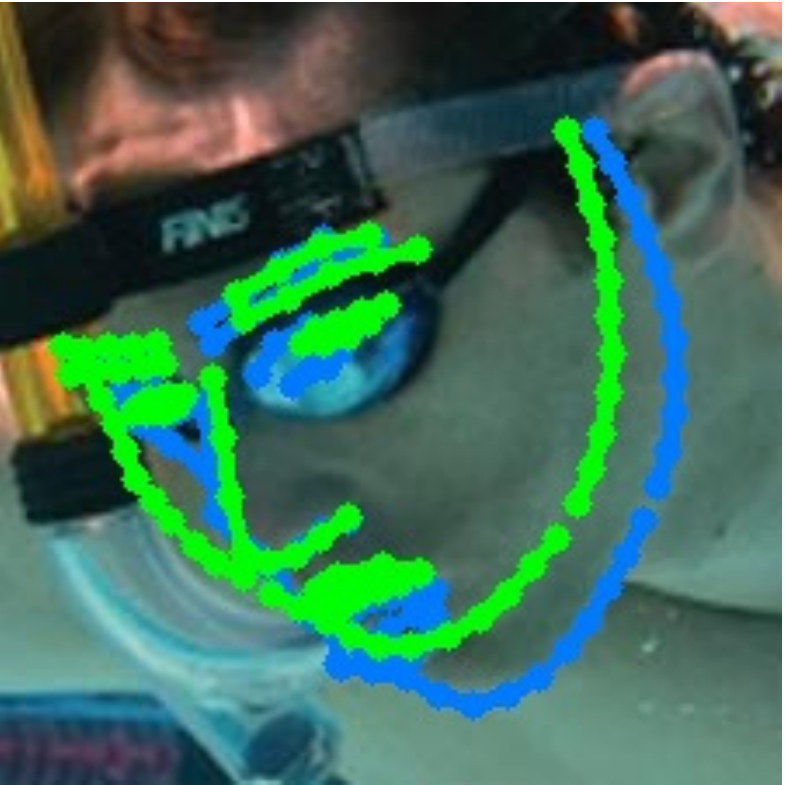}}{20.798}
\stackunder[5pt]{\includegraphics[width=0.11\textwidth]{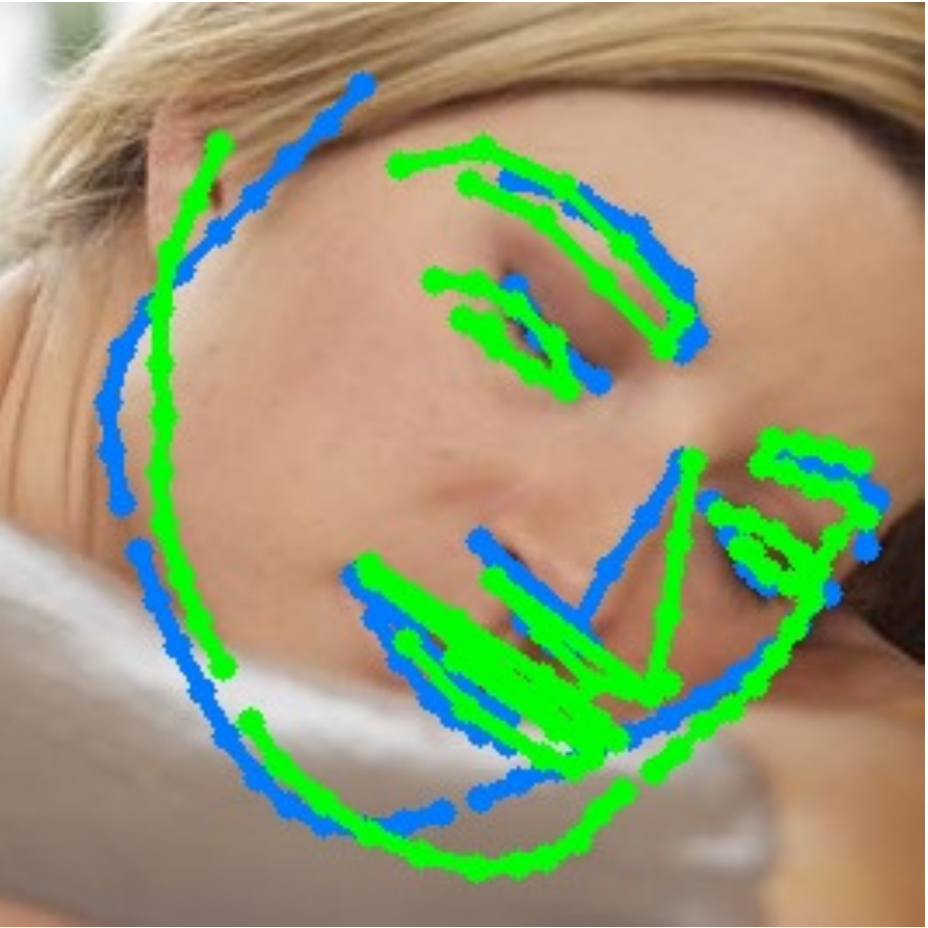}}{24.565}
\stackunder[5pt]{\includegraphics[width=0.11\textwidth]{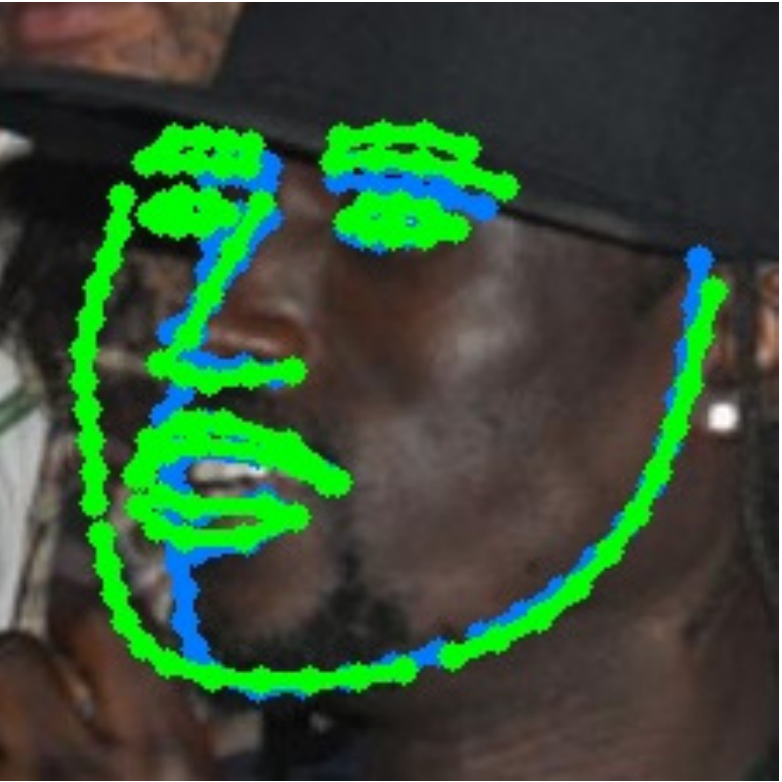}}{16.859}\\
\vspace{0.1cm}
\stackunder[5pt]{\includegraphics[width=0.11\textwidth]{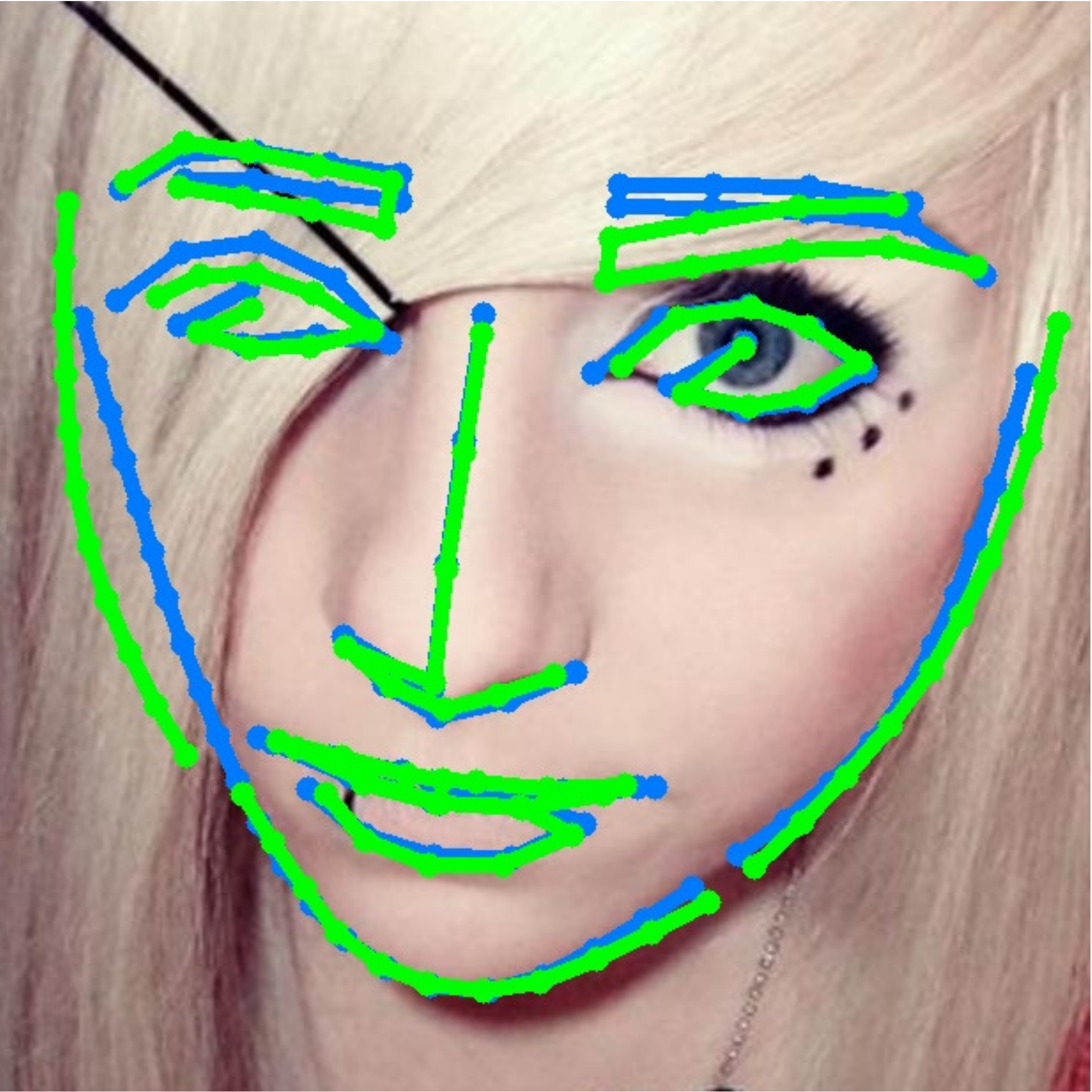}}{3.829}
\stackunder[5pt]{\includegraphics[width=0.11\textwidth]{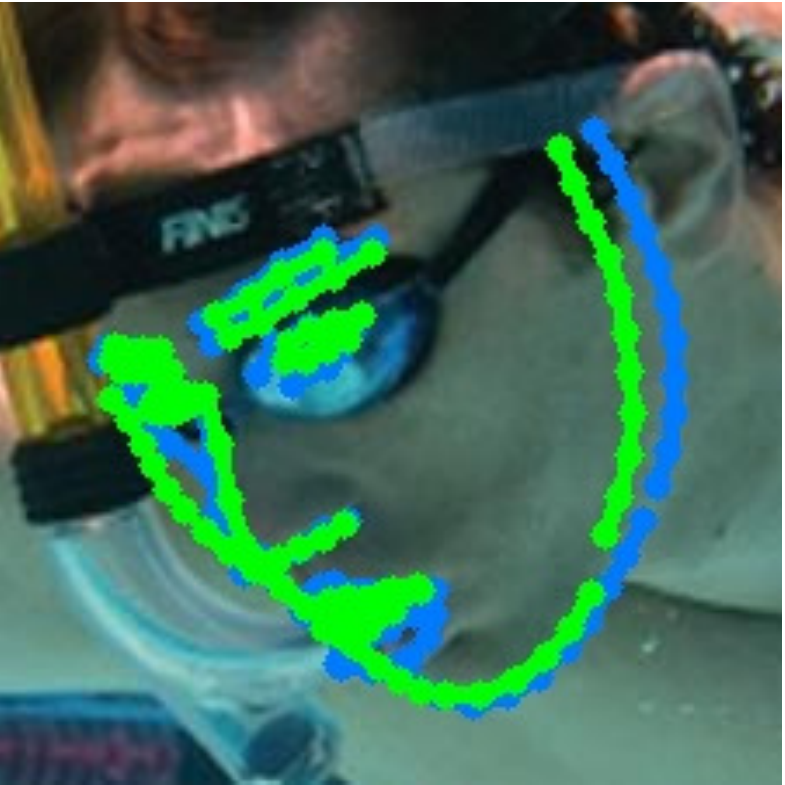}}{13.105}
\stackunder[5pt]{\includegraphics[width=0.11\textwidth]{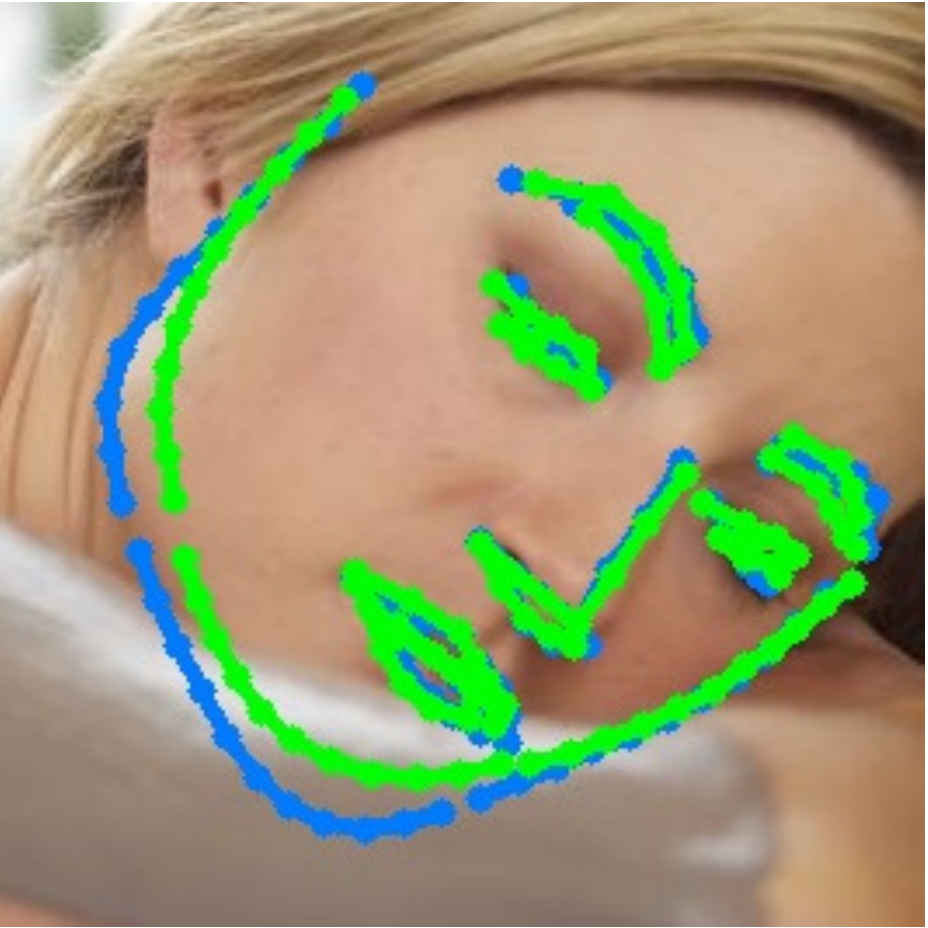}}{6.719}
\stackunder[5pt]{\includegraphics[width=0.11\textwidth]{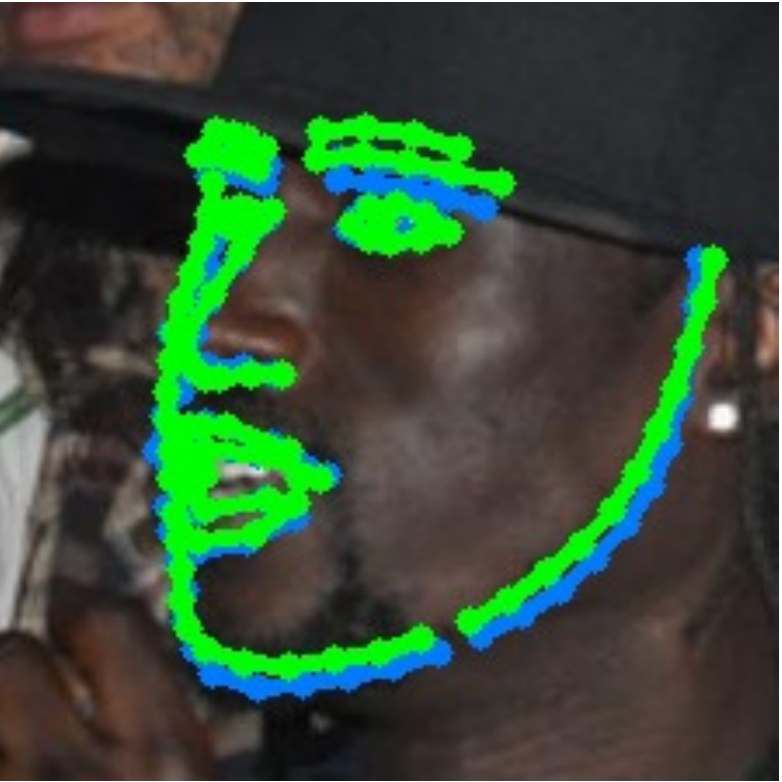}}{8.168}
\caption{First row shows LAB~\citep{Wu18} results, second row 3DDE results. We report the corresponding NME normalized by the eye corners distance. Blue and green colours represent ground truth and predictions respectively.}
\label{fig:lab_errors}
\end{figure}

Finally, we have also evaluated 3DDE with the newly released WFLW data set~\citep{Wu18}. In enables us to evaluate different sources of variability (\ie expressions, illumination, make-up, occlusions and blur). In Table~\ref{table:wflw} we provide the results of various competing methods~\citep{Wu18}, normalized by the eye corners distance. 3DDE outperforms its competitors in all the WFLW subsets by a large margin. We hypothesize that the reason for this is that the hybrid approach in 3DDE can be trained with less samples that some of its most prominent competitors and at the same time provide a very accurate face shape (see Fig.~\ref{fig:lab_errors}). Moreover, we achieve the best $AUC$ in all subsets, which determines that 3DDE is the best approach under all capture conditions (easy/frontal and difficult/profile) including all subsets that contain several types of difficulties.

\begin{table*}
\begin{center}
\tiny
\setlength\tabcolsep{0.25pt}
\begin{tabular}{l|ccc|ccc|ccc|ccc|ccc|ccc|ccc}
\hline
\multirow{3}{*}{Method} & \multicolumn{3}{c|}{Full} & \multicolumn{3}{c|}{Pose} & \multicolumn{3}{c|}{Expression} & \multicolumn{3}{c|}{Illumination} & \multicolumn{3}{c|}{Make-up} & \multicolumn{3}{c|}{Occlusion} & \multicolumn{3}{c}{Blur}\\
 & \multicolumn{3}{c|}{corners} & \multicolumn{3}{c|}{corners} & \multicolumn{3}{c|}{corners} & \multicolumn{3}{c|}{corners} & \multicolumn{3}{c|}{corners} & \multicolumn{3}{c|}{corners} & \multicolumn{3}{c}{corners}\\
 & $NME$ & $AUC_{10}$ & $FR_{10}$ & $NME$ & $AUC_{10}$ & $FR_{10}$ & $NME$ & $AUC_{10}$ & $FR_{10}$ & $NME$ & $AUC_{10}$ & $FR_{10}$ & $NME$ & $AUC_{10}$ & $FR_{10}$ & $NME$ & $AUC_{10}$ & $FR_{10}$ & $NME$ & $AUC_{10}$ & $FR_{10}$\\
\hline
ESR~\citep{Cao12} & 11.13 & 27.74 & 35.24 & 25.88 & 1.77 & 90.18 & 11.47 & 19.81 & 42.04 & 10.49 & 29.53 & 30.80 & 11.05 & 24.85 & 38.84 & 13.75 & 19.46 & 47.28 & 12.20 & 22.04 & 41.40\\
SDM~(Xiong \etal) & 10.29 & 30.02 & 29.40 & 24.10 & 2.26 & 84.36 & 11.45 & 22.93 & 33.44 & 9.32 & 32.37 & 26.22 & 9.38 & 31.25 & 27.67 & 13.03 & 20.60 & 41.85 & 11.28 & 23.98 & 35.32\\
CFSS~\citep{Zhu15} & 9.07 & 36.59 & 20.56 & 21.36 & 6.32 & 66.26 & 10.09 & 31.57 & 23.25 & 8.30 & 38.54 & 17.34 & 8.74 & 36.91 & 21.84 & 11.76 & 26.88 & 32.88 & 9.96 & 30.37 & 23.67\\
LAB~\citep{Wu18} & 5.27 & 53.23 & 7.56 & 10.24 & 23.45 & 28.83 & 5.51 & 49.51 & 6.37 & 5.23 & 54.33 & 6.73 & 5.15 & 53.94 & 7.77 & 6.79 & 44.90 & 13.72 & 6.32 & 46.30 & 10.74\\
\hline
3DDE & \textbf{4.68} & \textbf{55.44} & \textbf{5.04} & \textbf{8.62} & \textbf{26.40} & \textbf{22.39} & \textbf{5.21} & \textbf{51.75} & \textbf{5.41} & \textbf{4.65} & \textbf{56.02} & \textbf{3.86} & \textbf{4.60} & \textbf{55.36} & \textbf{6.79} & \textbf{5.77} & \textbf{46.92} & \textbf{9.37} & \textbf{5.41} & \textbf{49.57} & \textbf{6.72}\\
\hline
\end{tabular}
\end{center}
\caption{Error of face alignment methods on WFLW.}
\label{table:wflw}
\end{table*}

\subsection{Ablation study}\label{sec:ablation}
3DDE is based on three key ideas: 3D initialization, a cascaded ERT regressor operating on probabilistic CNN features and a coarse-to-fine scheme. 
In this section we analyze the contribution of each one to the overall performance of our algorithm.

In Table~\ref{table:ablation} we show the results obtained by different configurations of our framework when evaluated on WFLW. We have chosen WFLW in our study because it allows the analysis of results stratified by different types of difficulties (\ie facial expressions, large poses, illumination changes, etc.). In this case, since there are many profile faces, we use the height as normalization for the NME. So, the numerical values are not directly comparable to those in Table~\ref{table:wflw}. \verb$MS$ stands for ``mean shape initialization'' of the ERT. \verb$3D$ means to initialize the ERT with the procedure in section~\ref{sec:rigid}. \verb$SE$ denotes using plain gray level features for the ERT whereas \verb$DE$ denotes using probability maps produced by the CNN to train the ERT. Finally \verb$CF$ stands for using the coarse-to-fine scheme.

When combined with the cascaded ERT, the 3D initialization is key to achieve top overall performance, see \verb$CNN+MS+DE$ vs \verb$CNN+3D+DE$ in the \emph{full} subset. The reason for this is that, in the 3D case, the initialization takes care of the rigid component of face pose so that the ERT cascade only models non-rigid deformations. Moreover, the projection of the 3D face model is a correct 2D shape, a requirement for the ERT to converge to a valid face shape~\citep{Cao14}. Of course, the 3D initialization is fundamental to achieve good performance in presence of large face rotations. So, it provides the largest improvement in the \emph{pose} subset.

The use of CNN probability maps improves the NME in the \emph{full} data set in about 20\% (see \verb$CNN+3D+SE$ vs \verb$CNN+3D+DE$). The large receptive fields of CNNs are specially helpful in challenging situations, specifically those in the \emph{pose} and \emph{occlusion} subsets.

\begin{figure*}
  \centering
  \subfloat[Evolution of the NME through the different stages in the cascade]{
  \label{fig:learning:a}
  \includegraphics[width=0.5\textwidth]{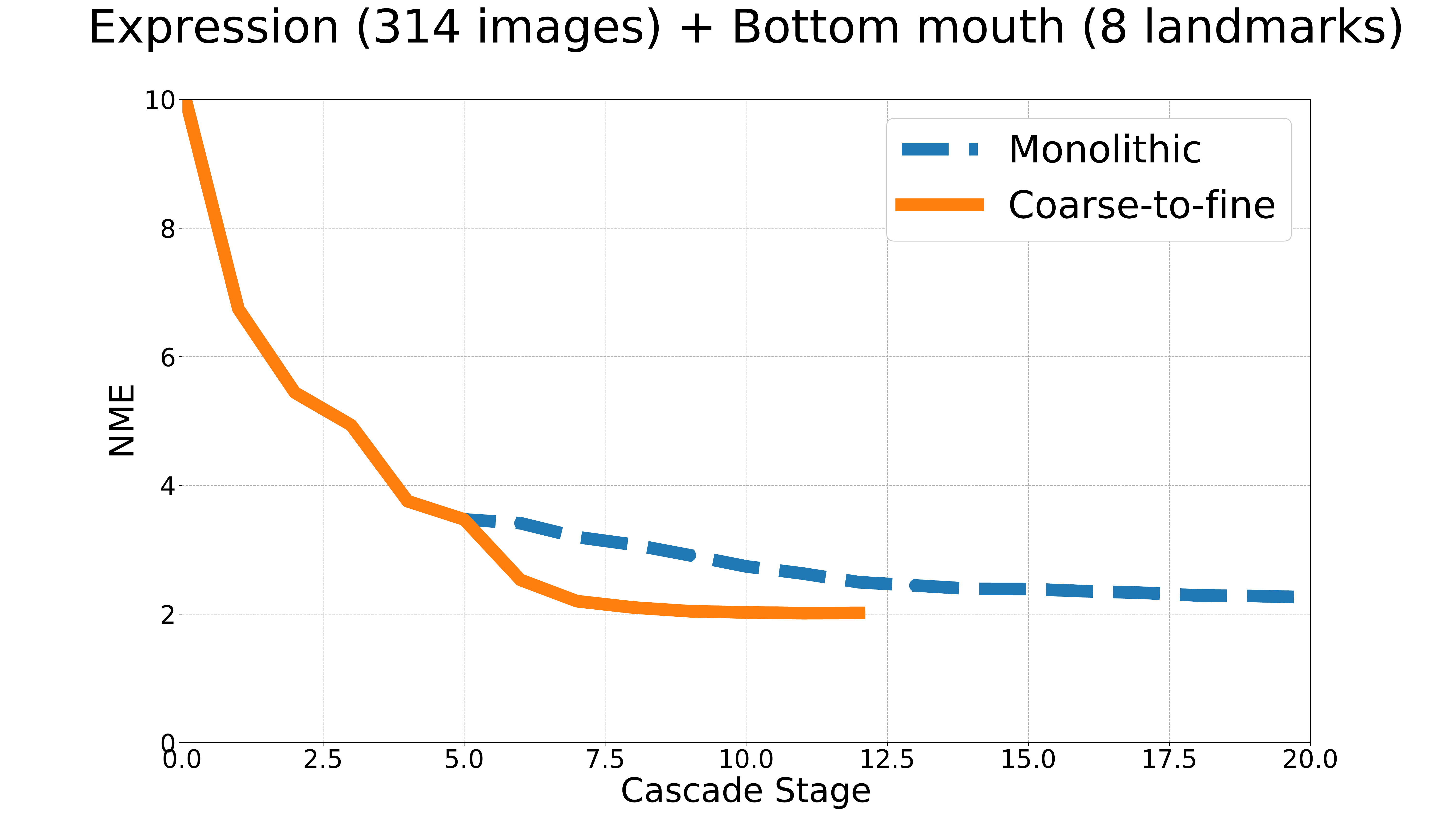}
  \includegraphics[width=0.5\textwidth]{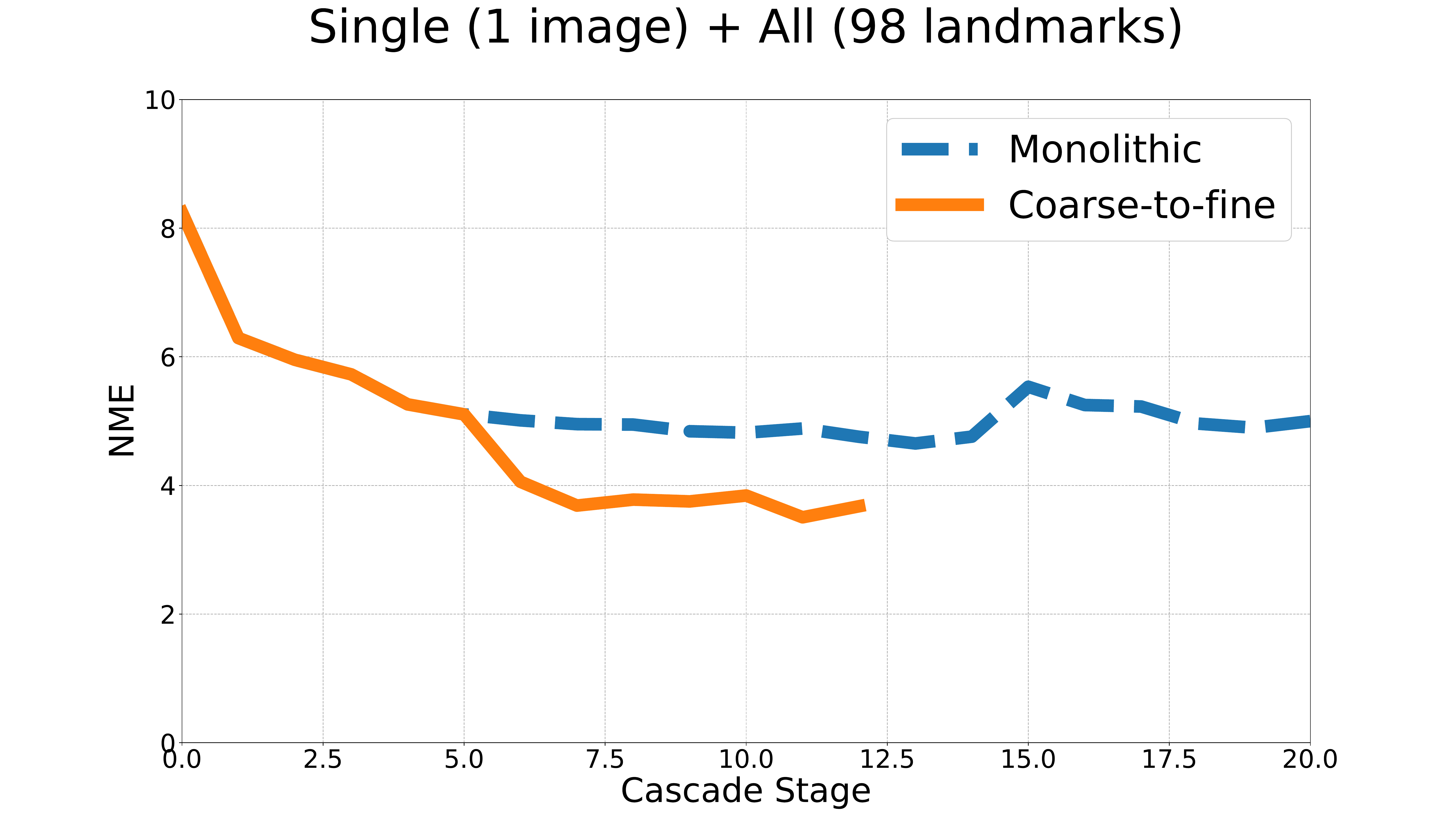}}\\
  \subfloat[Monolithic]{
  \label{fig:learning:b}
  \includegraphics[width=0.22\textwidth]{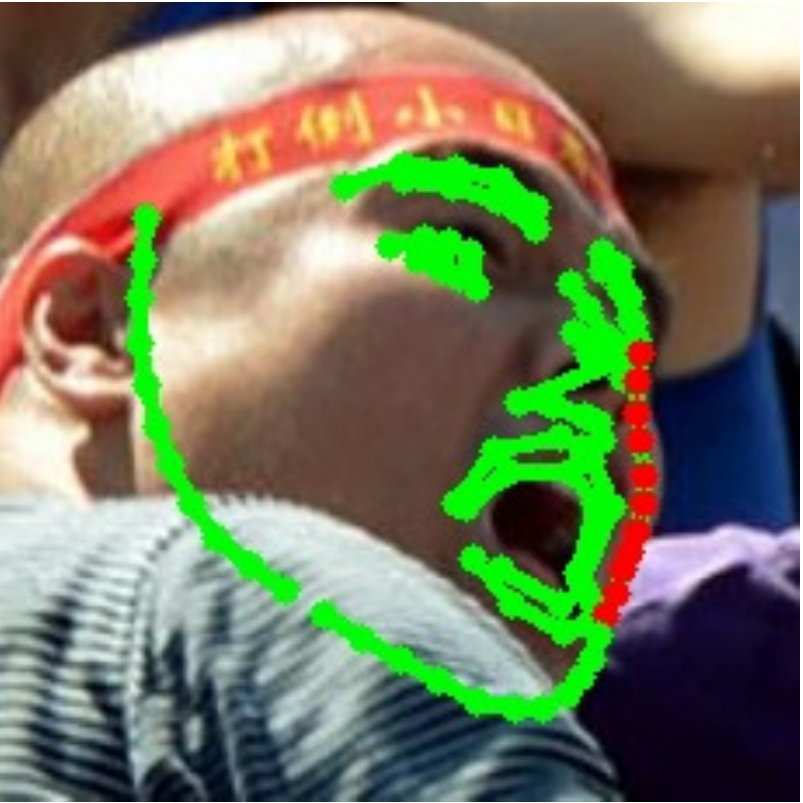}
  \includegraphics[width=0.22\textwidth]{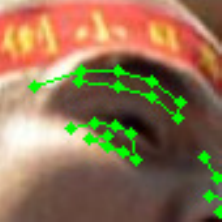}}
  \subfloat[Coarse-to-fine]{
  \label{fig:learning:c}
  \includegraphics[width=0.22\textwidth]{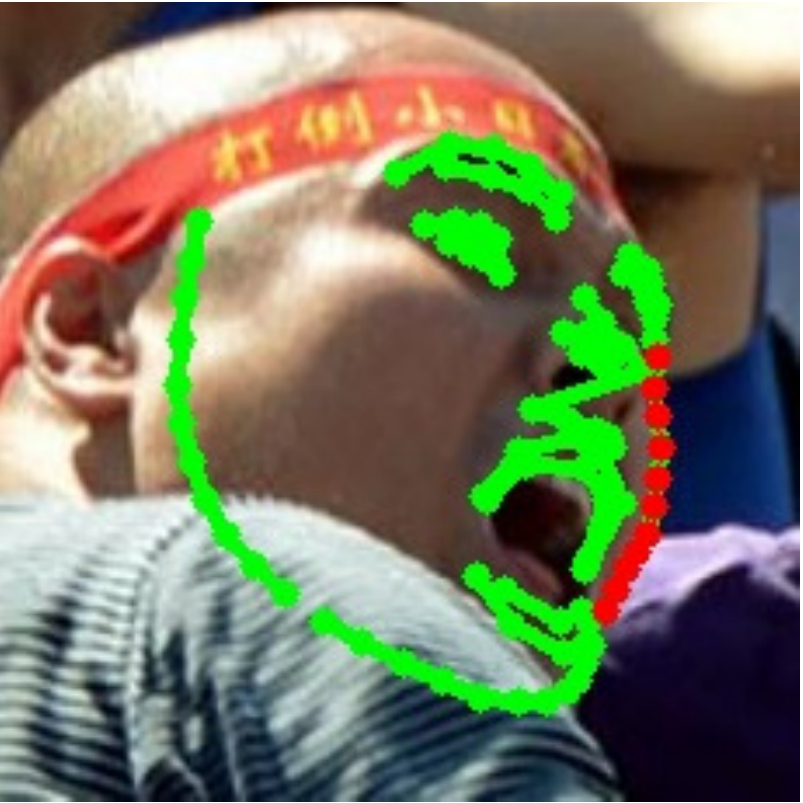}
  \includegraphics[width=0.22\textwidth]{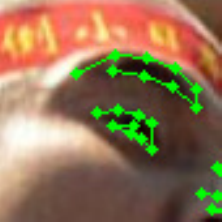}}
  \caption{Example of a monolithic ERT regressor vs. our coarse-to-fine approach. (a) NME evolution through the stages in the cascade (left plot, 8 mouth landmarks for all test images in the expression subset; right plot, all 98 landmarks in one image). (b) predicted shape and zoom-in with a monolithic regressor. (c) predicted shape and zoom-in with our coarse-to-fine approach.}
  \label{fig:learning}
\end{figure*}

The coarse-to-fine strategy in our cascaded ERT provides significative local improvements in difficult cases, with rare facial part combinations (see Fig.~\ref{fig:learning:a}). For this reason, the largest gain of \verb$CNN+3D+DE+CF$ vs \verb$CNN+3D+DE$ occurs in the \emph{expressions} subset. Although this strategy provides improvements in all the data base subsets, the actual NME differences are washed out when averaged over the number of landmarks in the face and the number of images in the subset. They may be appreciated by looking into specific data subsets or samples (see Fig.~\ref{fig:learning:a}),  such as the left eyebrow/eye location improvement in Fig.~\ref{fig:learning:b} and~\ref{fig:learning:c} (best viewed after zoom-in).

\begin{table*}
\tiny
\begin{center}
\setlength\tabcolsep{0.25pt}
\begin{tabular}{l|ccc|ccc|ccc|ccc|ccc|ccc|ccc}
\hline
\multirow{3}{*}{Method} & \multicolumn{3}{c|}{Full} & \multicolumn{3}{c|}{Pose} & \multicolumn{3}{c|}{Expression} & \multicolumn{3}{c|}{Illumination} & \multicolumn{3}{c|}{Make-up} & \multicolumn{3}{c|}{Occlusion} & \multicolumn{3}{c}{Blur}\\
 & \multicolumn{3}{c|}{height} & \multicolumn{3}{c|}{height} & \multicolumn{3}{c|}{height} & \multicolumn{3}{c|}{height} & \multicolumn{3}{c|}{height} & \multicolumn{3}{c|}{height} & \multicolumn{3}{c}{height}\\
 & $NME$ & $AUC_4$ & $FR_4$ & $NME$ & $AUC_4$ & $FR_4$ & $NME$ & $AUC_4$ & $FR_4$ & $NME$ & $AUC_4$ & $FR_4$ & $NME$ & $AUC_4$ & $FR_4$ & $NME$ & $AUC_4$ & $FR_4$ & $NME$ & $AUC_4$ & $FR_4$\\
\hline
CNN+3D+SE & 2.52 & 41.10 & 11.56 & 3.53 & 24.08 & 28.83 & 2.90 & 33.22 & 15.92 & 2.53 & 41.85 & 10.45 & 2.59 & 39.08 & 15.53 & 3.06 & 31.10 & 22.14 & 2.91 & 33.98 & 15.78\\
CNN+MS+DE & 2.23 & 49.77 & 7.04 & 3.33 & 35.13 & 17.79 & 2.56 & 45.15 & 8.91 & 2.17 & 49.29 & 5.87 & 2.33 & 46.85 & 9.70 & 2.69 & 40.33 & 12.90 & 2.53 & 42.71 & 9.57\\
CNN+3D+DE & 2.03 & 51.14 & 5.47 & 2.68 & 39.55 & 11.96 & 2.21 & 46.66 & 7.96 & 2.11 & 50.09 & 5.01 & 2.13 & 48.57 & 7.28 & 2.56 & 40.83 & 12.36 & 2.40 & 43.84 & 8.27\\
\hline
CNN+3D+DE+CF & 2.01 & 51.67 & 5.20 & 2.63 & 39.90 & 10.73 & 2.15 & 48.19 & 5.73 & 2.06 & 50.79 & 4.87 & 2.12 & 49.05 & 7.28 & 2.54 & 40.94 & 12.22 & 2.39 & 43.93 & 8.02\\
\hline
\end{tabular}
\end{center}
\caption{Ablation study. MS and 3D represent the 2D mean shape and 2D projections of the 3D mean face respectively. SE and DE represent the type of features used in the cascade being simple grayscale features and deep probability maps features respectively. The CNN+3D+DE+CF row represents the full 3DDE approach results.}
\label{table:ablation}
\end{table*}

\begin{figure}
  \centering
  \includegraphics[width=0.45\textwidth]{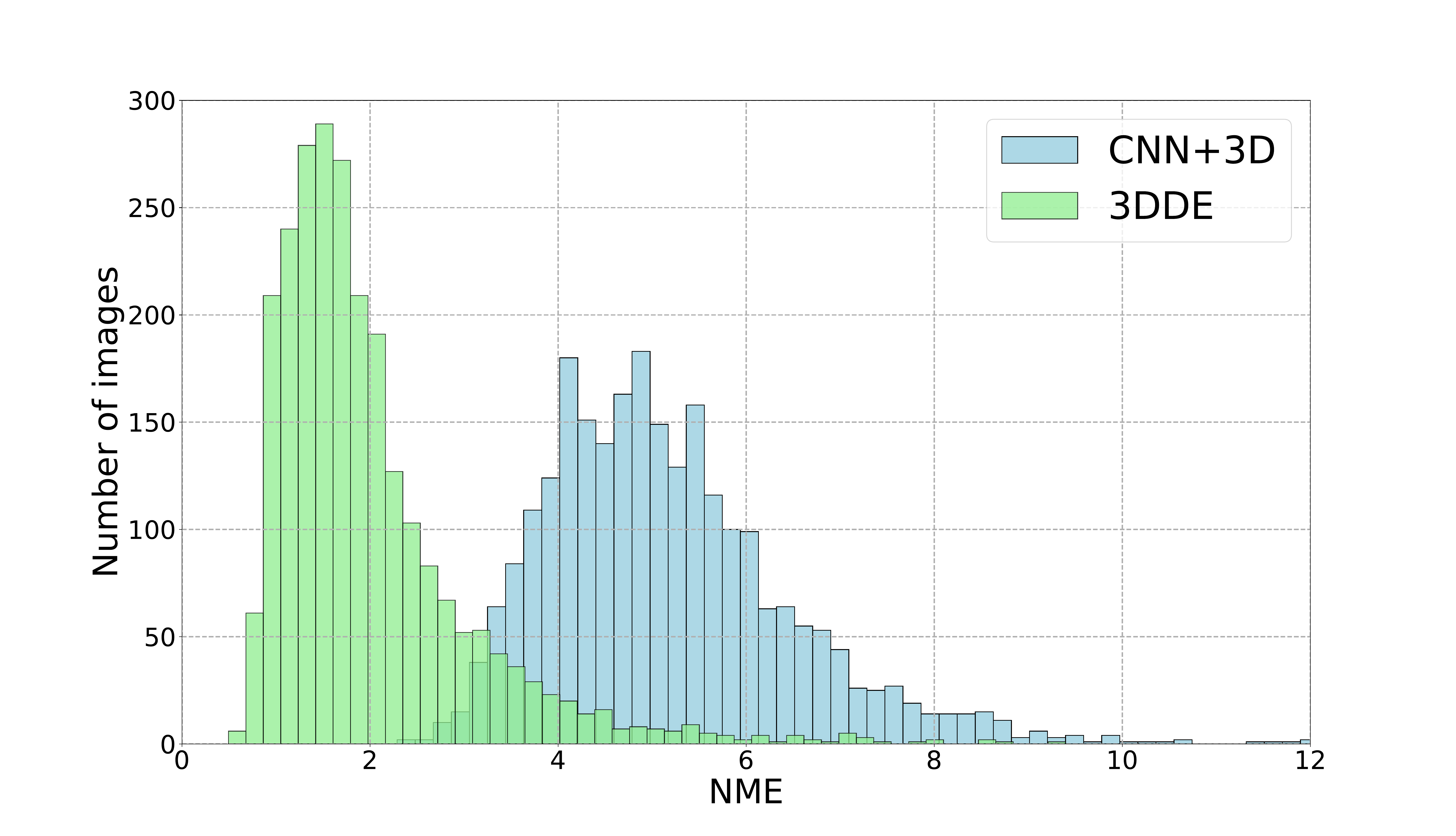}
  \caption{Sample distribution of NMEs produced by the CNN+3D and 3DDE regressors. We use the height as normalization for the NME.}
  \label{fig:improvement}
\end{figure}

Finally, we analyze the NME distribution produced by the rigid initialization and the final 3DDE model (see Fig.~\ref{fig:improvement}). Using the model trained for the WFLW experiment, we align the 2500 test samples of WFLW and plot the distribution of NMEs, produced both with the \verb$CNN+3D$ regressor (softPOSIT result) and the full \verb$CNN+3D+DE+CF$ regressor (3DDE result). The values of percentiles 10 and 90 of the NME distribution are 3.71 and 6.87 for the \verb$CNN+3D$ regressor and 1.03 and 3.32 for the \verb$CNN+3D+DE+CF$ one. So, on average, the full regressor reduces in about 60\% the NME achieved by the rigid initialization.


\subsection{Cross-dataset evaluation}
\label{sec:semi-supervised}

\begin{figure}
  \centering
  \subfloat[Distinct landmarks]{
    \label{fig:cross:a}
    \includegraphics[width=0.16\textwidth]{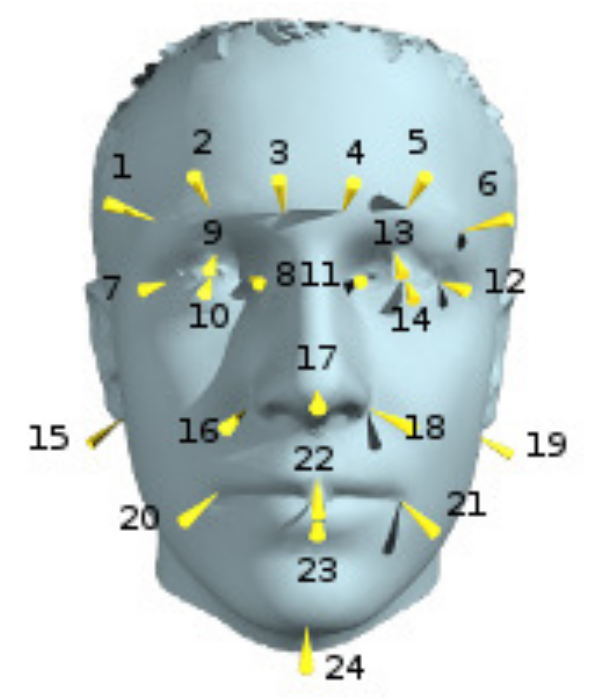}
    \includegraphics[width=0.15\textwidth]{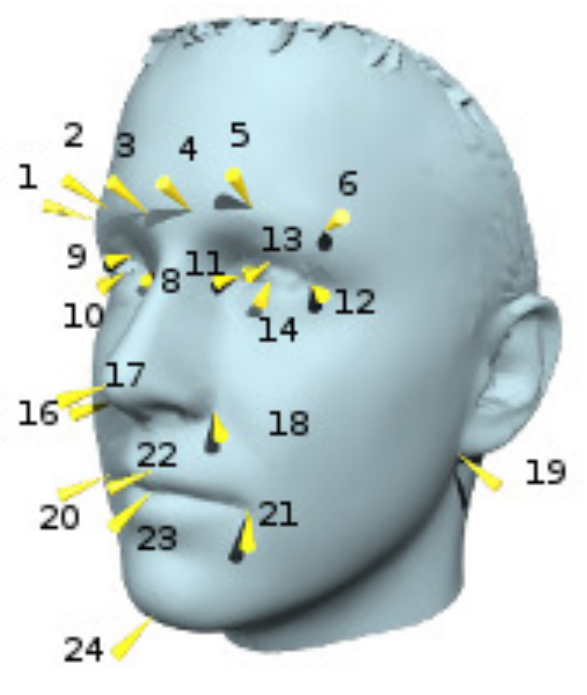}}\\
  \subfloat[NME per landmark]{
    \label{fig:cross:b}
    \includegraphics[width=0.45\textwidth]{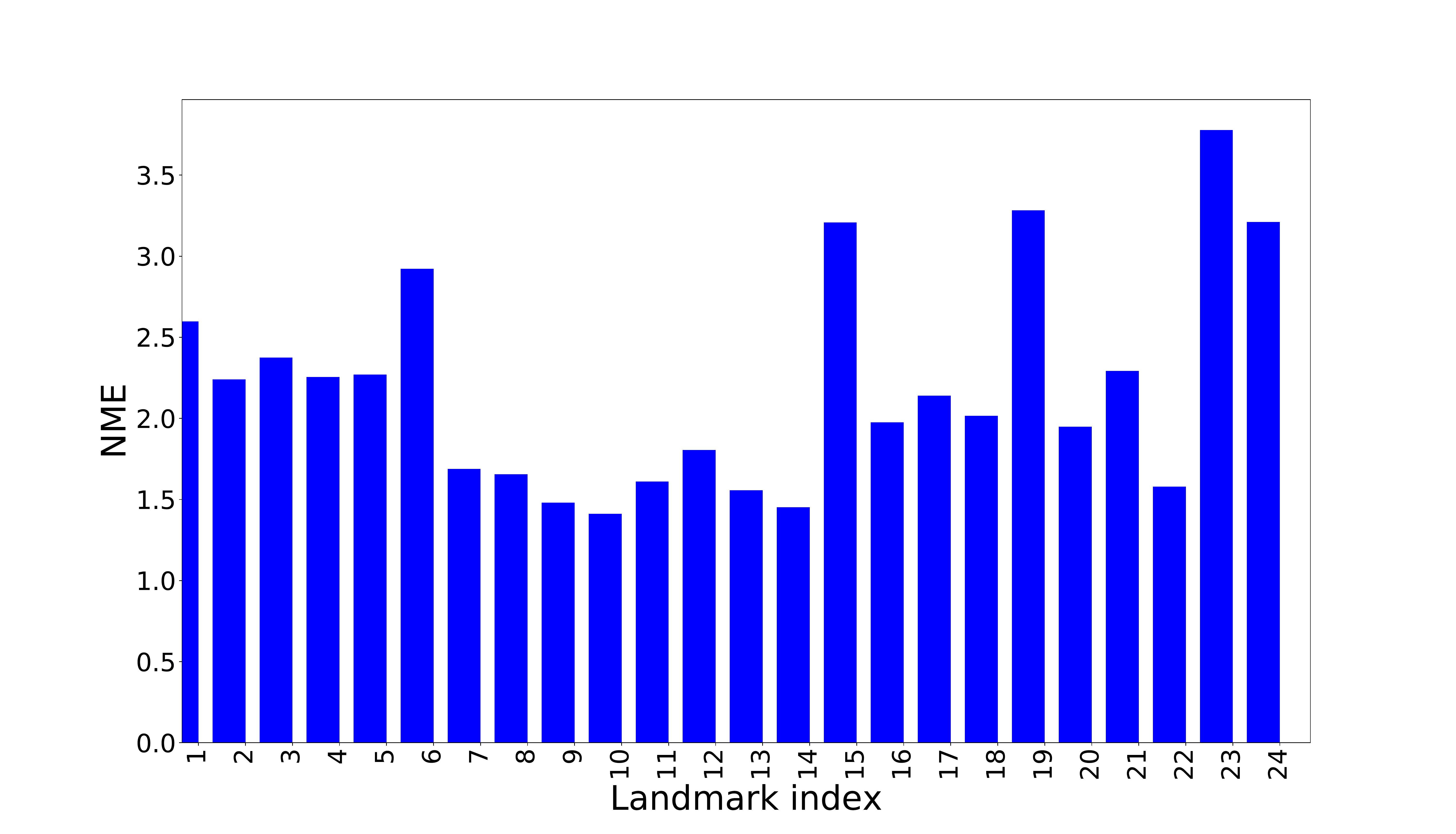}}
  \caption{Location of distinct face landmarks and the NME related to each landmark.}
  \label{fig:cross}
\end{figure}

In this section we perform cross-dataset experiments to evaluate the quality of present benchmarks and the generalization of the regressors trained on them. Here we benefit from the fact that 3DDE may be trained in a semi-supervised way, \ie using data sets with missing or unlabeled landmarks. To this end we select 24 distinct facial landmarks (see Fig.~\ref{fig:cross:a}). We consider them distinct because they may be accurately located by a human annotator. We train and evaluate 3DDE respectively with the training and test sets of each data base. We have also performed one more experiment training 3DDE with the training sets of all data bases and evaluating it successively with the tests sets of each of them, we denote this experiment with label \verb$All$.

In Table~\ref{table:cross} we show the results of our evaluation. The smallest data base, COFW, has the worst cross-dataset results. On the other hand, the data set with greatest diversity, WFLW, has the best results. Moreover, the model \verb$All$, trained with the training sets of all data bases, is able to improve, in all cross-dataset experiments, the models trained in a single data set. However, the most prominent outcome of this experiment is that we always achieve the best result when training with the train subset of the same data base. This holds even when compared against the model trained with all data sets, confirming the existence of the so-called ``data set bias" in current benchmarks~\citep{Torralba11}. 

In a final experiment we use model \verb$All$ to evaluate the NME of each landmark using the test sets of all data sets (see Fig.~\ref{fig:cross:b}). The landmarks with highest NME are those related to the ears, the bottom of the mouth and the chin.

\begin{table}
    \footnotesize
    \begin{center}
    \setlength\tabcolsep{2.5pt}
    \begin{tabular}{l|c|c|c|c|c}
    \hline
    \diagbox{Train}{Test} & 300W & COFW & AFLW & WFLW & All\\
    \hline
    300W & \textbf{2.00} & 3.11 & 4.90 & 3.44 & 4.15\\
    COFW & 3.68 & \textbf{2.09} & 4.56 & 4.03 & 4.19\\
    AFLW & 4.19 & 2.51 & \textbf{2.15} & 3.29 & 2.65\\
    WFLW & 2.57 & 2.53 & 3.28 & \textbf{1.70} & 2.71\\
    All & 2.34 & 2.23 & 2.41 & 1.96 & \textbf{2.26}\\
    \hline
    \end{tabular}
    \end{center}
    \caption{Cross-dataset experiment using only distinct landmarks to compute NME normalized by height.}
    \label{table:cross}
\end{table}

\section{Conclusions}\label{sec:conclusions}
We have introduced 3DDE, a robust face alignment method that leverages on good properties of CNNs, cascade of ERT and 3D face models. The CNN provides robust landmark estimations with weak face shape enforcement. The ERT is able to enforce the face shape and achieve better accuracy in landmark detection, but it only converges with a good initialization. Finally, 3D models exploit face orientation information to improve self-occlusion estimation. 

3DDE is initialized by robustly fitting a 3D face model to the probability maps produced by the CNN. The 3D model enables 3DDE to handle self-occlusions and successfully deal with both frontal and profile faces. Once initialized, the cascade of ERT only models the non-rigid component of face motion. It provides various benefits, namely, it enforces shape consistency, may be trained with unlabeled landmarks, estimate landmark visibility and efficiently parallelize the execution of the regression trees within each stage. We have additionally introduced a coarse-to-fine scheme within the cascade of  ERT that is able to deal with the combinatorial explosion of local parts deformation. In this case, the usual monolithic ERT will perform poorly when fitting faces with combinations of facial part deformations not present in the training set. This is a fundamental limitation of implicit shape models addressed by 3DDE.

In the experiments we have shown that 3DDE improves, as far as we know, the state-of-the-art performance in 300W, COFW, AFLW and WFLW data sets. In our ablation analysis we have shown that all the components of the system critically contribute to the final result. 

\begin{figure*}
  \centering
  \subfloat[300W public]{
    \label{fig:results:a}
    \includegraphics[width=0.15\textwidth]{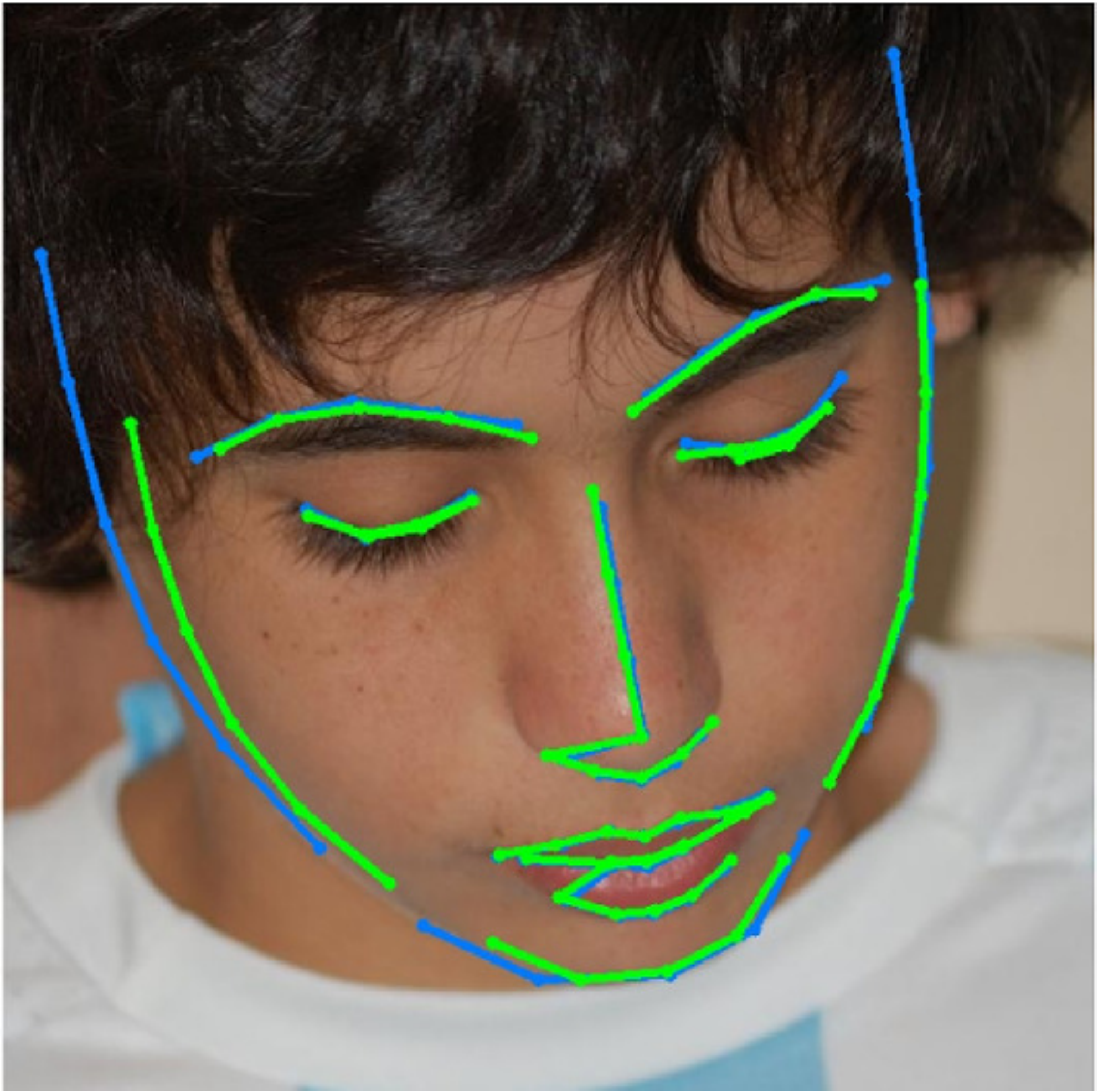}
    \includegraphics[width=0.15\textwidth]{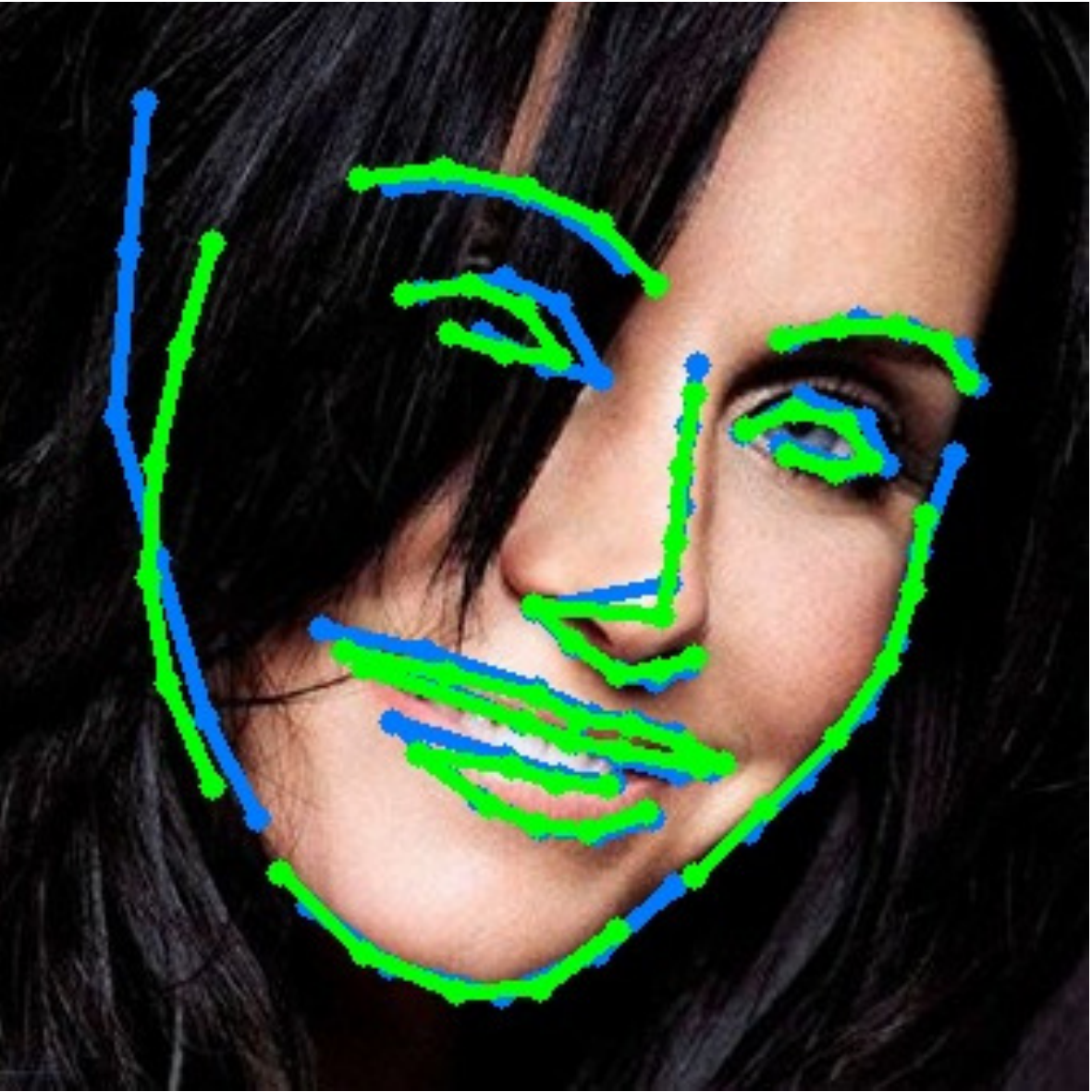}
    \includegraphics[width=0.15\textwidth]{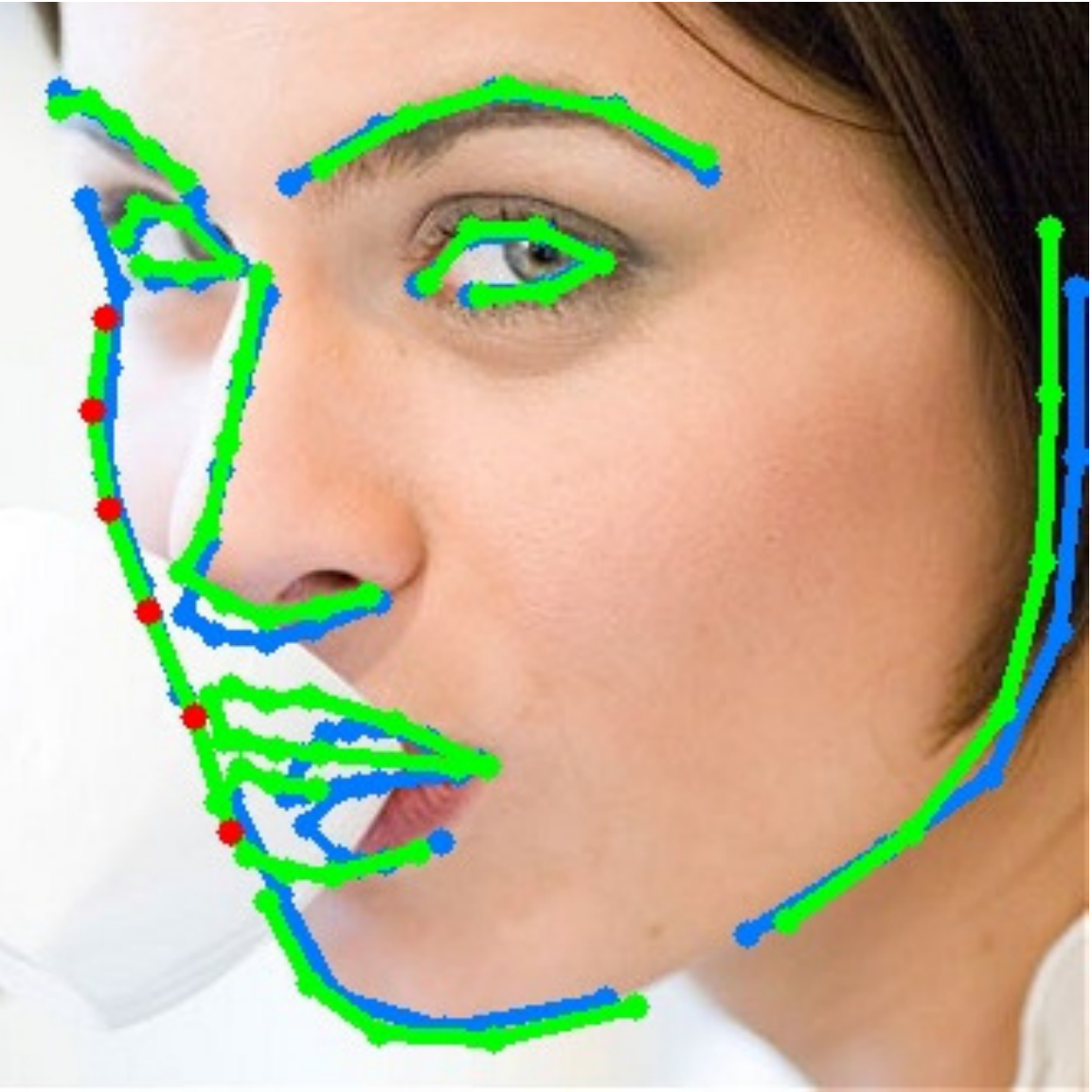}
    \includegraphics[width=0.15\textwidth]{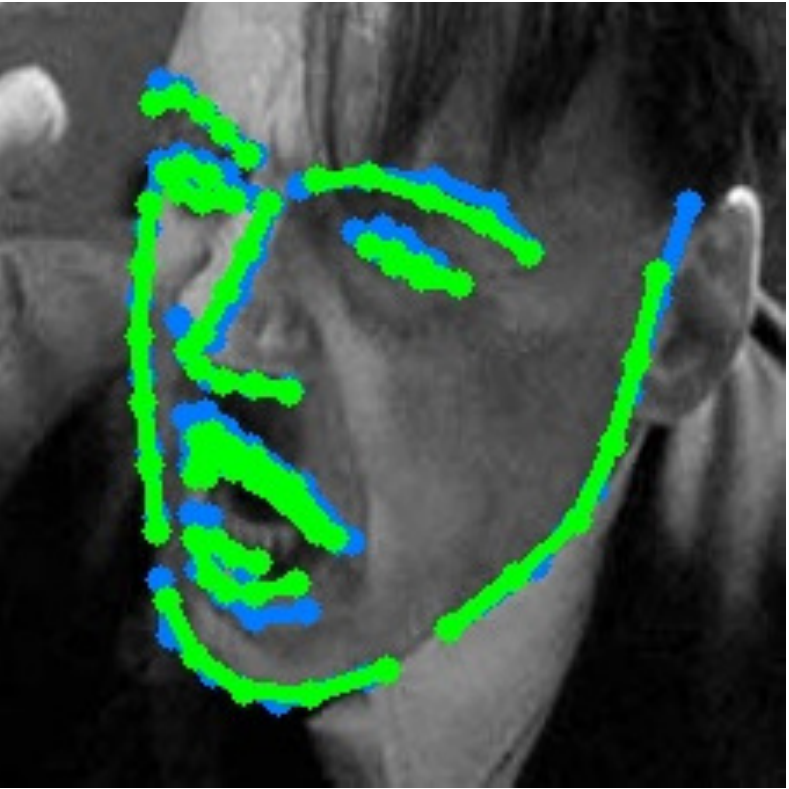}
    \includegraphics[width=0.15\textwidth]{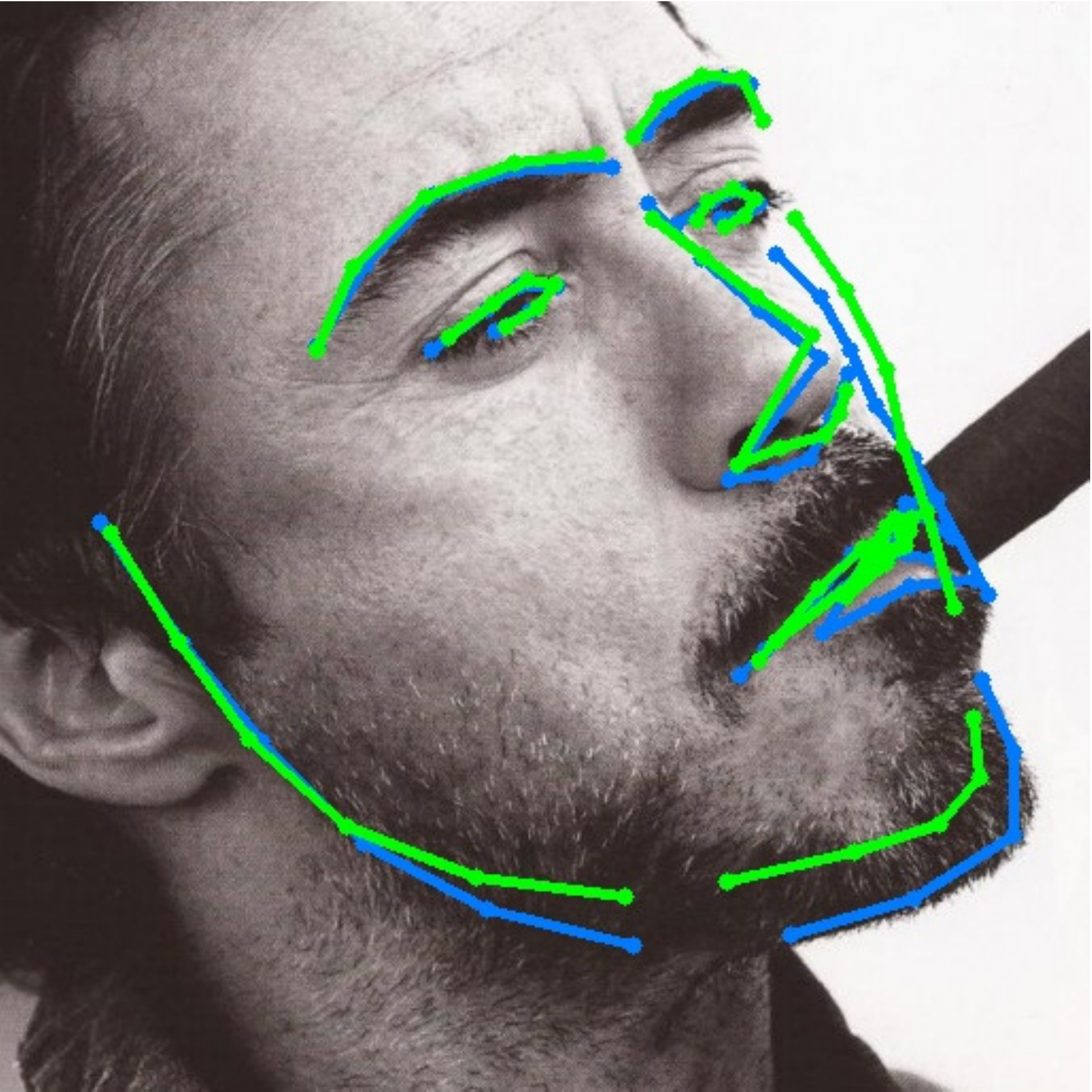}
    \includegraphics[width=0.15\textwidth]{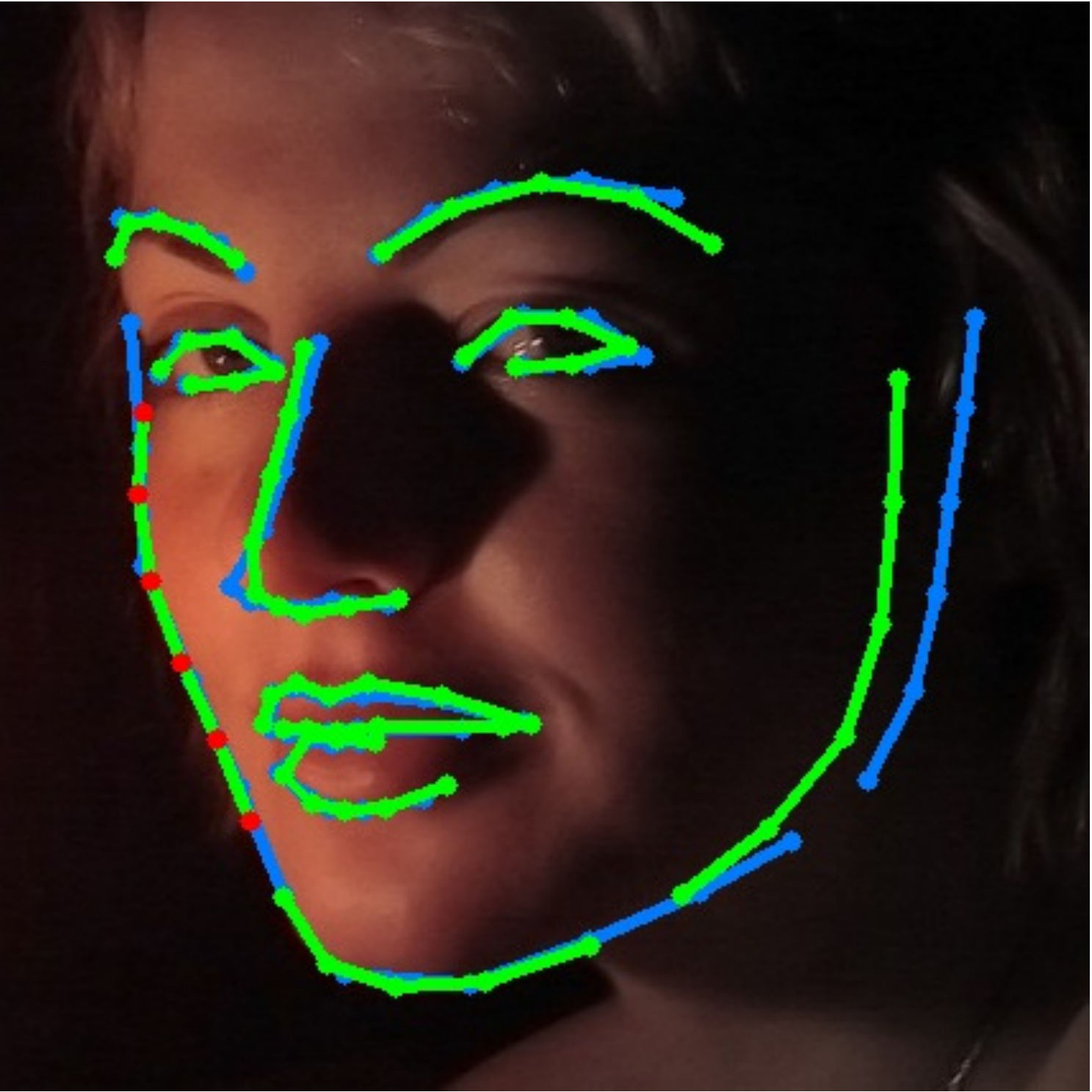}}
  \hfill
  \subfloat[300W private]{
    \label{fig:results:b}
    \includegraphics[width=0.15\textwidth]{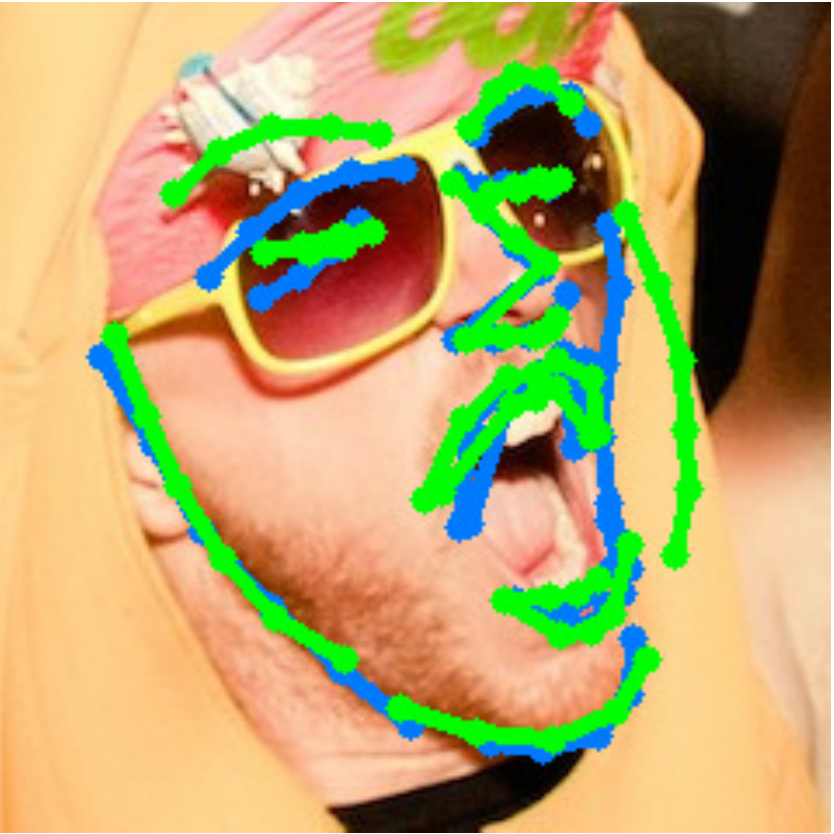}
    \includegraphics[width=0.15\textwidth]{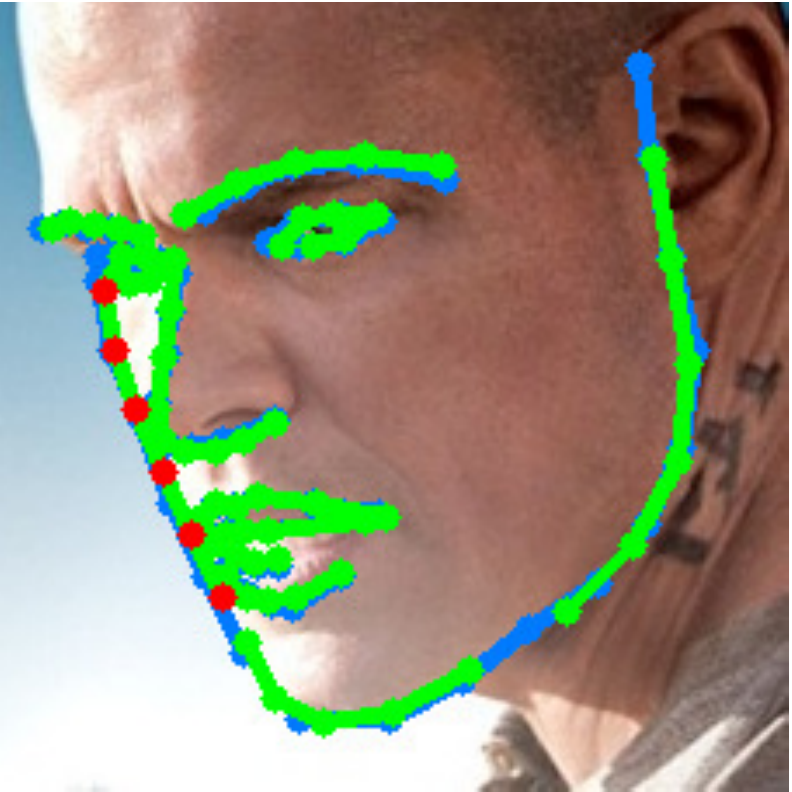}
    \includegraphics[width=0.15\textwidth]{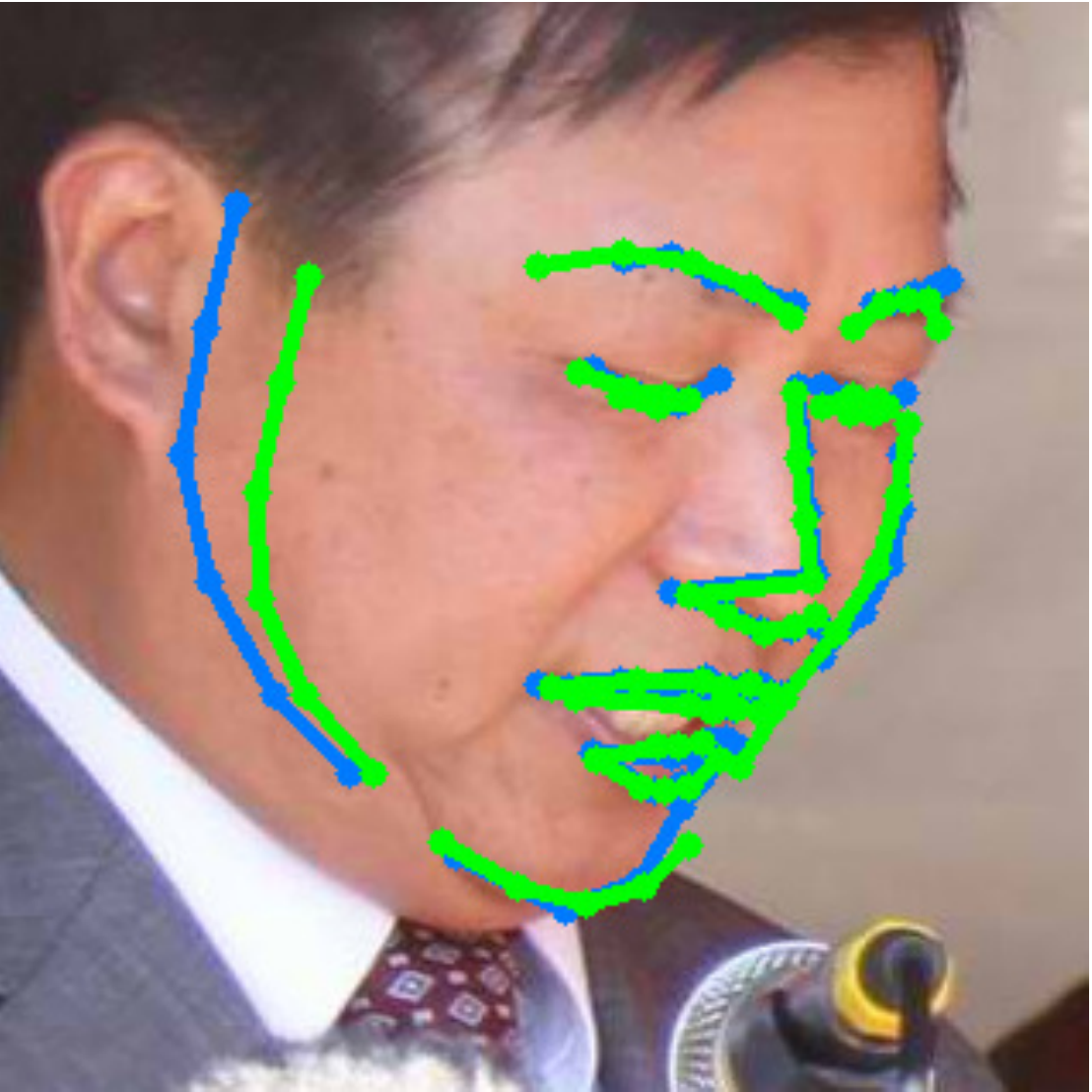}
    \includegraphics[width=0.15\textwidth]{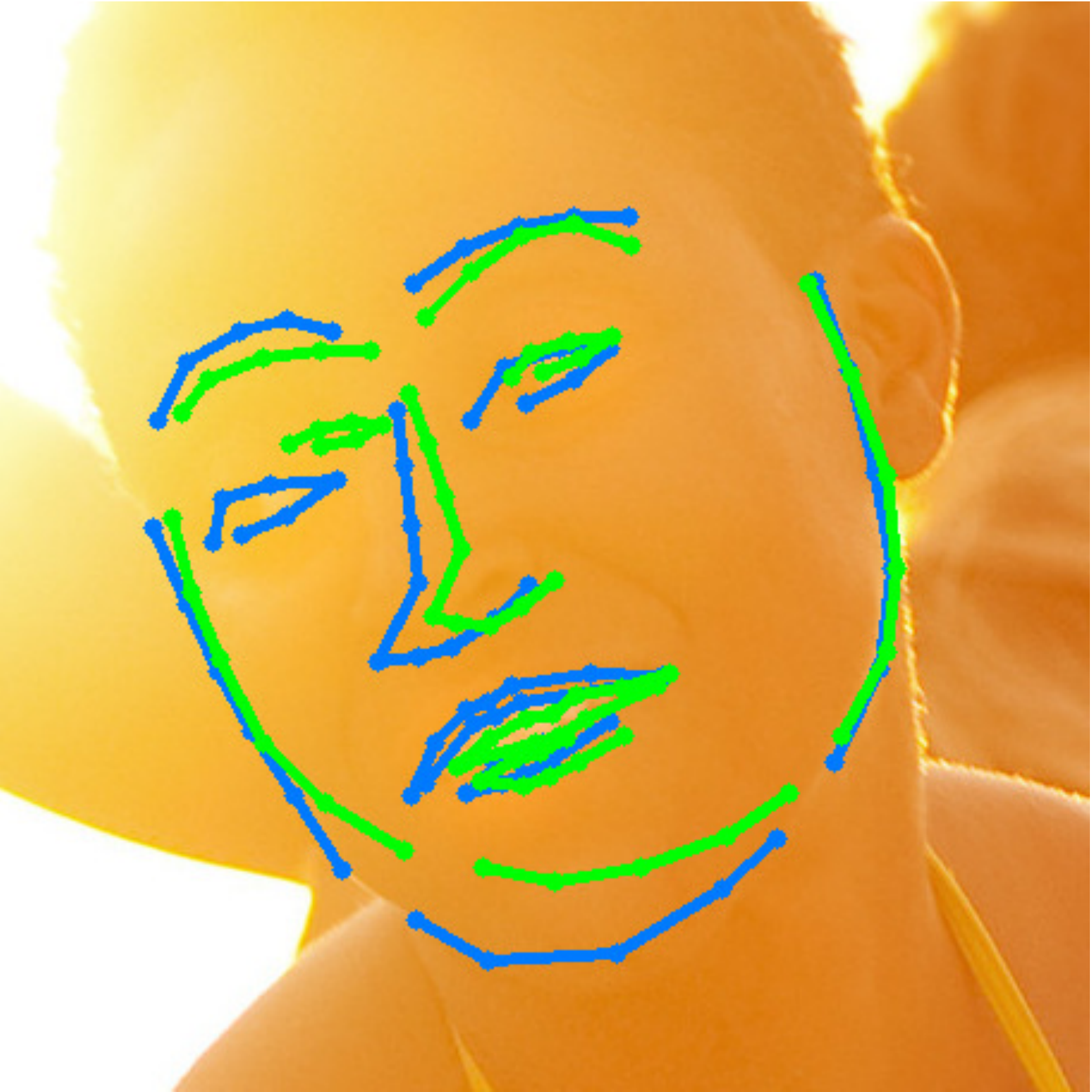}
    \includegraphics[width=0.15\textwidth]{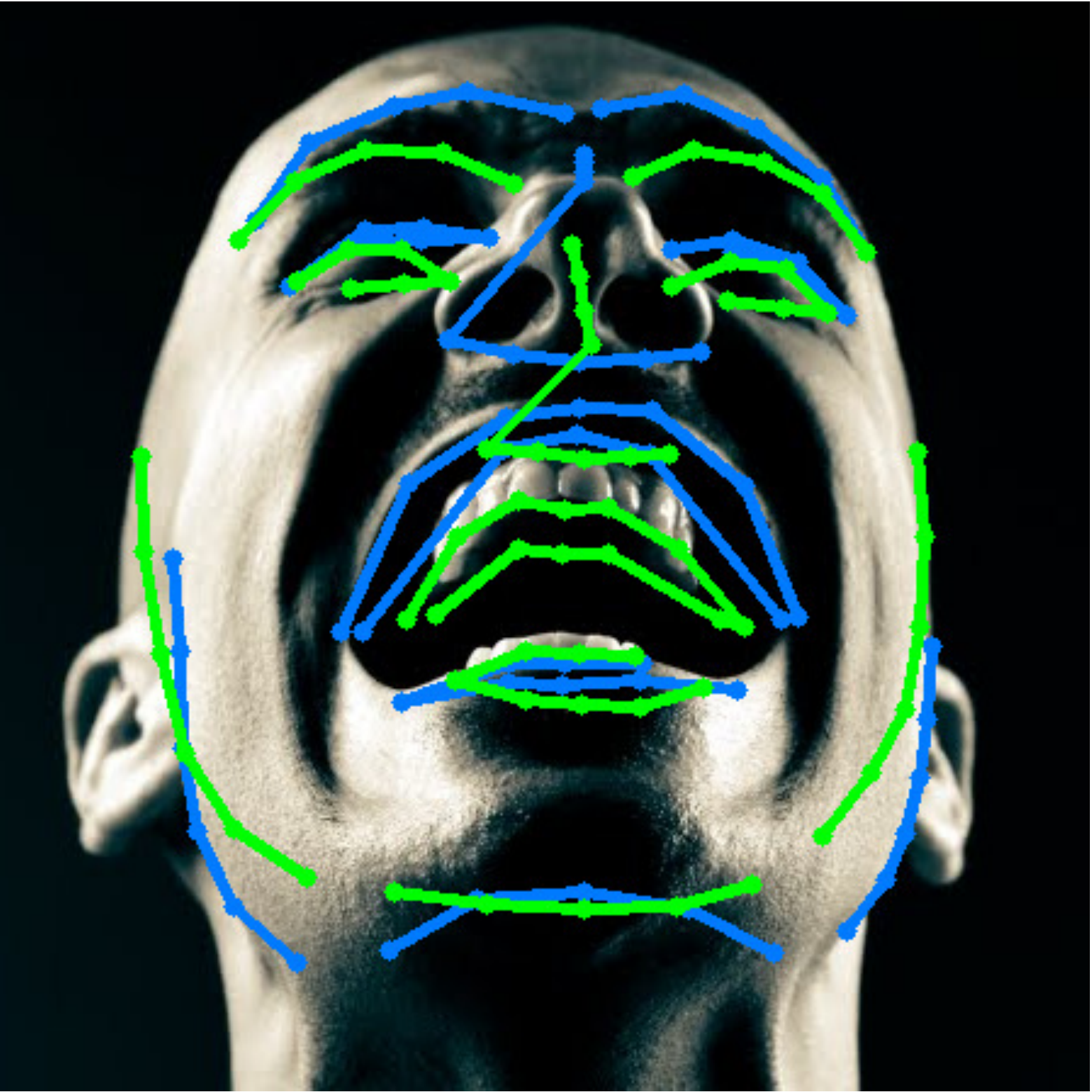}
    \includegraphics[width=0.15\textwidth]{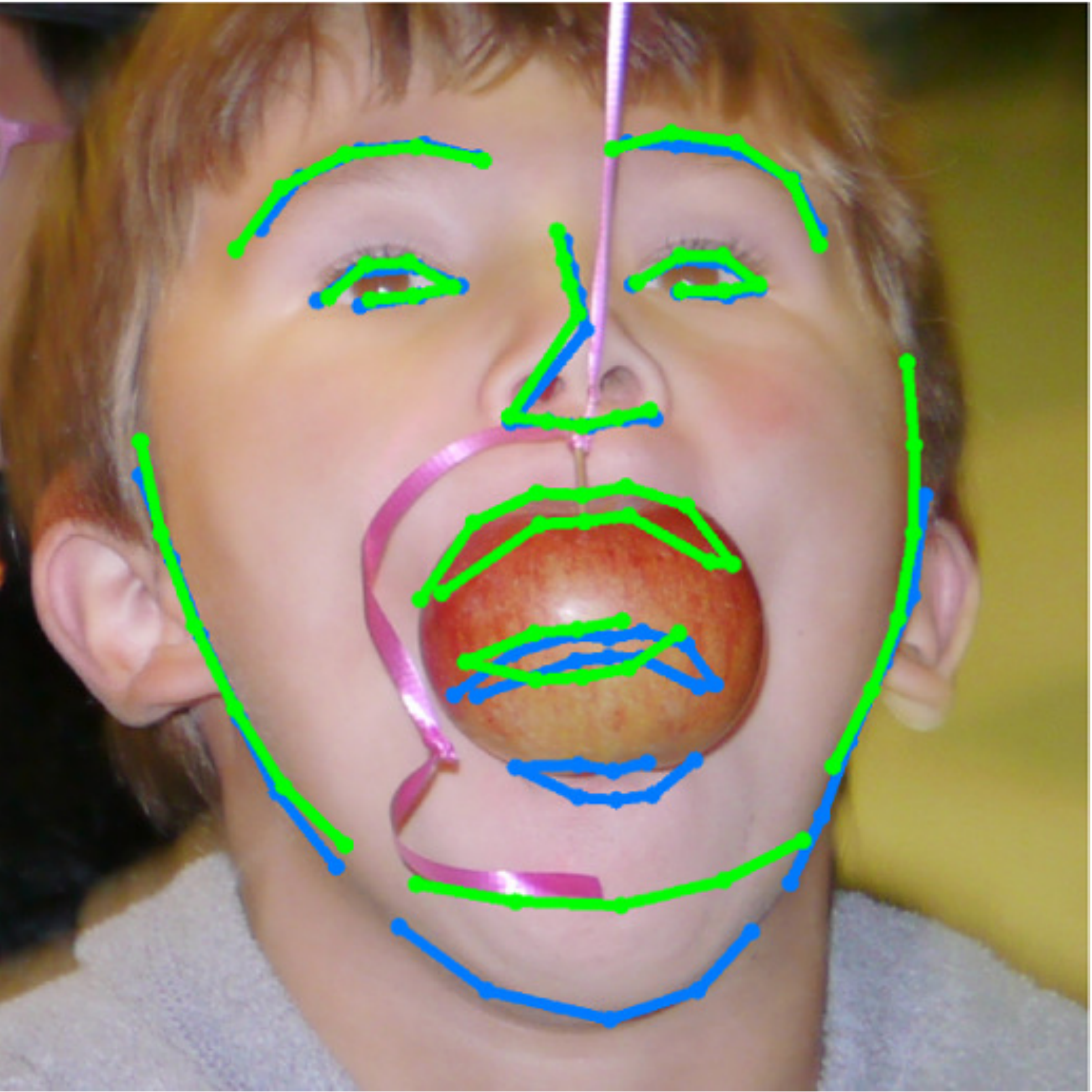}}
  \hfill
  \subfloat[COFW]{
    \label{fig:results:c}
    \includegraphics[width=0.15\textwidth]{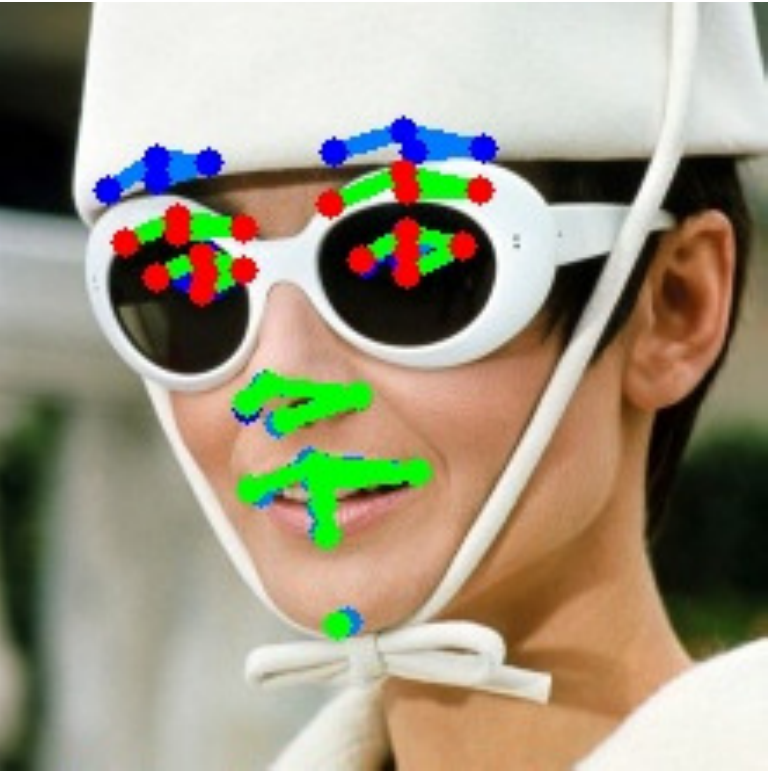}
    \includegraphics[width=0.15\textwidth]{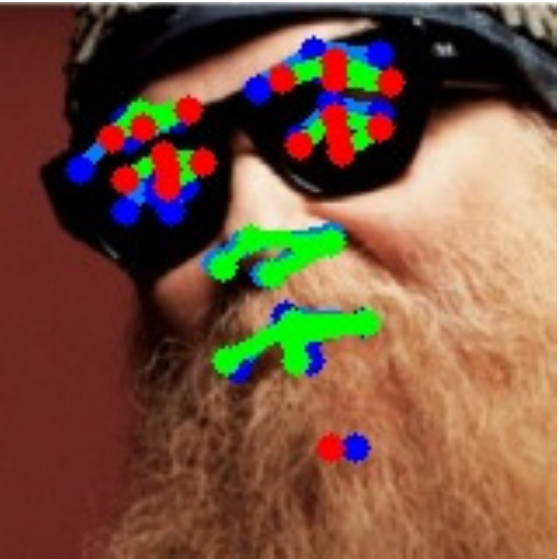}
    \includegraphics[width=0.15\textwidth]{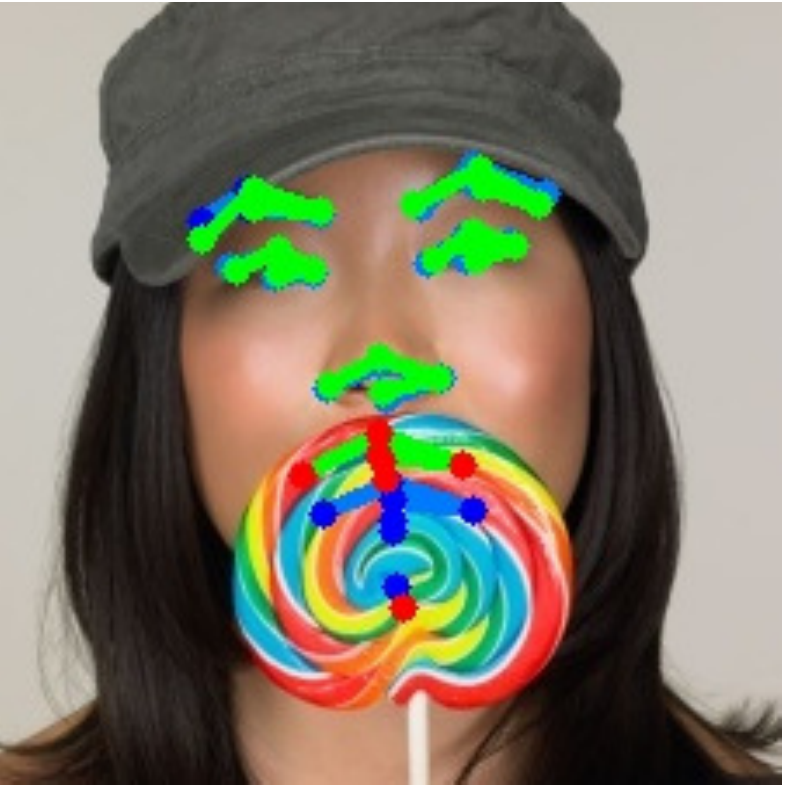}
    \includegraphics[width=0.15\textwidth]{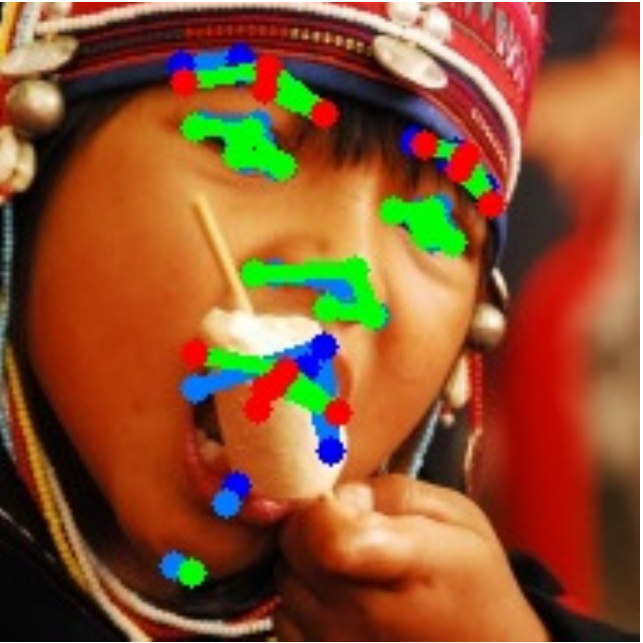}
    \includegraphics[width=0.15\textwidth]{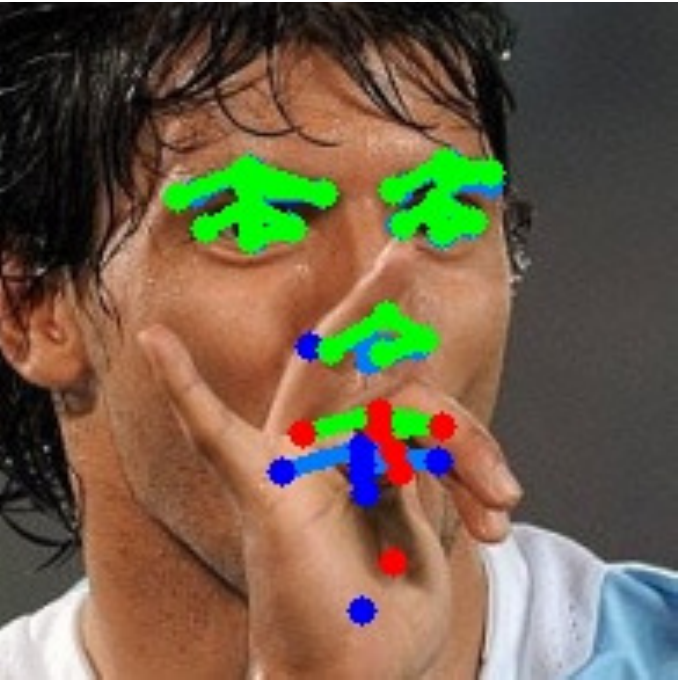}
    \includegraphics[width=0.15\textwidth]{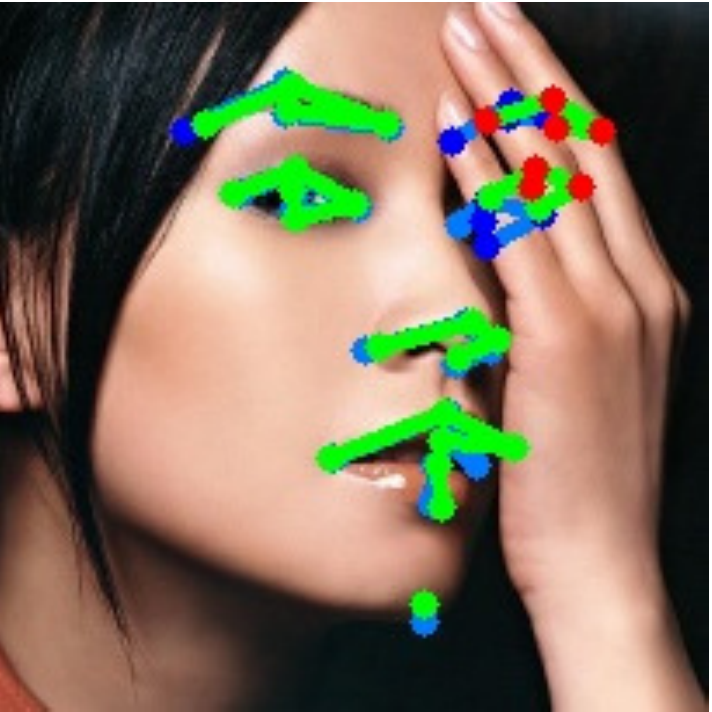}}
  \hfill
  \subfloat[AFLW]{
    \label{fig:results:d}
    \includegraphics[width=0.15\textwidth]{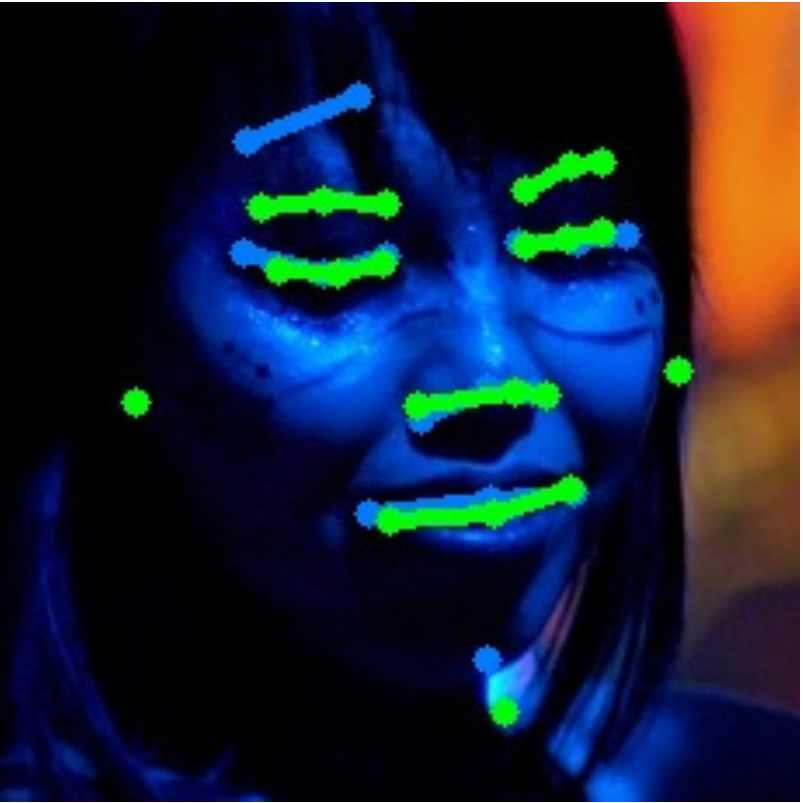}
    \includegraphics[width=0.15\textwidth]{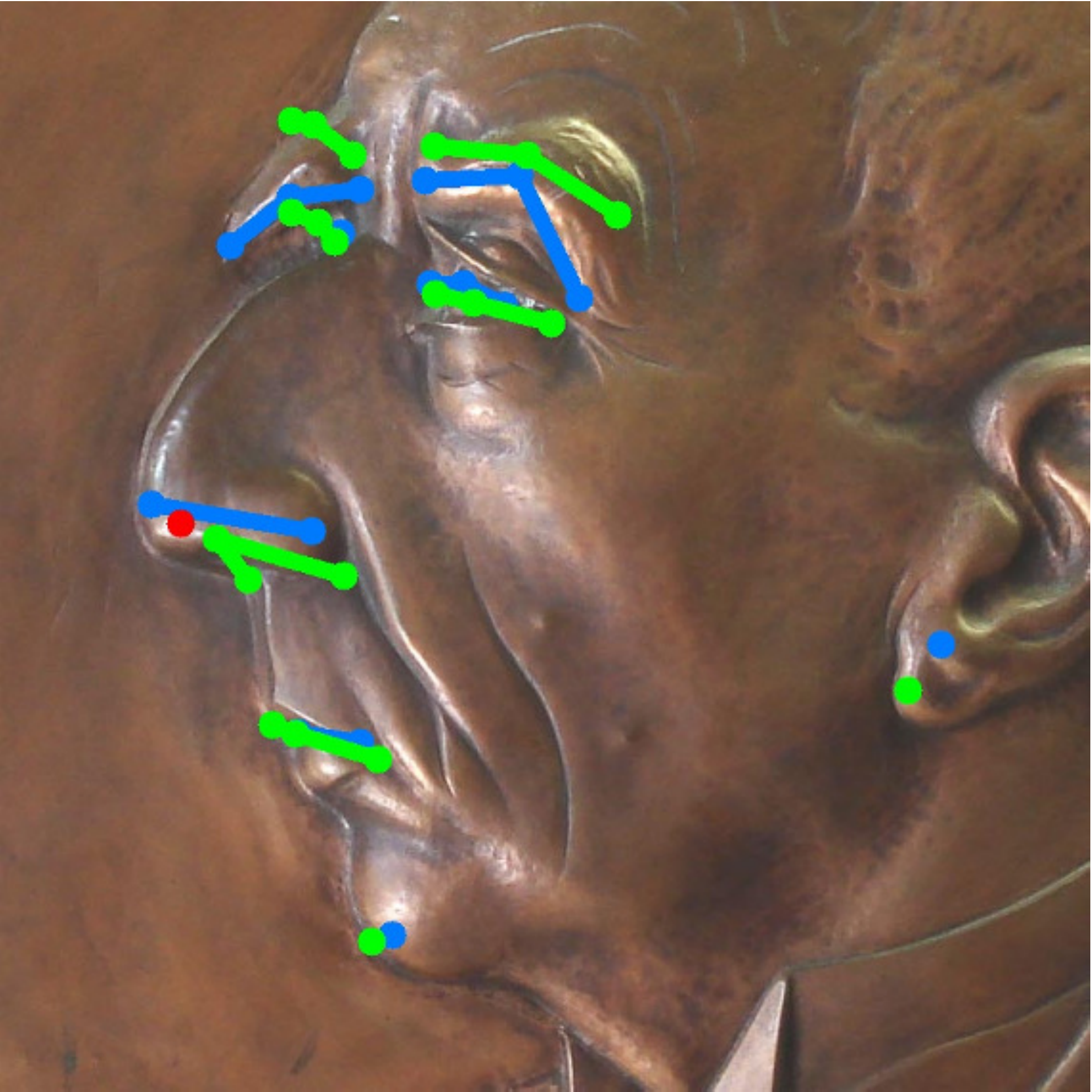}
    \includegraphics[width=0.15\textwidth]{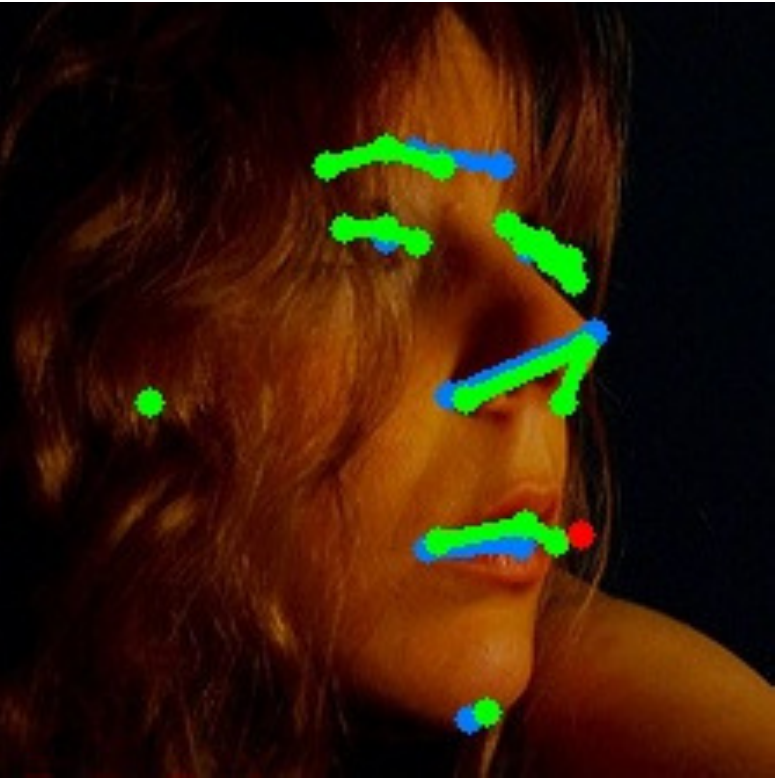}
    \includegraphics[width=0.15\textwidth]{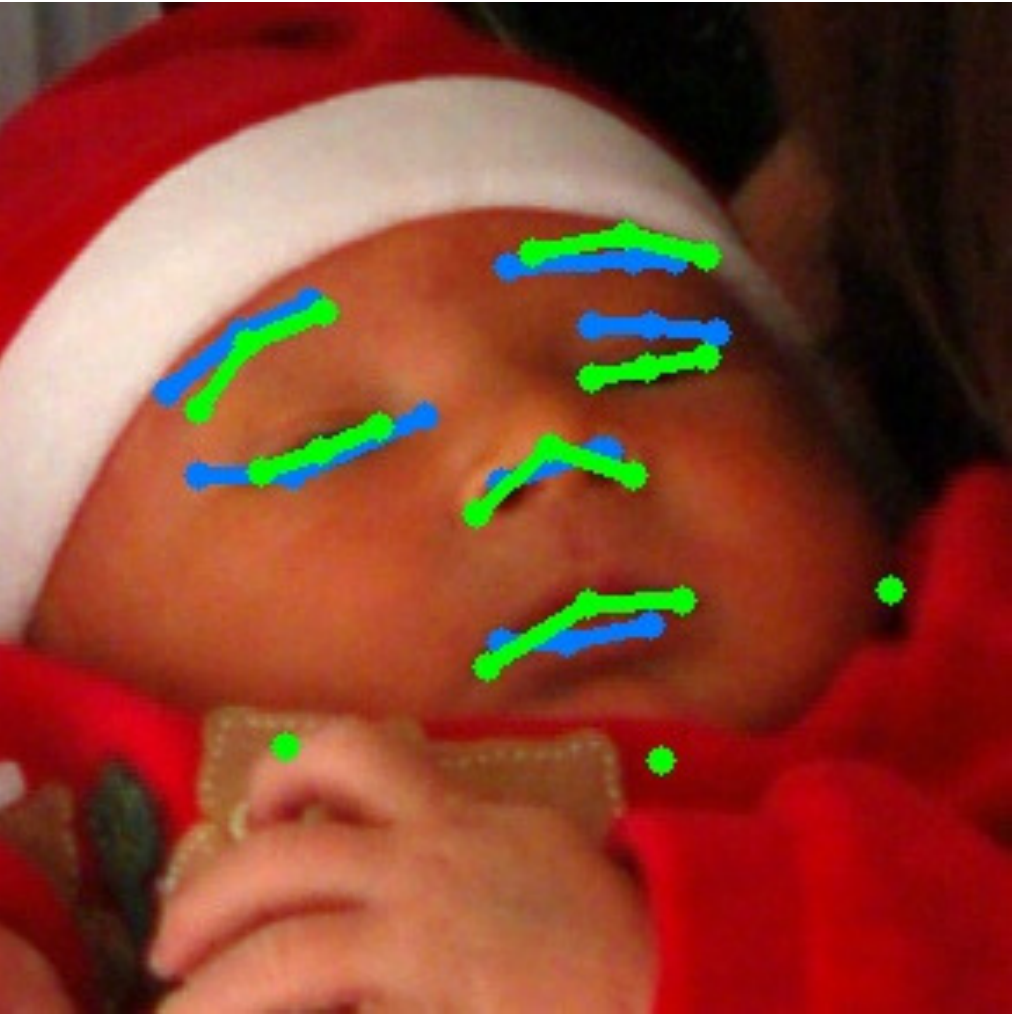}
    \includegraphics[width=0.15\textwidth]{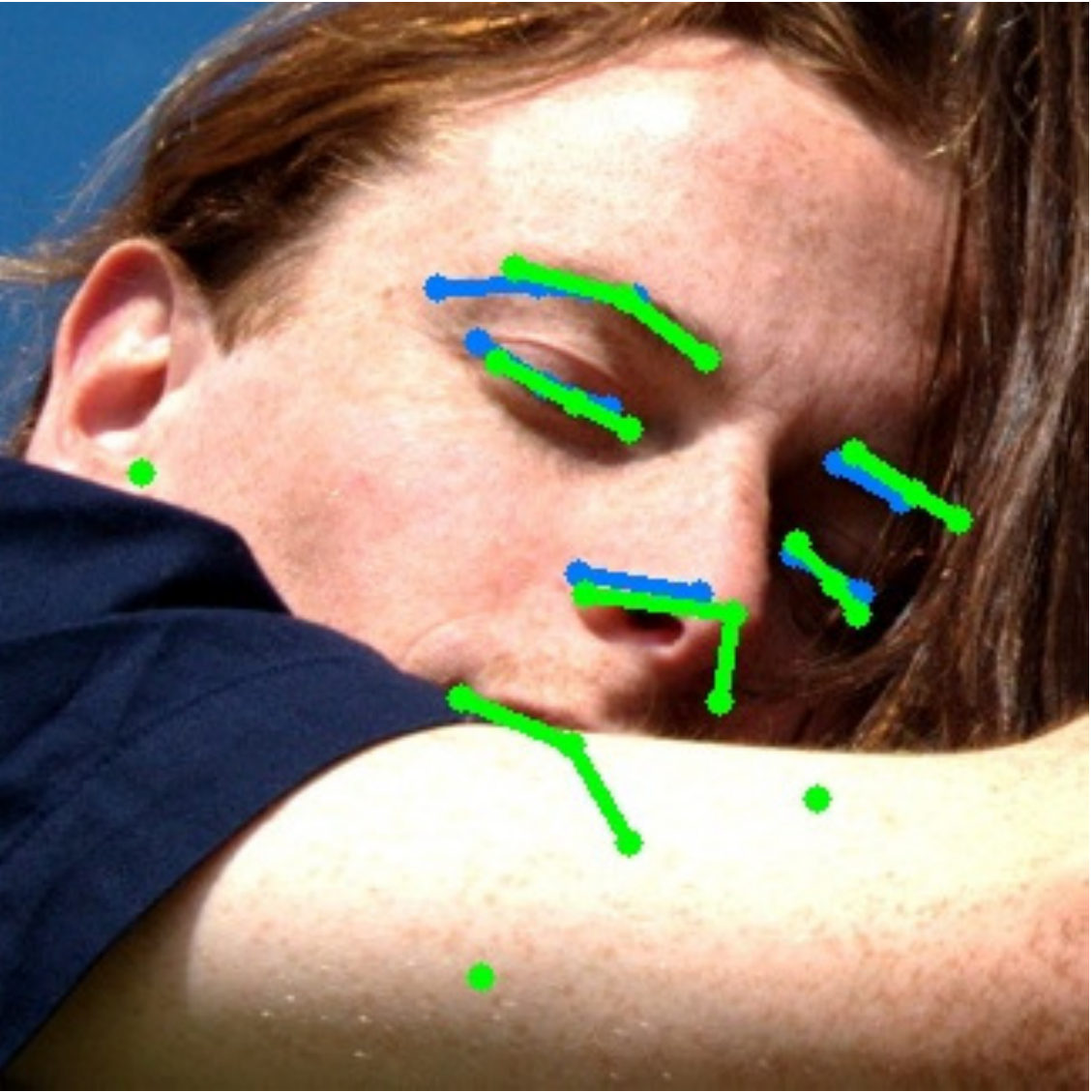}
    \includegraphics[width=0.15\textwidth]{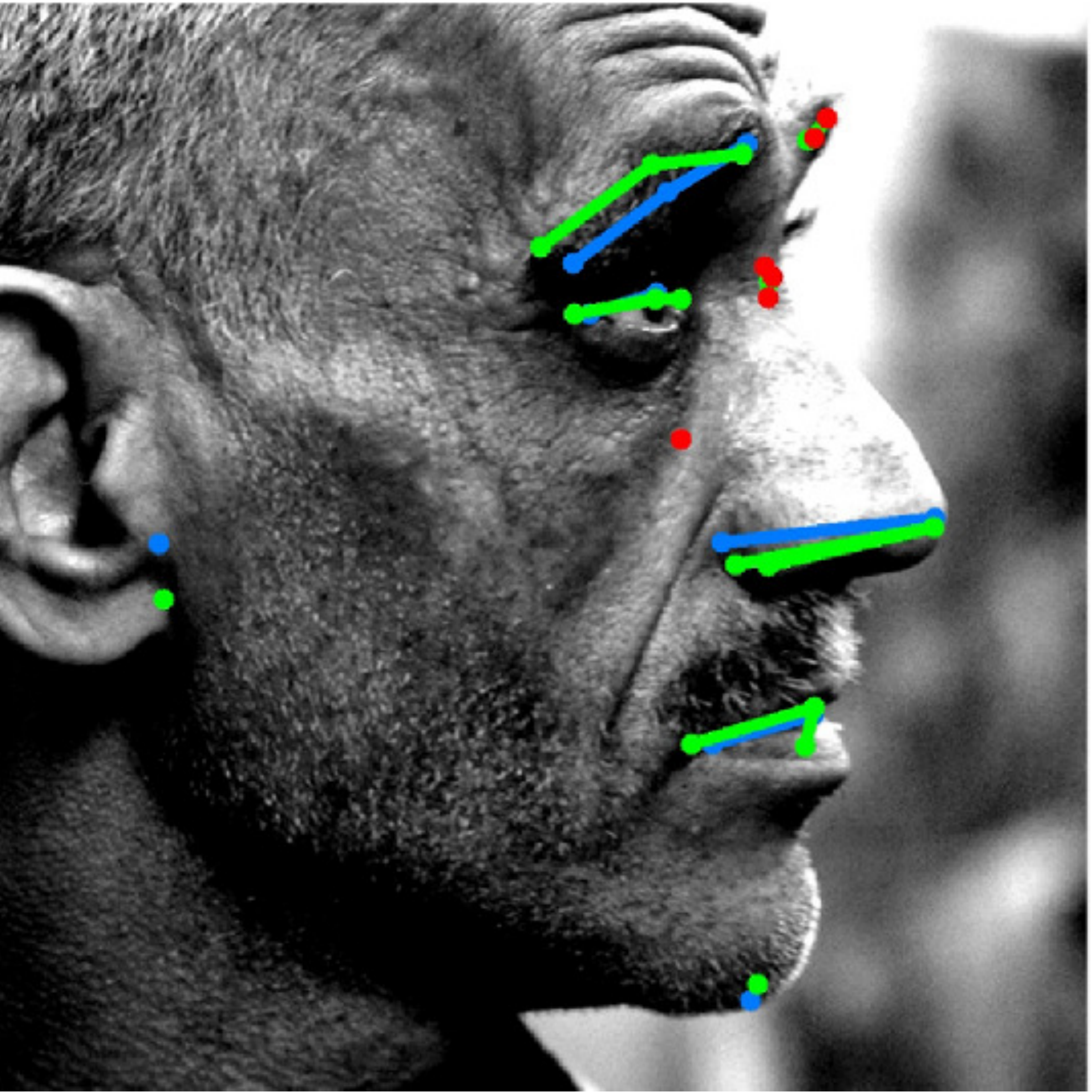}}
  \hfill
  \subfloat[WFLW]{
    \label{fig:results:e}
    \includegraphics[width=0.15\textwidth]{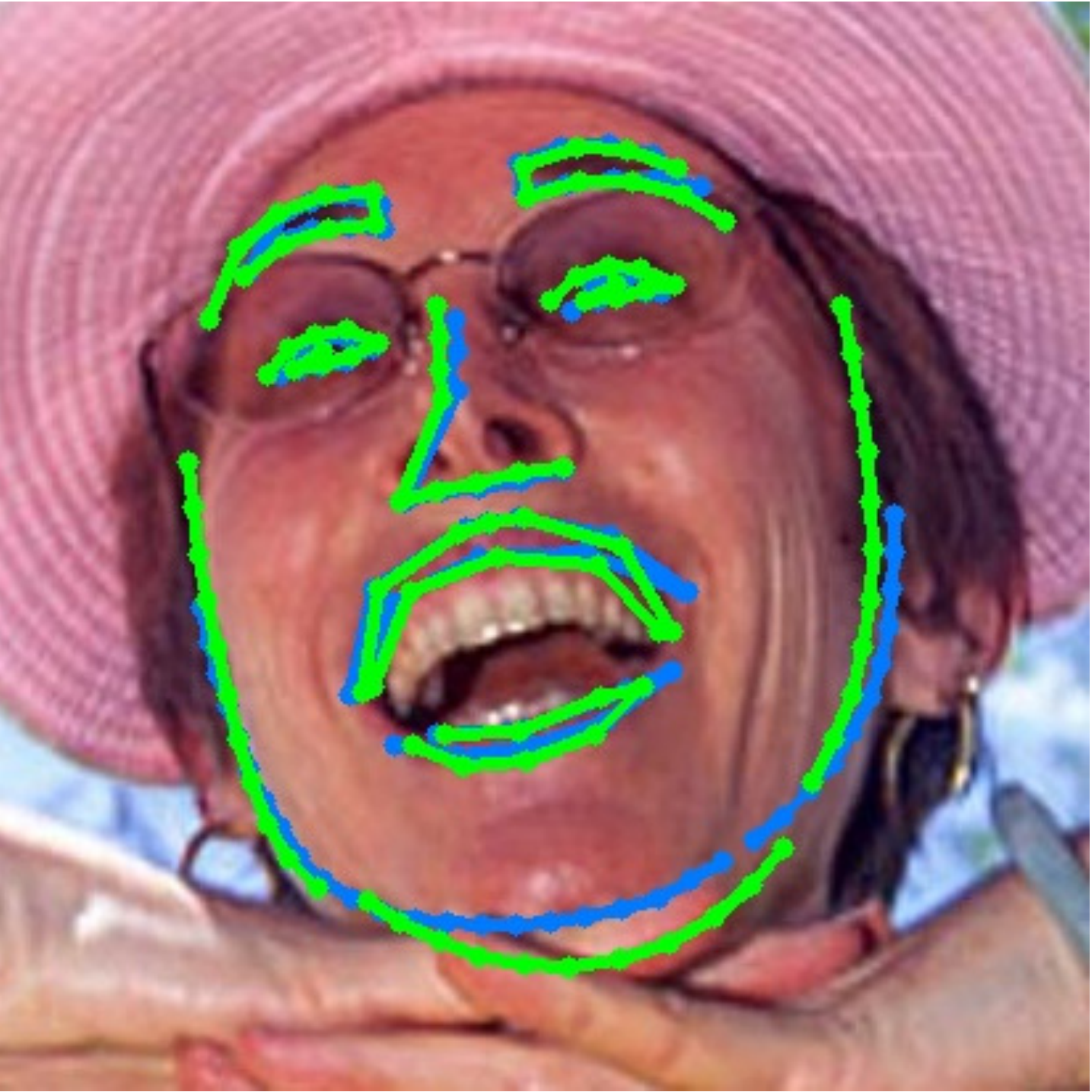}
    \includegraphics[width=0.15\textwidth]{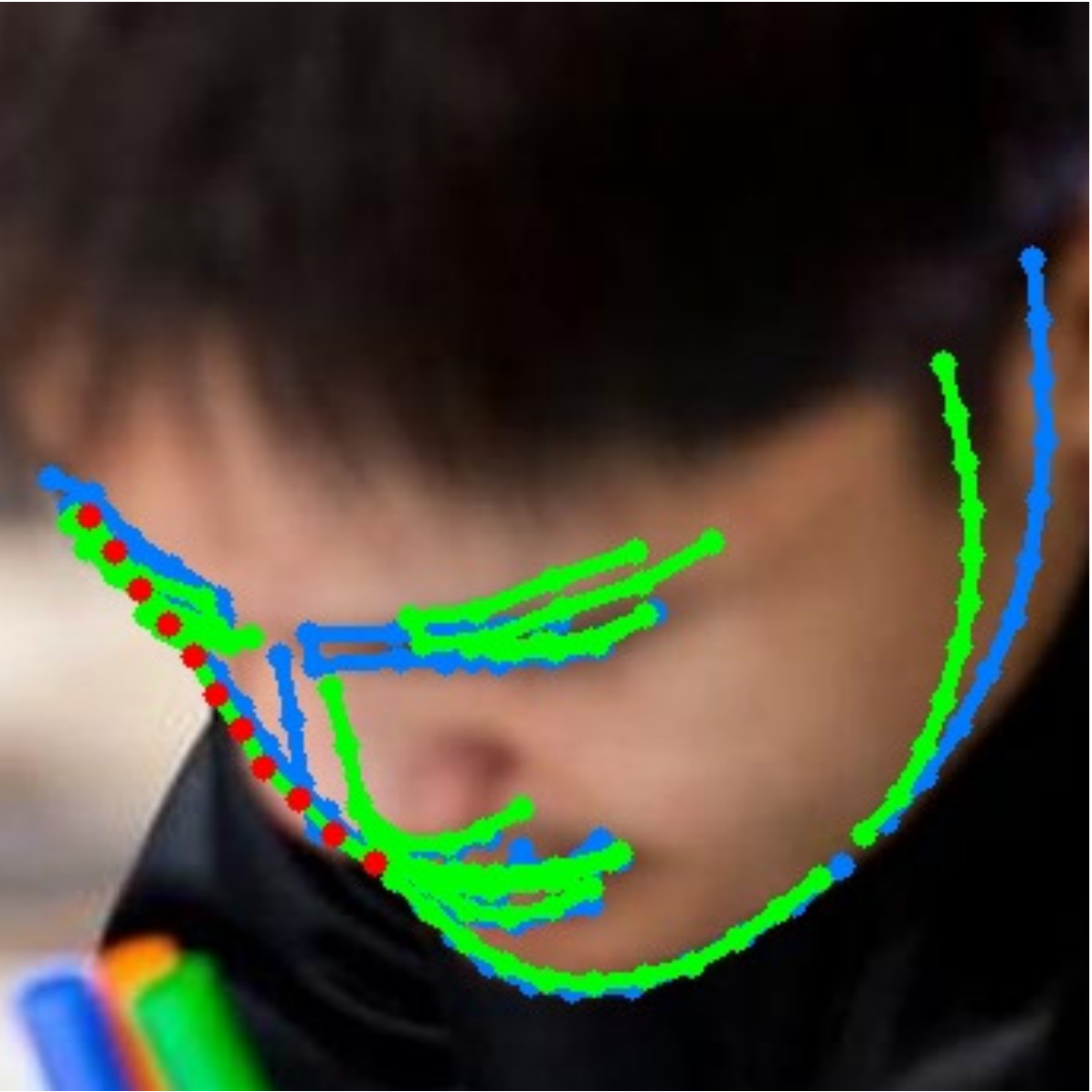}
    \includegraphics[width=0.15\textwidth]{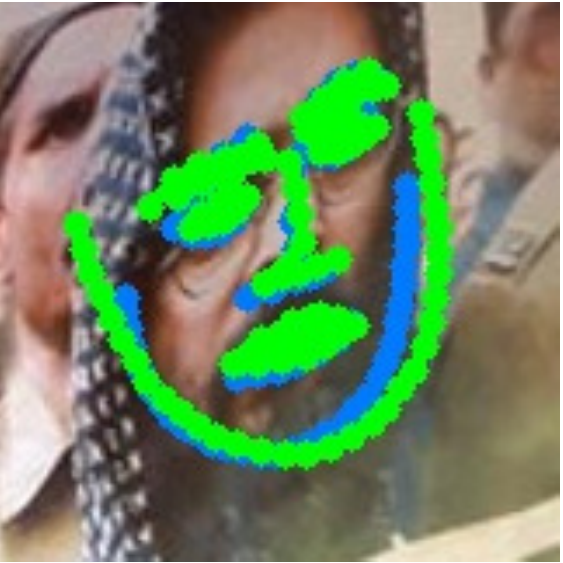}
    \includegraphics[width=0.15\textwidth]{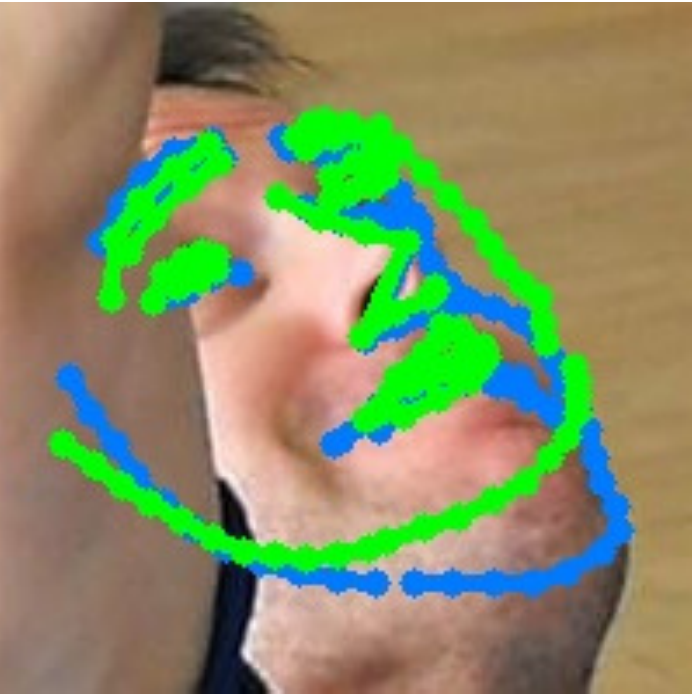}
    \includegraphics[width=0.15\textwidth]{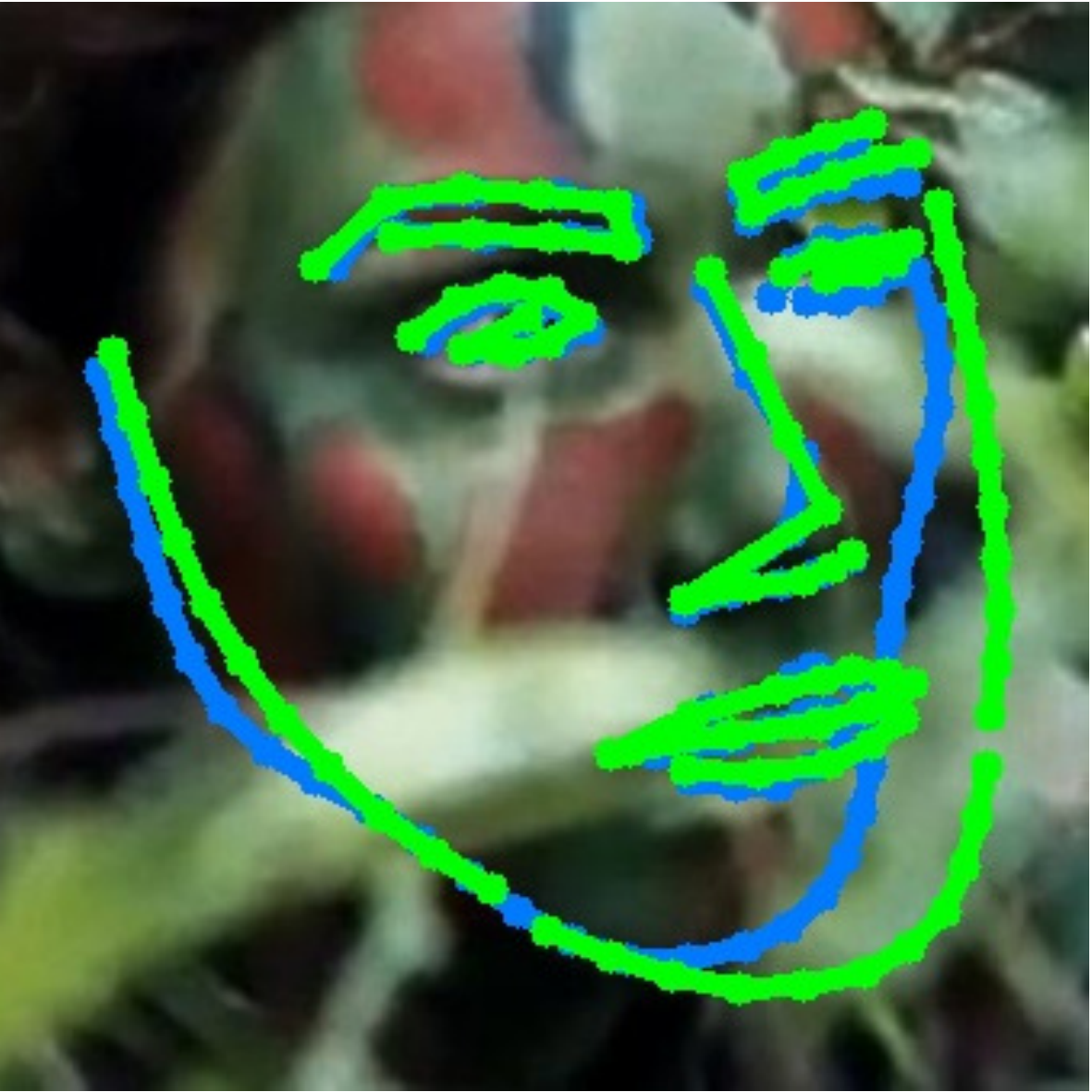}
    \includegraphics[width=0.15\textwidth]{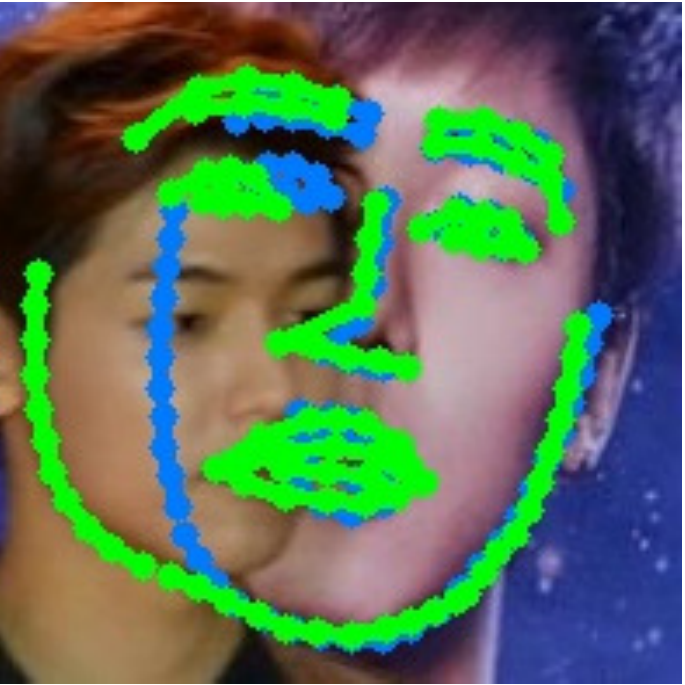}}
  \caption{Representative results considered errors using 3DDE in 300W, COFW, AFLW and WFLW testing subsets. Blue colour represents ground truth, green and red colours point out visible and non-visible shape predictions respectively.}
  \label{fig:results}
\end{figure*}

The availability of large annotated data sets has encouraged research in this area with important performance improvements in recent years. However, as shown in Fig.~\ref{fig:results}, this problem is still far from being completely solved. A critical question here is whether the models trained with present data sets will generalize to the situations present in real-life operation. The cross-dataset experiments performed reveal the existence of a significant data set bias in present benchmarks that limit the generalization of models trained with them. So, further work in this direction is required to improve the performance of present face alignment algorithms.

\ \\
\noindent{\bf Acknowledgments:} The authors gratefully acknowledge funding from the Spanish Ministry of Economy and Competitiveness, project TIN2016-75982-C2-2-R. They also thank the anonymous reviewers for their comments.


\bibliographystyle{model2-names}
\bibliography{faces}

\end{document}